\documentclass{article}

\usepackage{arxiv}

\usepackage[utf8]{inputenc}
\usepackage[T1]{fontenc}
\usepackage[numbers,sort&compress]{natbib}
\usepackage{amsmath,amssymb,amsfonts}
\usepackage{mathtools}
\usepackage{graphicx}
\usepackage{booktabs}
\usepackage{array}
\usepackage{caption}
\usepackage{subcaption}
\usepackage{float}
\usepackage{placeins}
\usepackage{url}
\usepackage{xcolor}
\usepackage{tikz}
\usetikzlibrary{positioning,arrows.meta,calc,fit,backgrounds}
\usepackage[hypertexnames=false]{hyperref}
\usepackage{authblk}

\let\ARXIVsection\section
\let\ARXIVsubsection\subsection
\let\ARXIVsubsubsection\subsubsection

\let\chapter\ARXIVsection
\let\section\ARXIVsubsection
\let\subsection\ARXIVsubsubsection

\numberwithin{equation}{section}
\numberwithin{figure}{section}
\numberwithin{table}{section}

\newcommand{\SR}{SR}
\newcommand{\HR}{HR}
\newcommand{\LR}{LR}
\newcommand{\D}{\mathbf{D}}

\title{Super-resolution as spatial downscaling: coherence, structure and task utility in Earth observation}

\renewcommand{\shorttitle}{}        
\renewcommand{\headeright}{}        
\renewcommand{\undertitle}{}        

\author[]{Bright Wiredu Nuakoh}
\affil[]{African Institute for Mathematical Sciences (AIMS), Rwanda}
\affil[]{\texttt{bright.wiredu@aims.ac.rw}}

\author[]{Ebenezer Afrifa-Yamoah}
\affil[]{School of Science, Edith Cowan University, Australia}
\affil[]{\texttt{e.afrifayamoah@ecu.edu.au}}
\date{}

\hypersetup{
  pdftitle={Super-resolution as spatial downscaling: coherence, structure and task utility in Earth observation},
  pdfauthor={Bright Wiredu Nuakoh,Ebenezer Afrifa-Yamoa},
  pdfkeywords={super-resolution, spatial downscaling, change of support, variogram, codispersion, remote sensing, task-based evaluation}
}

\begin{document}
\maketitle

\begin{abstract}
Learned satellite super-resolution is commonly assessed by pixel fidelity and potential downlink savings, but these criteria do not establish whether the reconstructed field preserves spatial structure or supports reliable downstream analysis. We treat single-image super-resolution as a spatial downscaling problem and evaluate four lightweight networks, including our compact 0.36 million-parameter model, against interpolation using pixel metrics, directional variograms, spectral band correlation, codispersion and observation coherence. On 735 satellite tiles from 190 scenes, the networks differ by at most 0.03\,dB and outperform bicubic interpolation by 1.39--1.42\,dB at $\times 2$. They recover about 30\% of the spectral power above the observation's Nyquist frequency and, at $\times 4$, produce more isotropic fields than the reference (range anisotropy 0.59 against 0.49). Energy-based metrics can favour invented detail. Enforcing coherence reaches the quantisation limit, yet at least 99.7\% of the remaining error lies in the observation operator's null space, which coherence cannot constrain. Under spatially blocked, paired evaluation with two backbones and a sharpening control, learned reconstruction changes land-use accuracy by $-1.0$ to $+3.1$ percentage points (significant in 7 of 24 comparisons) and building-change F1 by $-0.015$ to $+0.024$ (none significant) relative to bicubic interpolation, and amplifies spurious change between two observations of an unchanged scene by 1.3--1.5 times where interpolation attenuates it. Full-resolution JPEG\,2000 matches or exceeds reduced-resolution transmission with reconstruction at every matched bitrate. Learned reconstruction is most useful when only low-resolution imagery is available, and its evaluation should combine pixel metrics with spatial structure, observation coherence and task-level tests under spatial validation.




\end{abstract}
\keywords{Super-resolution, spatial downscaling, spatial evaluation, variogram, attention mechanism, lightweight neural networks}

\chapter{Introduction}
\label{ch:intro}

Satellite imagery is a record of the Earth's surface acquired by sensors carried on
orbiting platforms. It provides valuable information for applications in agriculture
\citep{10.1093/nsr/nwac290,ZHANG2020105584}, urban planning \citep{PATINO20131}, disaster
response \citep{7326589} and environmental monitoring \citep{chen2022remote}. These
applications ultimately depend
on spatial resolution: the ground area that a single pixel represents, conventionally
expressed as the ground sample distance (GSD). For example, a sensor with a $10$~m GSD
may separate fields but not the tracks between them. At a finer GSD of $1$~m, features
such as tracks, individual buildings and vehicles become detectable. Spatial resolution
therefore defines the limits of what the image can reveal and consequently the questions
it can answer. Downlink bandwidth is also often limited, so imagery may be transmitted
at a lower resolution than the sensor recorded.

This limit is largely determined at the time of image acquisition and is costly to
overcome through hardware alone. Improving the GSD typically requires larger optics,
longer focal lengths and smaller detector pixels, which increase mission cost; smaller
detector elements also capture fewer photons, which lowers the signal-to-noise ratio,
and such systems become more sensitive to diffraction \citep{park2003super}. For
agencies operating small satellites under tight budget and payload constraints, these
trade-offs are difficult to justify. An alternative is to retain the existing sensor
and enhance the spatial resolution computationally after the imagery has been acquired
and downlinked. This leads to the problem of super-resolution (\SR): estimating a
high-resolution (\HR) image from one or more low-resolution (\LR) observations of the
same scene. Reconstruction at the ground station shifts part of the computational
burden from the satellite to the ground, which makes reconstruction efficiency an
important consideration in the model design.

In remote sensing, single-image SR is a spatial downscaling problem: a value
observed over a coarse support, the footprint of a coarse pixel, is predicted over finer
supports inside it \citep{atkinson2013downscaling}. Two consequences follow. First, the
target is a spatial field, so its spatial structure (how it varies with distance and
direction, and which spatial frequencies are restored) matters as much as the error at
each pixel. Second, the prediction should be coherent with the observation:
aggregating it back to the coarse support should return the observed values
\citep{gotway2002combining,kyriakidis2004geostatistical}. Peak signal-to-noise ratio (PSNR) and the structural similarity index (SSIM), two widely used super-resolution evaluation metrics, quantify pixel-level fidelity and local structural similarity, respectively, but do not directly characterise spatial dependence and organisation as captured by measures such as directional variograms and codispersion.

\section{Motivation}

Do super-resolution models produce images that improve downstream remote-sensing tasks,
and when? A reconstruction may achieve a higher PSNR or SSIM without providing better
information for tasks such as land-use classification, change detection or object
recognition. Evaluating super-resolution only through image-quality metrics therefore
gives an incomplete picture of its value for remote sensing.

Deep learning has largely replaced classical interpolation and reconstruction methods
for single-image SR. Since the convolutional network introduced by
\citet{dong2014learning}, newer architectures have achieved substantial improvements in
reconstruction quality \citep{yue2016image,wang2021deep}, and the NTIRE efficient
super-resolution challenges \citep{li2022ntire} now compare networks of a few hundred
thousand parameters by their runtime at a fixed fidelity. Task-level evidence for satellite imagery is scarcer. It exists mainly for object
detection \citep{shermeyer2019effects}, and rarely with paired statistics and the
spatially blocked validation that autocorrelated imagery requires
\citep{roberts2017cross,ploton2020spatial}. Moreover, remote-sensing images contain
characteristics such as repeated field patterns, strong linear structures from roads and
drainage, and spatially repeated textures, and their radiometric properties vary across
seasons and locations. As \citet{dong2021remote} argue, these differences make separate
evaluation important.

This study examines whether learned reconstruction improves satellite image analysis and what an evaluation framework must measure to detect such gains. Four lightweight networks are trained under a common computational budget: three published architectures and a compact reference model, the Lightweight Single-Path Multi-Attention Network (LSMAN), designed to separate the effects of reconstruction from those of network architecture. Reconstruction quality is evaluated using complementary spatial, directional, spectral and pixel-level measures, including variograms, spectral band correlations, codispersion, coherence and conventional image quality metrics. Analytical utility is then assessed across land-use classification, building-change detection and object recognition using paired statistical tests, spatially blocked validation, a sharpening control and two feature extractors. Operational relevance is examined by comparing low-resolution transmission followed by reconstruction with full-resolution transmission at the same bitrate with JPEG\,2000, and by quantifying the artificial change introduced by reconstruction between repeated observations of an unchanged scene.

\section{Contributions}

\begin{enumerate} 
\item \textbf{Reconstruction errors lie almost entirely in the null space.} At least 99.7\% of error energy lies in the null space of the observation operator, for every LSMAN model trained under the block-average operator and at both scale factors. Observation coherence can therefore be enforced almost perfectly, but this constrains little of the actual reconstruction error and leaves substantial unquantified uncertainty. 

\item \textbf{Reconstruction manufactures change.} For unchanged scenes differing only in noise or compression, learned reconstruction amplifies differences by 1.3--1.5$\times$, whereas bicubic interpolation reduces them to 0.82--0.86$\times$. The coherence penalty does not prevent this because the effect arises in the null space. 

\item \textbf{Reconstruction does not outperform JPEG\,2000 at matched bitrate.} At matched bitrates, full-resolution JPEG\,2000 consistently matches or exceeds reconstruction in PSNR, high-frequency band correlation and land-use accuracy. Reconstruction is therefore most useful when only low-resolution imagery is available. 

\item \textbf{Structure-aware evaluation reveals what pixel metrics miss.} Energy and agreement measures must be interpreted together: the generative model restores energy but misplaces structure, whereas pixel replication adds energy through aliasing. Directional analysis further shows that reconstruction at $\times4$ increases isotropy relative to the reference, a loss undetected by pixel metrics. 

\item \textbf{Evaluation methodology matters.} Antialiased decimation and luminance definitions shift PSNR by more than 1\,dB, comparable to reported architectural gains. The calibrated pipeline reproduces published bicubic baselines within 0.022\,dB and explains the 2.14\,dB BSD100 deficit of our preliminary benchmark, which used misaligned low-resolution files from a public mirror. 
\end{enumerate} 

Despite substantial architectural differences, all four networks perform within 0.04\,dB of one another on satellite imagery, while removing attention from the reference model changes PSNR by less than 0.025\,dB. The conclusions are therefore robust to the choice among these networks.

\chapter{Related Work}
\label{ch:background}

Single-image super-resolution has moved from classical reconstruction, which treats recovery as an inverse problem solved by interpolation or regularised optimisation, to networks that learn the low-to-high-resolution mapping from data. This section reviews that development only as far as it bears on the present study: the attention mechanisms and lightweight designs that the compared networks use, the gap between the bicubic degradation these networks are trained on and the degradation of real satellite imagery, the small literature that evaluates reconstruction by its effect on analysis, and the geostatistical treatment of change of support, from which the structure and coherence measures of Section~\ref{ch:problem} are taken.

\section{Classical Reconstruction}
\label{sec:classical}

Classical reconstruction methods provide simple, non-learning-based approaches to single-image super-resolution by estimating high-resolution pixel values from the available low-resolution image. For example, bicubic interpolation estimates each output sample from the sixteen input samples in its
$4 \times 4$ neighbourhood with the cubic convolution kernel of \citet{keys1981cubic};
Lanczos interpolation uses a windowed-sinc kernel \citep{duchon1979lanczos}. Bicubic
interpolation plays two roles in this literature: it is the standard non-learned
baseline against which every method is compared, and the standard degradation operator
used to synthesise low-resolution data (Section~\ref{sec:degradation}). Interpolation
cannot recover high-frequency detail that decimation removed, so its output is blurred
at edges and textures. Iterative back-projection \citep{irani1991improving} starts from
an estimate, passes it through the imaging model, and corrects it with the difference
between the simulated and the actual observation; projection onto convex sets
\citep{stark1989high,tekalp1992high} and maximum-likelihood estimation also enforce
consistency with the observation. None of these methods gives a unique reconstruction:
decimation has a non-trivial null space, so many high-resolution images are consistent
with one observation, and the result depends on the prior or the initial estimate
(Section~\ref{sec:degradation}). Classical pipelines often follow interpolation with
Laplacian sharpening, which raises the contrast of edges and of sensor noise alike,
because a hand-specified prior cannot distinguish the two \citep{gonzalez2018digital}.
In this study, bicubic and Lanczos interpolation are the non-learned baselines,
iterative back-projection is reused as a post-hoc coherence projection
(Section~\ref{sec:coherence-results}), and bicubic interpolation followed by Laplacian
sharpening is a control condition in the downstream experiments
(Section~\ref{sec:protocol}).

\section{Learned Single-Image Super-Resolution}
\label{sec:learned}

Learning-based single-image SR methods have introduced a data-driven alternative to classical reconstruction approaches by learning mappings from low-resolution to high-resolution images. These methods learn image representations from training data and do not rely solely on predefined interpolation rules or hand-crafted image priors. The super-resolution convolutional neural network (SRCNN) of \citet{dong2014learning,dong2016image} was an early demonstration of this approach, using bicubic interpolation to upsample the low-resolution input before processing it with three convolutional layers. This pre-upsampling design increases computational cost because the convolutions operate at the target spatial resolution, although interpolation itself does not recover additional scene information. FSRCNN \citep{dong2016accelerating} and ESPCN \citep{shi2016real} therefore extract features at
low resolution and upsample only at the end. ESPCN uses sub-pixel convolution, which
produces $s^2$ feature maps and rearranges them into an image with $s$ times the spatial
resolution; LSMAN and all its competitors upsample this way. VDSR \citep{kim2016accurate}
trained a twenty-layer network to learn the residual between the interpolated input and
the high-resolution target. Residual blocks \citep{he2016deep} then became the standard
unit. EDSR \citep{lim2017enhanced} removed batch normalisation \citep{ioffe2015batch}
from them, because it can remove useful information about the intensity range, and
trained with an $\ell_1$ loss. RDN \citep{zhang2018residual} combined the outputs of all
blocks before upsampling (global feature fusion), which LSMAN also uses.

Pixel-wise losses such as $\ell_1$ and $\ell_2$ favour an average solution when several
high-resolution images correspond to the same low-resolution input, which produces smooth
reconstructions. Generative adversarial networks \citep{goodfellow2020generative} train a
generator together with a discriminator that learns to distinguish reconstructed images
from real ones. SRGAN \citep{ledig2017photo} applied this idea to SR, and ESRGAN
\citep{wang2019esrgan} extended it with residual-in-residual dense blocks and a
relativistic discriminator. These models produce sharper images with lower PSNR, and
\citet{blau2018perception} showed that distortion and perceptual quality cannot both be
optimised. A generative model must therefore be compared on a perceptual measure such as
LPIPS \citep{zhang2018unreasonable} as well as on PSNR. In this study, ESRGAN is used 
as a perceptual reference at $\times4$ (Section~\ref{sec:esrgan}).

\section{Attention Mechanisms}
\label{sec:attention}

Attention mechanisms enhance convolutional networks by allowing them to adaptively emphasise features that are more informative for a given input. Instead of treating all feature channels and spatial locations uniformly, attention modules use feature-dependent weights to recalibrate their relative contributions. Channel attention was popularised by the squeeze-and-excitation (SE) block of \citet{hu2018squeeze}, which uses global average pooling to summarise each channel, a two-layer bottleneck to generate channel-wise weights, and feature rescaling to emphasise or suppress individual channels.
 \citet{zhang2018rcan} brought it into SR in RCAN, arguing that
low-resolution inputs carry abundant low-frequency and scarce high-frequency information.
Spatial attention, on the other hand, weights locations. CBAM \citep{woo2018cbam} computes
a spatial map from the mean and maximum responses across channels and applies channel and
spatial attention in sequence, a design adopted in LSMAN. HAN \citep{niu2020single}
models channel and spatial dependencies jointly, with a parameter count comparable to
RCAN (about 16\,M). Lightweight networks use the same two forms. IMDN
\citep{hui2019lightweight} introduced contrast-aware channel attention (CCA), RFDN
\citep{liu2020residual} an enhanced spatial attention (ESA) block, and PAN
\citep{zhao2020efficient} a pixel attention that produces a full channel-and-spatial map
with 272\,K parameters. BSRN \citep{li2022blueprint} pairs ESA with CCA in a
352\,K-parameter network, so channel and spatial attention combined below one million
parameters is already a standard design.

\section{Lightweight Architectures}
\label{sec:lightweight}

Lightweight SR asks how much accuracy survives when the number of parameters,
operations or runtime is capped, conventionally at around two million parameters. CARN
\citep{ahn2018fast} used cascading residual connections. MSRN \citep{li2018multi} and
MSAN \citep{soh2020lightweight} extract features through parallel branches with
different kernel sizes or dilation rates, and MSAN replaces the CBAM spatial attention,
designed for high-level vision, with a multi-scale spatial attention module. For remote
sensing, \citet{dong2021remote} proposed SMSR, a single-path network with second-order
channel statistics evaluated on both natural and remote-sensing images. The
information-distillation lineage of IDN \citep{hui2018fast}, IMDN and RFDN splits the
features of each block into a retained and a refined part. E-RFDN, a smaller RFDN build,
won the AIM 2020 efficient SR challenge \citep{zhang2020aim}, and RLFN
\citep{kong2022residual} and BSRN led the NTIRE 2022 challenge \citep{li2022ntire}. The
NTIRE efficient SR challenges, which rank submissions by runtime, parameters and
operations at a fixed fidelity on DIV2K, are the benchmark of record of the field.
Lightweight transformers such as SwinIR-light \citep{liang2021swinir} and ELAN
\citep{zhang2022efficient}, and feature-modulation networks such as SAFMN
\citep{sun2023spatially}, reach the same fidelity with fewer parameters.
Table~\ref{tab:related} lists representative parameter budgets. Three of these networks
(IMDN, E-RFDN and CARN) are retrained in this study under the same budget as LSMAN
(Section~\ref{sec:training}).

\begin{table}[tbp]
  \centering
   \caption{Representative parameter budgets ($\times4$
    models unless stated). The last four rows are the builds trained in this study,
    with their parameter counts at $\times2$ / $\times4$.}
  \label{tab:related}
  \small
  \begin{tabular}{@{}lllr@{}}
    \toprule
    Model & Reference & Attention & Parameters \\
    \midrule
    VDSR    & \citet{kim2016accurate}     & none                          & $0.67$\,M \\
    EDSR    & \citet{lim2017enhanced}     & none                          & $43$\,M \\
    RCAN    & \citet{zhang2018rcan}       & channel                       & $16$\,M \\
    ESRGAN  & \citet{wang2019esrgan}      & none                          & $16.7$\,M \\
    CARN    & \citet{ahn2018fast}         & none                          & $1.59$\,M \\
    RFDN    & \citet{liu2020residual}     & enhanced spatial              & $550$\,K \\
    PAN     & \citet{zhao2020efficient}   & pixel (channel and spatial)   & $272$\,K \\
    BSRN    & \citet{li2022blueprint}     & enhanced spatial + contrast-aware channel & $352$\,K \\
    SwinIR-light & \citet{liang2021swinir} & window self-attention       & $878$\,K \\
    SAFMN   & \citet{sun2023spatially}    & spatially adaptive modulation & $240$\,K \\
    \midrule
    IMDN    & \citet{hui2019lightweight}  & contrast-aware channel        & $694$ / $715$\,K \\
    CARN, single-scale build &  \citet{ahn2018fast} & none                         & $0.96$ / $1.11$\,M ($\times2$ / $\times4$) \\
    E-RFDN  & \citet{liu2020residual}                 & enhanced spatial              & $417$ / $433$\,K \\
    LSMAN   & this study                 & channel then spatial          & $356$ / $605$\,K \\
    \bottomrule
  \end{tabular}
\end{table}

\section{Degradation, Remote-Sensing Reconstruction and Task-Driven Evaluation}
\label{sec:rsrelated}

Most SR models, including all those compared here, are trained on high-resolution images
decimated by a known bicubic operator. Real observations are blurred by an unknown
point-spread function, noisy and compressed. Blind and realistic degradation models
address this mismatch by training on randomised degradation pipelines
\citep{zhang2021designing,wang2021realesrgan}. A network trained for bicubic decimation
and applied to JPEG-compressed satellite tiles, as in this study, therefore operates
outside its training distribution. Remote-sensing reconstruction has also moved to
cross-sensor and real paired data: Sentinel-2 band sharpening \citep{lanaras2018super},
multi-image reconstruction of PROBA-V observations \citep{martens2019super}, and paired
low- and high-resolution acquisitions such as WorldStrat \citep{cornebise2022open} and
SEN2VEN$\mu$S \citep{michel2022sen2venus}.

Task-driven evaluation asks whether reconstruction helps an analysis.
\citet{dai2016image} found that SR helps some high-level vision tasks and not others,
\citet{haris2018task} trained SR jointly with a detector, and
\citet{shermeyer2019effects} measured object-detection performance on satellite imagery
at several resolutions and found the largest gains at the finest resolutions. Such
evaluations rarely use paired tests or spatially blocked validation, although random
splits over spatially autocorrelated data inflate accuracy
\citep{roberts2017cross,ploton2020spatial}. This study applies both.

\section{Spatial Downscaling and Change of Support}
\label{sec:downscaling}

Geostatistics treats a coarse pixel as an observation of a spatial field averaged over
its footprint, its support. Predicting the field over smaller supports is
downscaling, or area-to-point prediction, and the statistics of changing support are
well established \citep{gotway2002combining,atkinson2013downscaling}. Area-to-point
kriging is coherent: its predictions, averaged back over each coarse support,
reproduce the observed values \citep{kyriakidis2004geostatistical}. Coherence, also
called mass preservation, has recently been imposed on deep downscaling of climate fields
through constraint layers \citep{harder2023hard}. Spatial structure is described by the
variogram, the expected squared difference between values a given distance and direction
apart \citep{cressie1993statistics}, and the agreement between the structures of two
fields by their codispersion \citep{rukhin2008codispersion}. Section~\ref{ch:problem} uses
these tools to evaluate a reconstruction as a downscaled field.

\chapter{Methodology}
\label{ch:problem}

\section{The Degradation Model}
\label{sec:degradation}

Let $\mathbf{x} \in \mathbb{R}^{sH \times sW \times C}$ denote the high-resolution scene
and $\mathbf{y} \in \mathbb{R}^{H \times W \times C}$ its low-resolution observation,
where $s$ is the scale factor. The imaging chain is modelled as a blur and a decimation
followed by quantisation:
\begin{align}
\label{eq:degradation}
\mathbf{y} = \mathcal{Q}\big(\D\mathbf{x}\big),
\end{align}
where $\D$ is a linear operator that convolves with the point-spread function and
decimates, and $\mathcal{Q}$ rounds to 8 bits. Sensor noise, mainly photon shot noise,
is not modelled. Following the protocol of \citet{agustsson2017ntire},
\citet{soh2020lightweight} and \citet{dong2021remote}, low-resolution training images are generated by bicubic downsampling of high-resolution images. Models trained under this protocol therefore learn to reconstruct images under the assumed degradation process, which may differ from the actual imaging conditions of satellite sensors. For the observation-consistency analysis, we also use an averaging operator that computes the mean of each $s \times s$ block, representing the change from fine to coarse spatial support (Section~\ref{sec:downscaling}). Neither operator fully represents the effects of sensor-specific blur, noise, and compression.


Recovering $\mathbf{x}$ from $\mathbf{y}$ is ill-posed in the sense of Hadamard. $\D$
maps $s^2HWC$ unknowns to $HWC$ observations, so its null space $\mathcal{N}(\D)$ has
dimension at least $(s^2-1)HWC$, and any element of that null space can be added to a
solution without changing the observation. For the block-average operator the
decomposition is explicit. We write $\mathbf{U}$ for nearest-neighbour replication, so that
$\D\mathbf{U} = \mathbf{I}$ and $\mathbf{P} = \mathbf{U}\D$ is the orthogonal projection
onto block-constant fields. The error $\mathbf{e} = \hat{\mathbf{x}} - \mathbf{x}$ of any
estimate splits into:
\begin{align}
  \label{eq:nullsplit}
  \mathbf{e} = \underbrace{\mathbf{P}\mathbf{e}}_{\text{constrained by } \mathbf{y}}
  + \underbrace{(\mathbf{I}-\mathbf{P})\,\mathbf{e}}_{\in\,\mathcal{N}(\D)},
  \qquad
  \|\mathbf{e}\|^2 = \|\mathbf{P}\mathbf{e}\|^2 + \|(\mathbf{I}-\mathbf{P})\,\mathbf{e}\|^2 .
\end{align}
The first term is the part of the error that the observation can correct. The second is
invisible to every observation of the same support. Its share
$\|(\mathbf{I}-\mathbf{P})\mathbf{e}\|^2 / \|\mathbf{e}\|^2$ is reported in
Section~\ref{sec:coherence-results}, together with the spread of independently trained
networks as a first estimate of uncertainty.

\section{Super-Resolution as Supervised Learning}
\label{sec:learning}

Given the ill-posedness above, the learning formulation replaces the explicit prior of
classical methods with one learned from examples. Let
$\{(\mathbf{y}_n, \mathbf{x}_n)\}_{n=1}^{N}$ be a training set of low- and
high-resolution pairs generated by \eqref{eq:degradation}, and let
$F(\cdot\,;\boldsymbol{\theta})$ be a network with parameters $\boldsymbol{\theta}$.
Training minimises:
\begin{align}
  \label{eq:objective}
  \mathcal{L}(\boldsymbol{\theta}) = \frac{1}{N} \sum_{n=1}^{N}
    \Big( \big\| F(\mathbf{y}_n;\boldsymbol{\theta}) - \mathbf{x}_n \big\|_1
    + \lambda \, \big\| \D F(\mathbf{y}_n;\boldsymbol{\theta}) - \mathbf{y}_n \big\|_1 \Big),
\end{align}
where $\boldsymbol{\theta}$ collects the weights and biases of every layer. Following
\citet{lim2017enhanced}, $\ell_1$ is used in preference to $\ell_2$. The second term, with
weight $\lambda \ge 0$, penalises incoherence with the observation
(Section~\ref{sec:coherence}); $\lambda = 0$ gives the standard objective.

\section{Reconstruction as Change of Support}
\label{sec:coherence}

A downscaling estimator $F$ is coherent if aggregating its prediction to the
support of the observation returns the observation:
\begin{align}
  \label{eq:coherence}
  \D\,F(\mathbf{y}) = \mathbf{y}.
\end{align}
Area-to-point kriging satisfies \eqref{eq:coherence} by construction
\citep{kyriakidis2004geostatistical}. A convolutional network does not. Nothing in its
architecture or in the fidelity loss ties $\D F(\mathbf{y})$ to $\mathbf{y}$, and
predicting a residual over bicubic interpolation does not help, because bicubic
interpolation is not a right inverse of bicubic decimation. Coherence matters for
analysis because counting and boundary delineation depend on local mass being preserved:
a reconstruction that is brighter or darker on average over a coarse pixel than the
observation has added or removed material that the sensor did not record.

Coherence is measured by the coherence residual defined as:
\begin{align}
  \label{eq:cohres}
  r_D = \Big( \tfrac{1}{HWC} \big\| \D\hat{\mathbf{x}} - \mathbf{y} \big\|_2^2 \Big)^{1/2}
\end{align}
in grey levels. Because $\mathbf{y}$ is quantised, even the true scene is not exactly
coherent. On the 21 satellite test tiles the reference itself has $r_D = 0.29$ grey
levels for both operators and both scales, the RMS of uniform rounding error
($1/\sqrt{12}$), so residuals at or below this level are within the quantisation noise
of the observation. Section~\ref{sec:lsmancoherence} describes three ways of enforcing
coherence: the penalty in \eqref{eq:objective}, an exact constraint layer and a post-hoc
projection.

\section{Evaluation Metrics}
\label{sec:metrics}

All measures compare a reconstruction $\hat{\mathbf{x}}$ with its reference
$\mathbf{x}$ on luminance, after both are quantised to 8 bits.

\medskip
\noindent\textbf{Peak signal-to-noise ratio (PSNR).} PSNR is used to monitor reconstruction quality during training and to compare the pixel-level fidelity of the resulting models. It quantifies the difference between a reconstructed image $\hat{\mathbf{x}}$ and its reference $\mathbf{x}$ through the mean squared error (MSE), expressed in decibels (dB). It is computed as $10\log_{10}(255^2/\mathrm{MSE})$, computed on the studio-swing luma
$Y = 16 + 65.481R + 128.553G + 24.966B$ ($R,G,B \in [0,1]$, range $[16, 235]$) with a
border of $s$ pixels removed.

\medskip
\noindent\textbf{Structural similarity (SSIM).}
The SSIM
of \citet{wang2004image} compares local statistics of the reconstruction and the
reference as:
\begin{align}
  \label{eq:ssim}
  \mathrm{SSIM}(\hat{\mathbf{x}}, \mathbf{x}) =
    \frac{(2\mu_{\hat{x}}\mu_{x} + c_1)(2\sigma_{\hat{x}x} + c_2)}
         {(\mu_{\hat{x}}^2 + \mu_{x}^2 + c_1)(\sigma_{\hat{x}}^2 + \sigma_{x}^2 + c_2)},
\end{align}
averaged over the image, where $\mu$, $\sigma^2$ and $\sigma_{\hat{x}x}$ are the local
means, variances and covariance computed with an $11 \times 11$ Gaussian window of
standard deviation $1.5$ pixels, and $c_1 = (0.01 \cdot 255)^2$ and
$c_2 = (0.03 \cdot 255)^2$ are stabilising constants. Both are pixel-wise averages and say
nothing about which spatial structures were restored, so they are secondary measures in
this study.

\medskip
\noindent\textbf{Variograms.}
The empirical semivariogram of a field $z$ at lag vector $\mathbf{h}$ is:
\begin{align}
  \label{eq:variogram}
  \gamma(\mathbf{h}) = \frac{1}{2|N(\mathbf{h})|} \sum_{(i,j) \in N(\mathbf{h})}
  \big(z_i - z_j\big)^2 ,
\end{align}
where $N(\mathbf{h})$ is the set of pixel pairs separated by $\mathbf{h}$
\citep{cressie1993statistics}. It is computed for every lag up to 32 pixels with fast
Fourier transforms \citep{marcotte1996fast}, along four directions ($0^\circ$,
$45^\circ$, $90^\circ$, $135^\circ$; diagonal lags are $\sqrt{2}$ pixels per step). A
stable model $\gamma(h) = c_0 + c_1\big(1 - \exp(-3(h/a)^{\alpha})\big)$, $0<\alpha\le2$,
is fitted to each directional variogram of reconstruction and reference by weighted least
squares, giving the nugget $c_0$, the sill $c_0+c_1$, the practical range $a$ and the
anisotropy ratio (minimum over maximum directional range). The \emph{variogram error}
$\varepsilon_\gamma = \mathrm{mean}\,|\ln(\gamma_{\hat{x}}(h)/\gamma_{x}(h))|$ summarises
the four directions over lags 1--8.

\medskip
\noindent\textbf{Spectral power ratio and band correlation.}
With mean-removed, Hann-windowed Fourier transforms $\hat{X}(\mathbf{k})$ and
$X(\mathbf{k})$, radially binned by wavenumber $k$ in cycles per pixel, the power ratio
and the band correlation are:
\begin{align}
  \label{eq:spectral}
  R(k) = \frac{\sum_{|\mathbf{k}| \in k} |\hat{X}|^2}{\sum_{|\mathbf{k}| \in k} |X|^2},
  \qquad
  \rho(k) = \frac{\sum_{|\mathbf{k}| \in k} \mathrm{Re}\big(\hat{X}\,\overline{X}\big)}
  {\big(\sum_{|\mathbf{k}| \in k} |\hat{X}|^2 \, \sum_{|\mathbf{k}| \in k} |X|^2\big)^{1/2}} .
\end{align}
$R$ measures how much power a band carries and $\rho$ whether that power is in the right
place. Above the Nyquist frequency of the observation, $k \ge 1/(2s)$, nothing was
observed. A band there with $R \approx 1$ and $\rho \approx 0$ contains \emph{invented}
detail, one with $\rho \approx 1$ has been \emph{restored}, and one with $R \ll 1$ has
been \emph{lost}. The summaries $\log_{10}R_{\mathrm{sup}}$ and $\rho_{\mathrm{sup}}$ pool
all bins with $k \ge 1/(2s)$.

\medskip
\noindent\textbf{Codispersion.}
The codispersion coefficient between reconstruction and reference
\citep{rukhin2008codispersion},
\begin{align}
  \label{eq:codispersion}
  \rho_{\mathrm{c}}(\mathbf{h}) = \frac{\gamma_{\hat{x}x}(\mathbf{h})}
  {\sqrt{\gamma_{\hat{x}}(\mathbf{h})\,\gamma_{x}(\mathbf{h})}},
  \qquad
  \gamma_{\hat{x}x}(\mathbf{h}) = \frac{1}{2|N(\mathbf{h})|} \sum_{(i,j)}
  (\hat{x}_i - \hat{x}_j)(x_i - x_j),
\end{align}
is the correlation of the increments of the two fields at lag $\mathbf{h}$. It is
reported by lag and direction, and summarised by $\bar\rho_{\mathrm{c}}$, its mean over
the four directions and lags 1--4.

\medskip
\noindent\textbf{Coherence and perceptual similarity.} Coherence is measured by the
residual $r_D$ of \eqref{eq:cohres}, computed with the operator that produced the
observation. Perceptual similarity is measured by LPIPS \citep{zhang2018unreasonable}
with the AlexNet backbone, only where a generative model is compared.

\chapter{A Lightweight Single-Path Multi-Attention Network}
\label{ch:lsman}

The proposed Lightweight Single-Path Multi-Attention Network (LSMAN) combines channel and
spatial attention within a compact residual network while maintaining a single
feature-extraction path. Its design follows three considerations.

\medskip
\noindent\textbf{A single-path architecture.} Multi-scale architectures such as MSRN \citep{li2018multi} and MSAN \citep{soh2020lightweight} extract features through parallel branches with different kernel sizes or
dilation rates. These branches give access to features at different spatial scales, but
each one adds its own parameters and convolutions. LSMAN follows the single-path strategy
of \citet{dong2021remote}. The features pass sequentially through a series of residual
attention blocks, whose receptive field grows with depth, and the outputs of all blocks
are combined by global feature fusion.

\medskip
\noindent\textbf{Sequential channel and spatial attention.} Channel and spatial
attention can be modelled jointly, as in HAN \citep{niu2020single}, but at a substantial
cost in parameters. LSMAN applies channel attention first and spatial attention second,
following CBAM \citep{woo2018cbam}. The channel stage reweights the feature channels
according to their relative importance, and the spatial stage then assigns different
weights to locations within the feature map. Together the two gates cost 1.7\% of the
parameters of a block (Section~\ref{sec:complexity}).

\medskip
\noindent\textbf{Feature extraction in the low-resolution space.} LSMAN extracts
features while the feature maps remain at the size of the low-resolution input, and
upsamples only at the final stage, with a sub-pixel convolution \citep{shi2016real}.
Upsampling first would multiply the cost of every subsequent convolution by $s^2$, and
the upsampling operation does not create information that was absent from the
observation. This choice also explains why the $\times4$ network, whose input is four
times smaller, runs faster than the $\times2$ network (Section~\ref{sec:deployment}).

\begin{figure}[tbp]
  \centering
  \resizebox{\textwidth}{!}{
\definecolor{lsDark}{RGB}{28,42,51}
\definecolor{lsTeal}{RGB}{15,105,112}
\definecolor{lsOrange}{RGB}{180,84,10}
\tikzset{
  blk/.style={rounded corners=2pt, fill=#1, text=white, font=\scriptsize, align=center,
              minimum height=0.95cm, minimum width=1.0cm, inner sep=2pt},
  op/.style={circle, draw=lsDark, line width=0.5pt, fill=white, inner sep=0pt, minimum size=0.42cm, font=\footnotesize},
  io/.style={font=\footnotesize, inner sep=1pt},
  ar/.style={-{Stealth[length=1.6mm]}, line width=0.55pt, draw=lsDark},
  ln/.style={line width=0.55pt, draw=lsDark},
  skip/.style={-{Stealth[length=1.6mm]}, line width=0.55pt, draw=lsOrange},
  opt/.style={rounded corners=2pt, draw=lsDark, dashed, line width=0.5pt, text=lsDark, font=\scriptsize,
              align=center, minimum height=0.95cm, inner sep=2pt},
  lab/.style={font=\scriptsize\itshape, text=lsDark, inner sep=1pt},
  olab/.style={font=\scriptsize\itshape, text=lsOrange, inner sep=1pt},
}
\begin{tikzpicture}[node distance=0.24cm]
  \node[io] (y) {$\mathbf{y}$};
  \node[blk=lsDark, right=of y] (c0) {Conv\\$3{\times}3$};
  \node[blk=lsTeal, right=0.42cm of c0] (d1) {DAB\\1};
  \node[io, right=of d1] (dots) {$\cdots$};
  \node[blk=lsTeal, right=of dots] (dD) {DAB\\$D$};
  \node[blk=lsDark, right=0.42cm of dD] (gff) {Concat\\Conv $1{\times}1$};
  \node[op, right=of gff] (a1) {$+$};
  \node[blk=lsOrange, right=of a1] (up) {Conv, $\mathcal{PS}_s$,\\Conv};
  \node[op, right=of up] (a2) {$+$};
  \node[blk=lsDark, right=of a2, minimum width=0.8cm] (relu) {ReLU};
  \node[opt, right=of relu] (cons) {constraint\\layer (opt.)};
  \node[io, right=of cons] (xh) {$\hat{\mathbf{x}}$};
  \draw[ar] (y) -- (c0);
  \draw[ar] (c0) -- node[above, font=\scriptsize] {$\mathbf{F}_0$} (d1);
  \draw[ar] (d1) -- (dots);
  \draw[ar] (dots) -- (dD);
  \draw[ar] (dD) -- (gff);
  \draw[ar] (gff) -- (a1);
  \draw[ar] (a1) -- (up);
  \draw[ar] (up) -- (a2);
  \draw[ar] (a2) -- (relu);
  \draw[ar] (relu) -- (cons);
  \draw[ar] (cons) -- (xh);
  \coordinate (gtop) at ($(gff.north)+(0,0.42)$);
  \draw[ln] ($(c0.north)+(0.25,0)$) |- (gtop);
  \draw[ln] (d1.north) -- (d1.north |- gtop);
  \draw[ln] (dD.north) -- (dD.north |- gtop);
  \draw[ar] (gtop) -- (gff.north);
  \node[lab, anchor=south] at ($(dots |- gtop)+(0,0.05)$) {global feature fusion};
  \coordinate (lvA) at ($(c0.south)+(0,-0.40)$);
  \coordinate (lvB) at ($(c0.south)+(0,-1.00)$);
  \coordinate (lvC) at ($(c0.south)+(0,-1.60)$);
  \coordinate (r0) at ($(c0.south)+(0.25,0)$);
  \draw[skip] (r0) -- (r0 |- lvA) -| (a1.south);
  \node[olab, anchor=north] at ($(dots |- lvA)+(0,-0.04)$) {global residual};
  \draw[skip] (y.south) -- (y.south |- lvB) -| (a2.south);
  \node[olab, anchor=north] at ($(dD |- lvB)+(0,-0.04)$) {bicubic interpolation $\mathcal{B}_s(\mathbf{y})$ (residual formulation)};
  \draw[skip, dashed] (y.south |- lvB) -- (y.south |- lvC) -| (cons.south);
  \node[olab, anchor=north] at ($(gff |- lvC)+(0,-0.04)$) {observation $\mathbf{y}$: \ $\hat{\mathbf{x}} \leftarrow \hat{\mathbf{x}} + \mathbf{U}(\mathbf{y}-\mathbf{D}\hat{\mathbf{x}})$};
  \node[font=\small, anchor=south west] at ($(y.west |- gtop)+(-0.05,0.35)$) {(a) Network};

  \begin{scope}[yshift=-4.4cm]
  \node[io] (fin) {$\mathbf{F}_{d-1}$};
  \node[blk=lsDark, right=0.35cm of fin] (k1) {Conv $3{\times}3$\\ReLU};
  \node[blk=lsDark, right=of k1] (k2) {Conv\\$3{\times}3$};
  \node[blk=lsTeal, right=0.45cm of k2, minimum width=2.6cm] (ca) {channel gate\\GAP, FC, ReLU, FC, ReLU};
  \node[op, right=of ca] (m1) {$\times$};
  \node[blk=lsTeal, right=0.35cm of m1, minimum width=2.6cm] (sa) {spatial gate\\$[\text{mean};\text{max}]$, Conv $7{\times}7$, ReLU};
  \node[op, right=of sa] (m2) {$\times$};
  \node[op, right=of m2] (a3) {$+$};
  \node[io, right=0.3cm of a3] (fout) {$\mathbf{F}_{d}$};
  \draw[ar] (fin) -- (k1);
  \draw[ar] (k1) -- (k2);
  \draw[ar] (k2) -- node[below, font=\scriptsize] {$\mathbf{G}_d$} (ca);
  \draw[ar] (ca) -- (m1);
  \draw[ar] (m1) -- (sa);
  \draw[ar] (sa) -- (m2);
  \draw[ar] (m2) -- (a3);
  \draw[ar] (a3) -- (fout);
  \coordinate (gsplit) at ($(k2.east)!0.5!(ca.west)$);
  \draw[ar] (gsplit) -- ++(0,0.80) -| (m1.north);
  \coordinate (msplit) at ($(m1.east)!0.5!(sa.west)$);
  \draw[ar] (msplit) -- ++(0,0.80) -| (m2.north);
  \coordinate (lvD) at ($(k1.south)+(0,-0.40)$);
  \coordinate (r1) at ($(fin.east)+(0.1,0)$);
  \draw[skip] (r1) -- (r1 |- lvD) -| (a3.south);
  \node[olab, anchor=north] at ($(ca |- lvD)+(0,-0.04)$) {local residual connection};
  \node[font=\small, anchor=south west] at ($(fin.west)+(-0.05,0.95)$) {(b) Dual-attention block (DAB)};
  \end{scope}
\end{tikzpicture}}
  \caption{The proposed LSMAN architecture. (a) Shallow feature extraction, a single
    path of $D=6$ dual-attention blocks, fusion of all block outputs, a global residual
    connection, sub-pixel upsampling ($\mathcal{PS}_s$), and a residual over bicubic
    interpolation of the observation followed by a ReLU output. The constraint layer
    (dashed) is used only by the hard-coherence variant. (b) The dual-attention block:
    channel attention followed by spatial attention, both with ReLU gates, within a
    local residual connection.}
  \label{fig:architecture}
\end{figure}
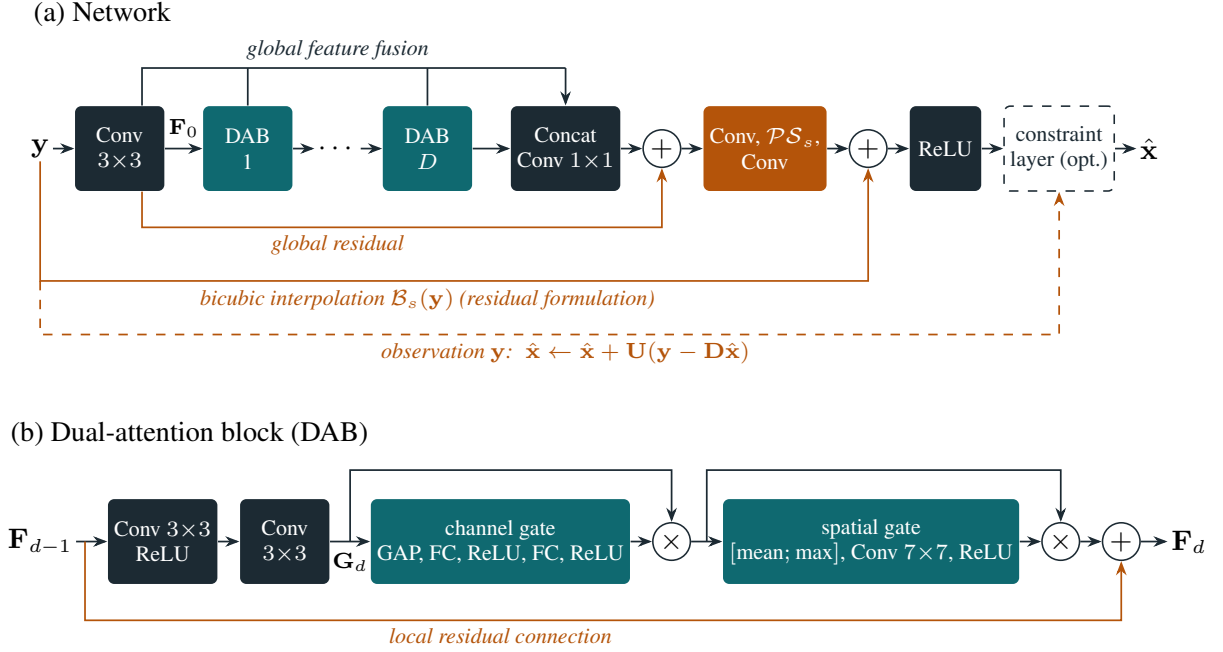

\section{The Dual-Attention Block}
\label{sec:dab}

The dual-attention block (DAB) is the basic processing unit of LSMAN and is repeated
throughout the network (Figure~\ref{fig:architecture}b). The block first transforms the
incoming features, then applies channel and spatial attention in sequence, and finally
adds the result to its input through a residual connection. Let
$\mathbf{F}_{d-1} \in \mathbb{R}^{H \times W \times C}$ denote the input to the $d$th
block, where $C$ is the number of feature channels.

\medskip
\noindent\textbf{Feature transformation.} Two $3 \times 3$ convolutions produce the
residual representation as follows:
\begin{align}
  \label{eq:dabconv}
  \mathbf{G}_d = \mathbf{W}_{d}^{(2)} * \delta\!\left(
      \mathbf{W}_{d}^{(1)} * \mathbf{F}_{d-1} + \mathbf{b}_{d}^{(1)}
    \right) + \mathbf{b}_{d}^{(2)},
\end{align}
where $\delta(t) = \max(0, t)$ is the rectified linear unit (ReLU), applied after the
first convolution only.

\medskip
\noindent\textbf{Channel attention.} Each channel is first summarised by global spatial
averaging:
\begin{align}
  \label{eq:squeeze}
  z_c = \frac{1}{HW} \sum_{i=1}^{H} \sum_{j=1}^{W} \mathbf{G}_d(i,j,c),
  \qquad c = 1,\dots,C,
\end{align}
and the descriptor $\mathbf{z} \in \mathbb{R}^{C}$ is passed through a two-layer
bottleneck to obtain a non-negative gate per channel:
\begin{align}
  \label{eq:excite}
  \mathbf{a} = \delta\!\left( \mathbf{W}_{\uparrow}\, \delta(\mathbf{W}_{\downarrow}\, \mathbf{z}
    + \mathbf{b}_{\downarrow}) + \mathbf{b}_{\uparrow} \right),
  \qquad
  \tilde{\mathbf{G}}_d(i,j,c) = a_c\, \mathbf{G}_d(i,j,c),
\end{align}
with $\mathbf{W}_{\downarrow} \in \mathbb{R}^{C' \times C}$,
$\mathbf{W}_{\uparrow} \in \mathbb{R}^{C \times C'}$, $C' = \max(\lceil C/r\rceil, 4)$ and
reduction ratio $r$. Since the descriptor is obtained from the whole feature map, each
channel weight reflects information aggregated across the image.

\medskip
\noindent\textbf{Spatial attention.} After channel reweighting, two maps are obtained by
pooling across the channels with the mean and maximum operations:
\begin{align}
  \label{eq:spatialpool}
  \mathbf{A}_{\mathrm{avg}}(i,j) =
  \frac{1}{C} \sum_{c=1}^{C} \tilde{\mathbf{G}}_d(i,j,c),
  \qquad
  \mathbf{A}_{\max}(i,j) = \max_{1 \le c \le C} \tilde{\mathbf{G}}_d(i,j,c),
\end{align}
and a $7 \times 7$ convolution of the two maps gives a non-negative spatial gate shared by
all channels:
\begin{align}
  \label{eq:spatialgate}
  \mathbf{M}_d = \delta\!\left(
    \mathbf{W}_{S} * \big[\, \mathbf{A}_{\mathrm{avg}} \,;\,
    \mathbf{A}_{\max} \,\big] + b_S \right),
  \qquad
  \hat{\mathbf{G}}_d(i,j,c) = \mathbf{M}_d(i,j)\, \tilde{\mathbf{G}}_d(i,j,c).
\end{align}
The $7 \times 7$ kernel provides a wider spatial context than the preceding $3 \times 3$
convolutions when estimating the importance of each location.

\medskip
\noindent\textbf{Local residual connection.} The attention-refined residual is added to
the input, $\mathbf{F}_d = \mathbf{F}_{d-1} + \hat{\mathbf{G}}_d$. Following
\citet{he2016deep}, the residual connection preserves a direct path for the input
features, so a block can retain its input when its learned contribution is small.

The gates in \eqref{eq:excite} and \eqref{eq:spatialgate} use the ReLU in place of the
logistic (sigmoid) function of the preliminary configuration of LSMAN, denoted LSMAN-$\sigma$, which has logistic gates and output and was trained for 15{,}000 steps (Section~\ref{sec:head}). They are therefore not
confined to $(0,1)$ and can amplify as well as suppress. They are identity-initialised:
the last weights of each gate start at zero and its bias at one, so a new block is a plain
residual block with gates equal to one, and training does not start from saturated or
inactive gates.

\section{Network Architecture}
\label{sec:lsmanarch}

The network has four stages (Figure~\ref{fig:architecture}a). Shallow feature extraction
maps the observation to $C$ channels with one $3 \times 3$ convolution,
$\mathbf{F}_0 = \mathbf{W}_0 * \mathbf{y} + \mathbf{b}_0$. Deep feature extraction passes
$\mathbf{F}_0$ through $D$ dual-attention blocks,
$\mathbf{F}_d = \mathrm{DAB}_d(\mathbf{F}_{d-1})$. Global feature fusion concatenates the
outputs of all blocks with $\mathbf{F}_0$ and reduces the $(D+1)C$ channels to $C$ with a
$1 \times 1$ convolution \citep{zhang2018residual}, and a global residual connection adds
$\mathbf{F}_0$:
\begin{align}
  \label{eq:gff}
  \mathbf{F}_{\mathrm{DF}} = \mathbf{F}_0 + \mathbf{W}_{\mathrm{GFF}} *
    \big[\, \mathbf{F}_1 ,\, \dots ,\, \mathbf{F}_D ,\, \mathbf{F}_0 \,\big]
    + \mathbf{b}_{\mathrm{GFF}} .
\end{align}
Since each $\mathbf{F}_d$ has passed through a different number of blocks, the fusion
lets the reconstruction draw on both shallow and deep features, and the global residual
gives the initial representation a direct path to the output. Upsampling and
reconstruction expands $\mathbf{F}_{\mathrm{DF}}$ to $Cs^2$ channels with a convolution,
rearranges them with the periodic shuffle $\mathcal{PS}_s$, which has no learnable
parameters, into an $sH \times sW \times C$ map, and maps the result to three channels
with a final convolution. The network predicts the residual over bicubic interpolation
$\mathcal{B}_s(\mathbf{y})$ of the observation, followed by a ReLU output:
\begin{align}
  \label{eq:reconstruct}
  \hat{\mathbf{x}} = \delta\!\left( \mathcal{B}_s(\mathbf{y}) +
    \mathbf{W}_R * \mathcal{PS}_s\big(\mathbf{W}_{U} * \mathbf{F}_{\mathrm{DF}}
      + \mathbf{b}_{U}\big) + \mathbf{b}_R \right).
\end{align}
The ReLU keeps the output non-negative without saturating, and the residual formulation
was selected on DIV2K validation PSNR (Section~\ref{sec:head}). This study uses $C = 48$
channels, $D = 6$ blocks and $r = 8$.

\section{Coherence-Constrained Variants}
\label{sec:lsmancoherence}

Coherence (Section~\ref{sec:coherence}) is enforced in three ways.

\medskip
\noindent\textbf{Soft penalty.} The network is trained with $\lambda > 0$ in
\eqref{eq:objective}, under either observation operator. $\lambda$ is the largest value
in $\{0.1, 1, 10\}$ whose DIV2K validation PSNR stays within 0.05\,dB of $\lambda = 0$.

\medskip
\noindent\textbf{Hard constraint layer.} For the block-average operator, whose right
inverse is nearest-neighbour replication $\mathbf{U}$, a final layer
\begin{align}
  \label{eq:hard}
  \hat{\mathbf{x}} \leftarrow \hat{\mathbf{x}} + \mathbf{U}\big(\mathbf{y} - \D\hat{\mathbf{x}}\big)
\end{align}
makes $\D\hat{\mathbf{x}} = \mathbf{y}$ exactly before quantisation, as in the constraint
layers of \citet{harder2023hard}. With a ReLU output, the network inside this layer can
stop learning. In two of three training seeds at $\times2$ its output became zero, the
layer returned $\mathbf{U}\mathbf{y}$ (the observation replicated), and the result was
perfectly coherent but uninformative (30.86\,dB on DIV2K validation, against 34.98\,dB
for the seed that trained). The constraint models therefore use a linear output, with
which all three seeds trained normally at both scales (34.95--34.96\,dB at $\times2$ and
29.67--29.69\,dB at $\times4$).

\medskip
\noindent\textbf{Post-hoc projection.} The output of an unconstrained network is
projected onto $\{\hat{\mathbf{x}} : \D\hat{\mathbf{x}} = \mathbf{y}\}$ at inference,
without retraining: by one step of \eqref{eq:hard} for the block-average operator, and by
iterative back-projection \citep{irani1991improving},
$\hat{\mathbf{x}}^{(t+1)} = \hat{\mathbf{x}}^{(t)} + \mathcal{B}_s(\mathbf{y} -
\D\hat{\mathbf{x}}^{(t)})$, for the bicubic operator.

\section{Parameter Complexity}
\label{sec:complexity}

Table~\ref{tab:params} gives the number of learnable parameters of each component. A
convolution with a $k \times k$ kernel from $C_{\mathrm{in}}$ to $C_{\mathrm{out}}$
channels has $k^2 C_{\mathrm{in}} C_{\mathrm{out}} + C_{\mathrm{out}}$ parameters, and the
derived totals match the built networks exactly. The two attention gates account for
729 of the $42{,}297$ parameters of a block (1.7\%), so most of the cost of a block lies
in its two feature convolutions. Both configurations remain well below the
two-million-parameter threshold associated with lightweight SR models
\citep{ahn2018fast,soh2020lightweight}. The small cost of attention also makes an
ablation with a parameter-matched control cheap to train (Section~\ref{sec:ablation}).

\begin{table}[tbp]
  \centering
  \caption{Parameter budget of LSMAN ($C = 48$, $D = 6$, $r = 8$, $C' = 6$). Only the
    upsampling convolution depends on the scale factor, which is why the $\times4$
    network is larger.}
  \label{tab:params}
  \small
  \begin{tabular}{@{}l l r r@{}}
    \toprule
    Component & Expression & $\times 2$ & $\times 4$ \\
    \midrule
    Shallow convolution & $9 \cdot 3 C + C$ & $1{,}344$ & $1{,}344$ \\
    Feature convolutions (per block) & $2(9C^2 + C)$ & $41{,}568$ & $41{,}568$ \\
    Channel attention (per block) & $2CC' + C' + C$ & $630$ & $630$ \\
    Spatial attention (per block) & $49 \cdot 2 + 1$ & $99$ & $99$ \\
    \addlinespace
    All $D = 6$ blocks & $D \times 42{,}297$ & $253{,}782$ & $253{,}782$ \\
    Global fusion convolution & $(D{+}1)C \cdot C + C$ & $16{,}176$ & $16{,}176$ \\
    Upsampling convolution & $9C \cdot Cs^2 + Cs^2$ & $83{,}136$ & $332{,}544$ \\
    Reconstruction convolution & $9C \cdot 3 + 3$ & $1{,}299$ & $1{,}299$ \\
    \midrule
    \textbf{Total} & & $\mathbf{355{,}737}$ & $\mathbf{605{,}145}$ \\
    \bottomrule
  \end{tabular}
\end{table}

\chapter{Experimental Setup}
\label{ch:experiments}

\section{Datasets} \label{sec:setup} \noindent\textbf{Training and benchmarks.} All networks are trained on the 800 training images of DIV2K \citep{agustsson2017ntire}, a standard dataset for image super-resolution. Low-resolution images are generated by cropping each high-resolution image to dimensions divisible by the scale factor $s$, applying antialiased bicubic downsampling, and quantising the result to 8 bits. On the DIV2K validation set, the regenerated images differ from the officially released low-resolution images by an average of 0.149 grey levels at $\times2$ and 0.104 grey levels at $\times4$, with fewer than 0.02\% of pixels differing by more than one grey level. Evaluation uses Set5 \citep{bevilacqua2012low}, Set14 \citep{zeyde2012single}, BSD100 \citep{martin2001database}, Urban100 \citep{huang2015single}, and the 100 DIV2K validation images. All images are evaluated at full size, trimmed only to a multiple of $s$.

\medskip
\noindent\textbf{Satellite tiles.}
The satellite imagery is derived from the DeepGlobe Land Cover Classification
dataset \citep{demir2018deepglobe}. It comprises RGB images from DigitalGlobe's
Vivid+ product at a 50\,cm ground sample distance, covering areas in Thailand,
Indonesia, and India. The source scenes are JPEG images of
$2{,}448 \times 2{,}448$ pixels, with no acquisition dates provided. The tiles were collected without a record of their source; by matching them
to the source scenes, we identified a pool of 5{,}108 tiles, including the
21 test tiles, as exact crops from 190 scenes in the training split
(Appendix~\ref{app:tiles}). Each scene is divided into a $9 \times 9$ grid of
contiguous $256 \times 256$ tiles, each covering approximately
$128 \times 128$\,m. Two evaluation sets are used. \emph{Sat-Test21} contains the 21 tiles of the preliminary experiment, from 15 scenes,
with no two tiles touching; the closest pair is separated by 362 pixels
(181\,m) (Appendix~\ref{app:data}). \emph{Sat-Pool} is a scene-balanced sample of 735 pool tiles from all
190 scenes, at most four per scene; it shares no tiles with Sat-Test21 but includes its 15 scenes.
No satellite imagery is used for training, allowing evaluation of how well
networks trained on natural images transfer to remote sensing. Because tiles
from the same scene are spatially dependent, statistical analyses resample
scenes rather than individual tiles (Section~\ref{sec:stats}).

Because the source scenes are JPEG files, the reference tiles are not an uncompressed scene but a JPEG rendering of one. Part of the structure above the Nyquist frequency of the observation may be block structure introduced by the codec, which no reconstruction should reproduce, so the band correlation and power ratio reported in Section~\ref{sec:structure} are bounded below unity by the reference itself and are read as relative comparisons between methods rather than as absolute measures of recovery.

\medskip
\noindent\textbf{Downstream data.} UC Merced \citep{yang2010bag} provides 2{,}100 aerial
images of $256 \times 256$ pixels at about 0.3\,m GSD in 21 land-use classes. LEVIR-CD
\citep{chen2020spatial} provides bi-temporal $256 \times 256$ patches of the same
locations, cut from $1024 \times 1024$ images, with building-change masks; a patch is
labelled as changed when more than 1\% of its pixels change, and the test split has
2{,}048 patches. NWPU VHR-10 \citep{cheng2014multi} provides 3{,}896 annotated objects in ten classes (Appendix~\ref{app:tasks}).

\section{Calibration of the Measurement Path}
\label{sec:calibration}

Reliable comparison of super-resolution (SR) methods requires consistent
low-resolution image generation and evaluation metrics. We calibrated the
measurement pipeline using antialiased bicubic downsampling and the luminance
definition described in Section~\ref{sec:metrics}. Before model comparisons,
the pipeline was checked against the reference bicubic PSNR values for all
four benchmark datasets at both scale factors. The resulting values differed
from the reference values by no more than 0.022\,dB. Detailed calibration
results are provided in Appendix~\ref{app:calibration-path}.
Benchmark calibration identified discrepancies arising from image resizing
and luminance computation. In particular, non-integer
downsampling factors can misalign reconstructed images with their references,
while single-channel images require intensity-based evaluation instead of
colour-to-luminance conversion. No image in the
present study is centre-cropped, ensuring that the reported
results reflect the complete image content. 
Consistent image degradation and luminance handling are important for
reproducible super-resolution evaluation. Calibration experiments showed
that these conventions can shift bicubic PSNR by more than 1\,dB for some
benchmarks. We therefore calibrated the measurement pipeline against
reference bicubic results before comparing the evaluated networks (Appendix~\ref{app:calibration}).

\section{Training Configuration and Competitors}
\label{sec:training}

Training follows the protocols of \citet{soh2020lightweight} and
\citet{dong2021remote} (Table~\ref{tab:training}), using $48 \times 48$
low-resolution patches, batches of 16, random flips and rotations, an
$\ell_1$ objective \eqref{eq:objective}, and the Adam optimiser
\citep{kingma2015adam}. The initial learning rate is $4 \times 10^{-4}$,
halved every 10{,}000 steps. Each network is trained for 30{,}000 steps
with three random seeds, using the final weights without checkpoint selection
or early stopping. The competitors are IMDN \citep{hui2019lightweight}, E-RFDN
\citep{liu2020residual,zhang2020aim}, and single-scale CARN
\citep{ahn2018fast}, retrained under the same budget. Their published results
are included for context only because they used substantially longer training.
ESRGAN \citep{wang2019esrgan} is evaluated from a public checkpoint at
$\times4$ as a perceptual reference. Training curves are provided in
Appendix~\ref{app:training}. At $\times4$, all networks gain less than
0.02\,dB over the final 10\% of training; at $\times2$, CARN, LSMAN, and
E-RFDN gain 0.03--0.09\,dB.

\begin{table}[tbp]
  \centering
  \caption{Training configuration, identical for every network, scale factor and seed.}
  \label{tab:training}
  \small
  \begin{tabular}{@{}l l@{}}
    \toprule
    Setting & Value \\
    \midrule
    Training data & DIV2K, 800 images, low-resolution images regenerated (Section~\ref{sec:setup}) \\
    Patch size, batch size & $48 \times 48$ low-resolution patches, batch of 16 \\
    Augmentation & random horizontal and vertical flips, $90^\circ$ rotations \\
    Loss & $\ell_1$, Eq.~\eqref{eq:objective} with $\lambda = 0$ unless stated \\
    Optimiser & Adam \citep{kingma2015adam}, $\beta = (0.9, 0.999)$, $\epsilon = 10^{-8}$ \\
    Learning rate & $4 \times 10^{-4}$, halved every $10{,}000$ steps \\
    Steps & $30{,}000$ (60 epochs of 500 steps) \\
    Seeds & 0, 1, 2; final weights, no checkpoint selection \\
    Hardware & four NVIDIA RTX 3090 GPUs, PyTorch 2.5.1 \\
    \bottomrule
  \end{tabular}
\end{table}



\section{Statistical Analysis}
\label{sec:stats}

All method comparisons are paired by image. For satellite imagery, differences
are averaged within scenes, with 95\% confidence intervals estimated by
bootstrap resampling of scenes (10{,}000 resamples) and $p$-values calculated
using the Wilcoxon signed-rank test \citep{wilcoxon1945individual}. Results
from three random seeds are averaged per image before testing. Classifier
accuracies are compared using the exact McNemar test
\citep{mcnemar1947note,dietterich1998approximate}, while F1 scores are
compared using bootstrap resampling of source images (2{,}000 resamples)
\citep{efron1993introduction}. Holm's procedure adjusts $p$-values within
each comparison family, separately at each scale factor
\citep{holm1979simple}. Statistical significance is interpreted alongside
effect size and seed-to-seed variability, since small differences may be
statistically significant without being practically meaningful. The gap recovered by a condition is
$G = (M_{\mathrm{cond}} - M_{\mathrm{LR}})/(M_{\mathrm{HR}} - M_{\mathrm{LR}})$, where $M$
is the task metric under the condition, under the observation and under the
high-resolution reference. $G<0$ means worse than the observation and $G>1$ better than
the reference. Because the denominator can be small and noisy, $G$ is reported with a
cluster-bootstrap 95\% interval (5{,}000 resamples) and suppressed whenever the interval
of the denominator $M_{\mathrm{HR}} - M_{\mathrm{LR}}$ contains zero. A
pixel-versus-structure \emph{disagreement} is counted for a pair of methods when PSNR and
a structure measure both differ significantly between them (Holm-adjusted Wilcoxon
$p<0.05$ on scene means) but favour different methods.

\section{Downstream Protocol}
\label{sec:protocol}

At $\times2$ and $\times4$, each task compares the HR reference, native LR
observation (pixel-replicated to the reference size), bicubic interpolation,
bicubic interpolation with Laplacian sharpening
($g=f-\nabla^2 f$, using the four-neighbour discrete Laplacian), LSMAN,
IMDN (the strongest matched-budget competitor on DIV2K validation), and
coherence-penalised LSMAN. Each reference image is decimated to simulate
a coarser observation, and all conditions are evaluated at the reference
size without further resizing. Sharpening controls for gains attributable
to input sharpness rather than learned reconstruction.

Classification uses frozen ImageNet ResNet-50 \citep{he2016deep} and
MobileNetV2 \citep{sandler2018mobilenetv2} feature extractors with
logistic-regression probes on standardised pooled features, and an
end-to-end fine-tuned ResNet-50 on UC Merced (12 epochs, AdamW,
one-cycle learning-rate schedule). Probes use matched-condition training
and testing or are trained on HR references and tested on each condition.
For change detection, the probe classifies the absolute difference between
features extracted from the two dates.

Validation accounts for spatial dependence. LEVIR-CD uses its official
split, with F1 bootstrapped over source images. For UC Merced and NWPU
VHR-10, near-duplicate images are grouped by connected components of
cosine similarity above 0.9 between centred ResNet-50 features; five-fold
stratified cross-validation keeps clusters within a single fold. UC Merced
contains 1,892 clusters, the largest with 70 images.

False-change sensitivity is tested on 1{,}113 unchanged LEVIR-CD test
patches. For each patch, two observations differ in Gaussian noise
($\sigma=1$ or 2 grey levels), JPEG quality (85 versus 95), or both,
and are reconstructed independently. We report the fraction of patches
flagged as changed, the fraction of pixels with RGB change-vector
magnitude above 10 grey levels, and amplification, defined as the RMS
difference between reconstructions divided by that between observations.
\chapter{Reconstruction Results}
\label{ch:fidelity}

\section{Output Formulation and Activation}
\label{sec:head}

\begin{figure}[tbp]
  \centering
  \includegraphics[width=\textwidth]{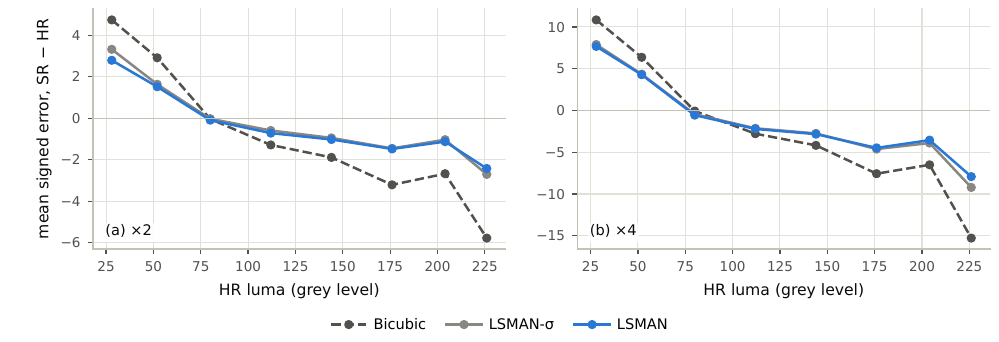}
  \caption{Mean signed reconstruction error by reference brightness on
  Sat-Pool at $\times2$ (a) and $\times4$ (b), comparing bicubic interpolation,
  LSMAN-$\sigma$ and LSMAN (seed 0).}
  \label{fig:bias}
\end{figure}

We evaluated six output configurations: direct prediction, residual
prediction over bicubic interpolation and EDSR-style mean shifting, each with a
ReLU or linear output activation (Table~\ref{tab:head}). Residual prediction achieves the highest DIV2K validation PSNR at both scales (35.43\,dB at $\times2$ and 29.92\,dB at $\times4$),
improving by 0.09--0.17\,dB over direct prediction with the same activation.
Direct prediction with ReLU has the greatest seed variability ($\pm0.16$\,dB
at $\times2$), while residual prediction yields the lowest Sat-Pool coherence
residual (0.20 grey levels at $\times2$, versus 0.40--0.42 for direct prediction).
We select residual prediction based on mean DIV2K validation PSNR
across scales, without using test results for model selection; its ReLU and linear variants tie, and the ReLU is kept so that the output stays non-negative (Section~\ref{sec:lsmanarch}).
\begin{table}[tbp]
  \centering
  \caption{Output-head comparison using ReLU attention gates. PSNR (dB) is
  reported as mean $\pm$ standard deviation over three seeds after
  30{,}000 training steps. $^{\dagger}$ marks the configuration selected
  using DIV2K validation results.}
  \label{tab:head}
  \small
\begin{tabular}{@{}lrrrrrr@{}}
\toprule
Formulation, output & DIV2K-val & Set14 & BSD100 & Urban100 & Sat-Test21 & Sat-Pool \\
\midrule
\multicolumn{7}{@{}l}{\emph{Scale $\times2$}} \\
direct, linear & 35.32{\scriptsize\,$\pm$\,0.02} & 32.86{\scriptsize\,$\pm$\,0.01} & 31.74{\scriptsize\,$\pm$\,0.01} & 30.55{\scriptsize\,$\pm$\,0.00} & 34.36{\scriptsize\,$\pm$\,0.02} & 35.71{\scriptsize\,$\pm$\,0.02} \\
direct, ReLU & 35.26{\scriptsize\,$\pm$\,0.16} & 32.83{\scriptsize\,$\pm$\,0.09} & 31.72{\scriptsize\,$\pm$\,0.05} & 30.54{\scriptsize\,$\pm$\,0.06} & 34.34{\scriptsize\,$\pm$\,0.03} & 35.67{\scriptsize\,$\pm$\,0.07} \\
mean-shift, linear & 35.41{\scriptsize\,$\pm$\,0.02} & 32.95{\scriptsize\,$\pm$\,0.02} & 31.78{\scriptsize\,$\pm$\,0.01} & 30.68{\scriptsize\,$\pm$\,0.02} & 34.38{\scriptsize\,$\pm$\,0.03} & 35.73{\scriptsize\,$\pm$\,0.03} \\
mean-shift, ReLU & 35.42{\scriptsize\,$\pm$\,0.02} & 32.94{\scriptsize\,$\pm$\,0.03} & 31.79{\scriptsize\,$\pm$\,0.01} & 30.69{\scriptsize\,$\pm$\,0.04} & 34.37{\scriptsize\,$\pm$\,0.02} & 35.72{\scriptsize\,$\pm$\,0.01} \\
residual, linear & 35.43{\scriptsize\,$\pm$\,0.01} & 32.94{\scriptsize\,$\pm$\,0.02} & 31.80{\scriptsize\,$\pm$\,0.02} & 30.73{\scriptsize\,$\pm$\,0.02} & 34.39{\scriptsize\,$\pm$\,0.01} & 35.74{\scriptsize\,$\pm$\,0.01} \\
residual, ReLU$^{\dagger}$ & 35.43{\scriptsize\,$\pm$\,0.02} & 32.93{\scriptsize\,$\pm$\,0.01} & 31.79{\scriptsize\,$\pm$\,0.02} & 30.72{\scriptsize\,$\pm$\,0.04} & 34.38{\scriptsize\,$\pm$\,0.02} & 35.73{\scriptsize\,$\pm$\,0.02} \\
\addlinespace
\multicolumn{7}{@{}l}{\emph{Scale $\times4$}} \\
direct, linear & 29.79{\scriptsize\,$\pm$\,0.05} & 27.93{\scriptsize\,$\pm$\,0.04} & 27.19{\scriptsize\,$\pm$\,0.01} & 25.06{\scriptsize\,$\pm$\,0.01} & 29.62{\scriptsize\,$\pm$\,0.00} & 31.00{\scriptsize\,$\pm$\,0.01} \\
direct, ReLU & 29.83{\scriptsize\,$\pm$\,0.01} & 27.96{\scriptsize\,$\pm$\,0.01} & 27.20{\scriptsize\,$\pm$\,0.01} & 25.08{\scriptsize\,$\pm$\,0.00} & 29.62{\scriptsize\,$\pm$\,0.00} & 31.01{\scriptsize\,$\pm$\,0.00} \\
mean-shift, linear & 29.89{\scriptsize\,$\pm$\,0.01} & 28.01{\scriptsize\,$\pm$\,0.02} & 27.24{\scriptsize\,$\pm$\,0.00} & 25.15{\scriptsize\,$\pm$\,0.01} & 29.63{\scriptsize\,$\pm$\,0.01} & 31.02{\scriptsize\,$\pm$\,0.01} \\
mean-shift, ReLU & 29.88{\scriptsize\,$\pm$\,0.00} & 28.02{\scriptsize\,$\pm$\,0.01} & 27.24{\scriptsize\,$\pm$\,0.00} & 25.15{\scriptsize\,$\pm$\,0.01} & 29.63{\scriptsize\,$\pm$\,0.01} & 31.02{\scriptsize\,$\pm$\,0.01} \\
residual, linear & 29.92{\scriptsize\,$\pm$\,0.01} & 28.04{\scriptsize\,$\pm$\,0.01} & 27.27{\scriptsize\,$\pm$\,0.01} & 25.22{\scriptsize\,$\pm$\,0.01} & 29.63{\scriptsize\,$\pm$\,0.00} & 31.02{\scriptsize\,$\pm$\,0.01} \\
residual, ReLU$^{\dagger}$ & 29.92{\scriptsize\,$\pm$\,0.00} & 28.04{\scriptsize\,$\pm$\,0.01} & 27.26{\scriptsize\,$\pm$\,0.00} & 25.22{\scriptsize\,$\pm$\,0.01} & 29.64{\scriptsize\,$\pm$\,0.01} & 31.03{\scriptsize\,$\pm$\,0.01} \\
\bottomrule
\end{tabular}

\end{table}
To test whether the logistic output of LSMAN-$\sigma$ biases dark and bright pixels,
Figure~\ref{fig:bias} shows the mean signed error by brightness. All methods, including
interpolation, brighten dark pixels and darken bright pixels. At $\times4$, bicubic interpolation exhibits the
largest bias ($+10.8$ and $-15.3$ grey levels in the darkest and brightest
bins, respectively). LSMAN-$\sigma$ slightly increases the negative
bias in the brightest bin ($-9.2$ versus $-7.9$ grey levels for LSMAN),
although these pixels represent fewer than 0.05\% of the tiles.

\section{Attention Ablation}
\label{sec:ablation}

Six attention configurations were evaluated with three seeds at each scale:
no attention, channel-only, spatial-only, channel-then-spatial (the design of
Section~\ref{ch:lsman}), spatial-then-channel, and a parameter-matched control
that adds the two gate outputs instead of multiplying them (Table~\ref{tab:ablation}).
All configurations differ by at most 0.04\,dB, with paired differences from the
no-attention network below 0.025\,dB, comparable to seed variability. Although
some Sat-Pool differences are statistically detectable, their direction reverses
across scales: channel attention gains 0.012\,dB at $\times2$ but loses
0.013\,dB at $\times4$. The parameter-matched control performs similarly,
providing no evidence of a measurable benefit from attention at this model
budget. Both gates are retained for consistency across experiments.

\begin{figure}[tbp]
  \centering
  \includegraphics[width=\textwidth]{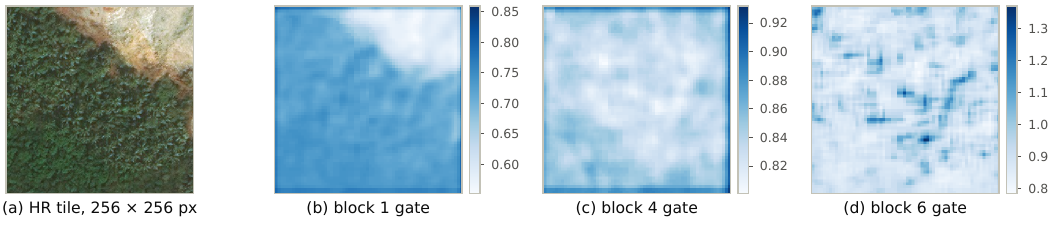}
  \caption{Spatial attention gates of the $\times4$ LSMAN (seed 0) on the most textured
    Sat-Test21 tile. (a) Reference tile. (b)--(d) Gate $\mathbf{M}_d$ of blocks 1, 4 and
    6, computed on the $64 \times 64$ observation and shown at the size of the tile (each
    gate value repeated over its $4 \times 4$ block); colour scales differ per block.
    The gates use a ReLU, so values above 1 amplify.}
  \label{fig:attention}
\end{figure}

The spatial gates in the $\times4$ network vary with depth
(Figure~\ref{fig:attention}). The first follows land cover, assigning lower
values to bare soil than to canopy; the middle is nearly uniform (0.80--0.93),
while the last is mostly between 0.8 and 1, with isolated peaks of 1.37 that
amplify the residual. Given the negligible ablation differences, these maps
illustrate gate behaviour but do not establish its contribution to reconstruction.
\begin{table}[tbp]
  \centering
  \caption{Attention ablation (PSNR, dB, mean $\pm$ s.d. over three seeds). The last
    column is the paired per-image difference from the network without attention on
    Sat-Pool (seed-averaged; scene-cluster bootstrap and Wilcoxon tests in the data
    files). Parameter counts are for the built networks.}
  \label{tab:ablation}
  \small
\begin{tabular}{@{}lrrrrrr@{}}
\toprule
 & & \multicolumn{4}{c}{PSNR (dB), mean $\pm$ s.d. over 3 seeds} & $\Delta$ vs none \\
\cmidrule(lr){3-6}
Attention configuration & Parameters & DIV2K-val & Urban100 & Sat-Test21 & Sat-Pool & Sat-Pool (dB) \\
\midrule
\multicolumn{7}{@{}l}{\emph{Scale $\times2$}} \\
no attention & 351{,}363 & 35.45{\scriptsize\,$\pm$\,0.02} & 30.74{\scriptsize\,$\pm$\,0.03} & 34.38{\scriptsize\,$\pm$\,0.02} & 35.73{\scriptsize\,$\pm$\,0.02} & -- \\
channel only & 355{,}143 & 35.44{\scriptsize\,$\pm$\,0.01} & 30.74{\scriptsize\,$\pm$\,0.02} & 34.39{\scriptsize\,$\pm$\,0.01} & 35.74{\scriptsize\,$\pm$\,0.01} & $+$0.012 \\
spatial only & 351{,}957 & 35.46{\scriptsize\,$\pm$\,0.01} & 30.76{\scriptsize\,$\pm$\,0.03} & 34.39{\scriptsize\,$\pm$\,0.02} & 35.74{\scriptsize\,$\pm$\,0.02} & $+$0.010 \\
channel $\rightarrow$ spatial (LSMAN) & 355{,}737 & 35.43{\scriptsize\,$\pm$\,0.02} & 30.74{\scriptsize\,$\pm$\,0.03} & 34.38{\scriptsize\,$\pm$\,0.02} & 35.74{\scriptsize\,$\pm$\,0.02} & $+$0.008 \\
spatial $\rightarrow$ channel & 355{,}737 & 35.43{\scriptsize\,$\pm$\,0.01} & 30.72{\scriptsize\,$\pm$\,0.02} & 34.37{\scriptsize\,$\pm$\,0.03} & 35.72{\scriptsize\,$\pm$\,0.03} & $-$0.004 \\
parameter-matched control & 355{,}737 & 35.44{\scriptsize\,$\pm$\,0.03} & 30.76{\scriptsize\,$\pm$\,0.05} & 34.38{\scriptsize\,$\pm$\,0.05} & 35.73{\scriptsize\,$\pm$\,0.05} & $+$0.004 \\
\addlinespace
\multicolumn{7}{@{}l}{\emph{Scale $\times4$}} \\
no attention & 600{,}771 & 29.92{\scriptsize\,$\pm$\,0.00} & 25.20{\scriptsize\,$\pm$\,0.01} & 29.64{\scriptsize\,$\pm$\,0.01} & 31.03{\scriptsize\,$\pm$\,0.00} & -- \\
channel only & 604{,}551 & 29.92{\scriptsize\,$\pm$\,0.00} & 25.22{\scriptsize\,$\pm$\,0.00} & 29.62{\scriptsize\,$\pm$\,0.02} & 31.02{\scriptsize\,$\pm$\,0.02} & $-$0.013 \\
spatial only & 601{,}365 & 29.92{\scriptsize\,$\pm$\,0.01} & 25.21{\scriptsize\,$\pm$\,0.02} & 29.64{\scriptsize\,$\pm$\,0.01} & 31.03{\scriptsize\,$\pm$\,0.01} & $-$0.002 \\
channel $\rightarrow$ spatial (LSMAN) & 605{,}145 & 29.92{\scriptsize\,$\pm$\,0.01} & 25.21{\scriptsize\,$\pm$\,0.01} & 29.63{\scriptsize\,$\pm$\,0.02} & 31.03{\scriptsize\,$\pm$\,0.01} & $-$0.005 \\
spatial $\rightarrow$ channel & 605{,}145 & 29.92{\scriptsize\,$\pm$\,0.00} & 25.21{\scriptsize\,$\pm$\,0.01} & 29.63{\scriptsize\,$\pm$\,0.01} & 31.03{\scriptsize\,$\pm$\,0.01} & $-$0.003 \\
parameter-matched control & 605{,}145 & 29.92{\scriptsize\,$\pm$\,0.01} & 25.22{\scriptsize\,$\pm$\,0.01} & 29.64{\scriptsize\,$\pm$\,0.01} & 31.03{\scriptsize\,$\pm$\,0.01} & $+$0.001 \\
\bottomrule
\end{tabular}

\end{table}

\section{Matched-Budget Comparison}
\label{sec:matched}

Table~\ref{tab:benchmarks} compares LSMAN with IMDN, E-RFDN and CARN, all trained with the
same data, patches, optimiser, schedule, steps and seeds. Although the three competitors
achieve higher PSNR on natural-image benchmarks, with gains of 0.06--0.32\,dB at $\times2$
and 0.05--0.22\,dB at $\times4$, LSMAN remains competitive with fewer parameters than all of them at $\times2$ and all but E-RFDN at
$\times4$. The gaps are largest on Urban100, where every paired per-image comparison on
Urban100 and BSD100 favours the competitors (Holm-adjusted $p < 0.001$; for example,
IMDN gains 0.33\,dB [0.28, 0.37] on Urban100 at $\times2$). Published results for the same
architectures are a further 0.2--1.1\,dB higher. SSIM follows the same ordering
(Appendix~\ref{app:training}).

\begin{figure}[tbp]
  \centering
  \includegraphics[width=\textwidth]{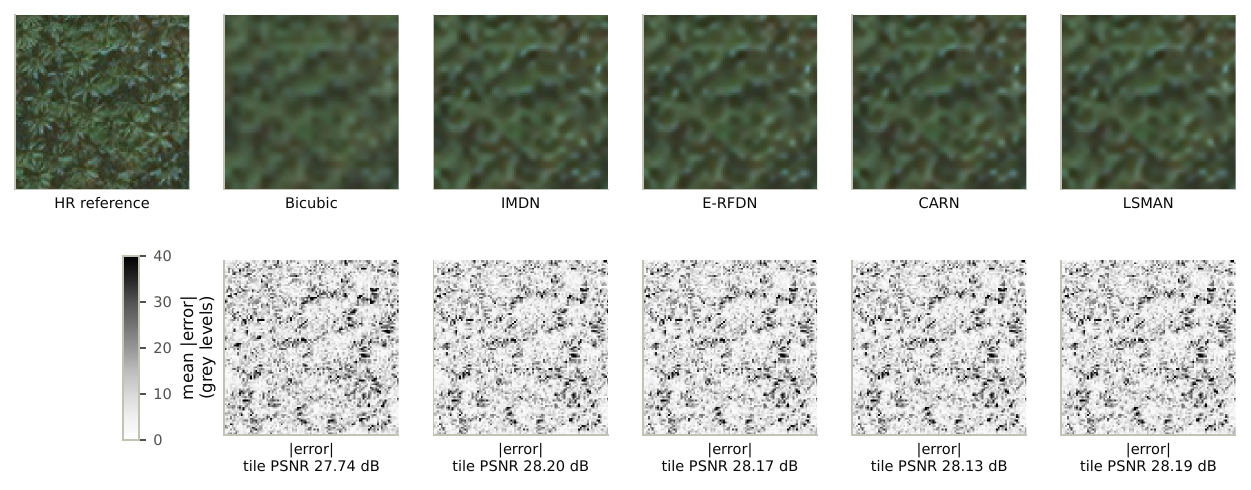}
  \caption{Reconstructions at $\times4$ of a $96 \times 96$ crop of the most textured
    Sat-Test21 tile (top) and their absolute errors in grey levels, averaged over the
    three colour channels (bottom), with the PSNR of the whole tile. All panels share one
    error scale.}
  \label{fig:visual}
\end{figure}

On satellite imagery, the four networks achieve practically equivalent results
(Table~\ref{tab:satellite}). Each improves on bicubic interpolation by 1.39--1.48\,dB at
$\times2$ and 0.68--0.71\,dB at $\times4$, approximately three times the gain of Lanczos
interpolation. All 190 Sat-Pool scenes improve at both scales (Appendix~\ref{app:structure}),
and the networks differ by at most 0.04\,dB. At $\times2$, E-RFDN and IMDN exceed LSMAN by
just 0.02\,dB, while CARN trails by 0.01\,dB. At $\times4$, LSMAN, IMDN and E-RFDN are
indistinguishable on Sat-Test21 and differ by at most 0.004\,dB on Sat-Pool. These results
show that LSMAN delivers competitive satellite-image fidelity with a compact architecture,
although it does not exceed the competitors. Figure~\ref{fig:visual} compares reconstructions of a $96 \times 96$ crop at $\times4$,
selected from the window with the most gradient energy in the most textured Sat-Test21
tile, alongside their absolute-error maps. The four networks score within 0.07\,dB of one
another on this tile (28.13--28.20\,dB, against 27.74\,dB for bicubic), and their error maps
are nearly identical: all struggle with canopy texture unresolved by the observation.
The networks therefore differ mainly in how they reconstruct detail within the observation
operator's null space (Section~\ref{sec:degradation}).

\begin{table}[tbp]
  \centering
  \caption{Matched-budget comparison (PSNR, dB; mean $\pm$ s.d. over three seeds). Every
    network is trained with identical data, patches, optimiser, schedule, 30{,}000 steps
    and seeds. Rows in italics are the values published by the original authors after
    much longer training; they are context, not part of the comparison. The published
    RFDN row is its larger 534\,K/550\,K build.}
  \label{tab:benchmarks}
  \small
\begin{tabular}{@{}lrrrrrr@{}}
\toprule
Method & Parameters & Set5 & Set14 & BSD100 & Urban100 & DIV2K-val \\
\midrule
\multicolumn{7}{@{}l}{\emph{Scale $\times2$}} \\
Bicubic & -- & 33.68 & 30.24 & 29.56 & 26.88 & 32.44 \\
Lanczos & -- & 34.32 & 30.70 & 29.93 & 27.27 & 32.92 \\
\addlinespace
CARN & 964{,}163 & 37.55{\scriptsize\,$\pm$\,0.08} & 33.04{\scriptsize\,$\pm$\,0.04} & 31.88{\scriptsize\,$\pm$\,0.02} & 30.97{\scriptsize\,$\pm$\,0.07} & 35.54{\scriptsize\,$\pm$\,0.04} \\
E-RFDN & 417{,}212 & 37.55{\scriptsize\,$\pm$\,0.07} & 33.01{\scriptsize\,$\pm$\,0.02} & 31.90{\scriptsize\,$\pm$\,0.02} & 31.04{\scriptsize\,$\pm$\,0.06} & 35.58{\scriptsize\,$\pm$\,0.04} \\
IMDN & 694{,}404 & 37.62{\scriptsize\,$\pm$\,0.05} & 33.08{\scriptsize\,$\pm$\,0.00} & 31.89{\scriptsize\,$\pm$\,0.03} & 31.05{\scriptsize\,$\pm$\,0.07} & 35.60{\scriptsize\,$\pm$\,0.03} \\
LSMAN & 355{,}737 & 37.46{\scriptsize\,$\pm$\,0.05} & 32.95{\scriptsize\,$\pm$\,0.02} & 31.79{\scriptsize\,$\pm$\,0.02} & 30.73{\scriptsize\,$\pm$\,0.02} & 35.43{\scriptsize\,$\pm$\,0.02} \\
\addlinespace
\emph{CARN, published} & & \emph{37.76} & \emph{33.52} & \emph{32.09} & \emph{31.92} & -- \\
\emph{IMDN, published} & & \emph{38.00} & \emph{33.63} & \emph{32.19} & \emph{32.17} & -- \\
\emph{RFDN, published} & & \emph{38.05} & \emph{33.68} & \emph{32.16} & \emph{32.12} & -- \\
\addlinespace
\multicolumn{7}{@{}l}{\emph{Scale $\times4$}} \\
Bicubic & -- & 28.43 & 26.00 & 25.96 & 23.14 & 28.11 \\
Lanczos & -- & 28.81 & 26.23 & 26.13 & 23.33 & 28.34 \\
\addlinespace
CARN & 1{,}111{,}875 & 31.59{\scriptsize\,$\pm$\,0.05} & 28.15{\scriptsize\,$\pm$\,0.02} & 27.32{\scriptsize\,$\pm$\,0.01} & 25.38{\scriptsize\,$\pm$\,0.00} & 29.99{\scriptsize\,$\pm$\,0.01} \\
E-RFDN & 433{,}448 & 31.58{\scriptsize\,$\pm$\,0.06} & 28.12{\scriptsize\,$\pm$\,0.01} & 27.31{\scriptsize\,$\pm$\,0.01} & 25.30{\scriptsize\,$\pm$\,0.03} & 30.00{\scriptsize\,$\pm$\,0.01} \\
IMDN & 715{,}176 & 31.68{\scriptsize\,$\pm$\,0.02} & 28.15{\scriptsize\,$\pm$\,0.01} & 27.33{\scriptsize\,$\pm$\,0.01} & 25.36{\scriptsize\,$\pm$\,0.01} & 30.02{\scriptsize\,$\pm$\,0.01} \\
LSMAN & 605{,}145 & 31.46{\scriptsize\,$\pm$\,0.02} & 28.04{\scriptsize\,$\pm$\,0.01} & 27.26{\scriptsize\,$\pm$\,0.00} & 25.21{\scriptsize\,$\pm$\,0.00} & 29.92{\scriptsize\,$\pm$\,0.01} \\
\addlinespace
\emph{CARN, published} & & \emph{32.13} & \emph{28.60} & \emph{27.58} & \emph{26.07} & -- \\
\emph{IMDN, published} & & \emph{32.21} & \emph{28.58} & \emph{27.56} & \emph{26.04} & -- \\
\emph{RFDN, published} & & \emph{32.24} & \emph{28.61} & \emph{27.57} & \emph{26.11} & -- \\
\bottomrule
\end{tabular}

\end{table}

\begin{table}[tbp]
  \centering
  \caption{Satellite tiles (PSNR, dB; seed-averaged). Differences are each method minus
    LSMAN, with 95\% scene-cluster bootstrap intervals and Holm-adjusted Wilcoxon
    $p$-values on scene means.}
  \label{tab:satellite}
  \small
\begin{tabular}{@{}lrlrrlr@{}}
\toprule
 & \multicolumn{3}{c}{Sat-Test21 (21 tiles, 15 scenes)} & \multicolumn{3}{c}{Sat-Pool (735 tiles, 190 scenes)} \\
\cmidrule(lr){2-4}\cmidrule(lr){5-7}
Method & PSNR & $-$ LSMAN [95\% CI] & $p_{\mathrm{Holm}}$ & PSNR & $-$ LSMAN [95\% CI] & $p_{\mathrm{Holm}}$ \\
\midrule
\multicolumn{7}{@{}l}{\emph{Scale $\times2$}} \\
Bicubic & 32.94 & $-$1.450 [$-$1.571, $-$1.325] & $<$0.001 & 34.33 & $-$1.404 [$-$1.436, $-$1.373] & $<$0.001 \\
Lanczos & 33.38 & $-$1.005 [$-$1.104, $-$0.905] & $<$0.001 & 34.79 & $-$0.951 [$-$0.974, $-$0.927] & $<$0.001 \\
CARN & 34.38 & $-$0.012 [$-$0.019, $-$0.005] & 0.007 & 35.73 & $-$0.011 [$-$0.013, $-$0.009] & $<$0.001 \\
E-RFDN & 34.41 & $+$0.026 [$+$0.016, $+$0.040] & $<$0.001 & 35.76 & $+$0.021 [$+$0.017, $+$0.025] & $<$0.001 \\
IMDN & 34.40 & $+$0.017 [$+$0.011, $+$0.025] & $<$0.001 & 35.76 & $+$0.018 [$+$0.016, $+$0.021] & $<$0.001 \\
LSMAN & 34.39 & -- & -- & 35.74 & -- & -- \\
\addlinespace
\multicolumn{7}{@{}l}{\emph{Scale $\times4$}} \\
Bicubic & 28.93 & $-$0.708 [$-$0.790, $-$0.618] & $<$0.001 & 30.32 & $-$0.706 [$-$0.736, $-$0.678] & $<$0.001 \\
Lanczos & 29.14 & $-$0.500 [$-$0.560, $-$0.430] & $<$0.001 & 30.53 & $-$0.501 [$-$0.523, $-$0.479] & $<$0.001 \\
CARN & 29.62 & $-$0.012 [$-$0.024, $-$0.002] & 0.144 & 31.00 & $-$0.026 [$-$0.029, $-$0.023] & $<$0.001 \\
E-RFDN & 29.63 & $-$0.004 [$-$0.014, $+$0.006] & 1.000 & 31.03 & $-$0.004 [$-$0.006, $-$0.001] & 0.002 \\
IMDN & 29.64 & $+$0.004 [$-$0.005, $+$0.017] & 1.000 & 31.03 & $-$0.004 [$-$0.006, $-$0.001] & $<$0.001 \\
LSMAN & 29.64 & -- & -- & 31.03 & -- & -- \\
\bottomrule
\end{tabular}

\end{table}

\section{Spatial Structure}
\label{sec:structure}

\begin{figure}[tbp]
  \centering
  \includegraphics[width=\textwidth]{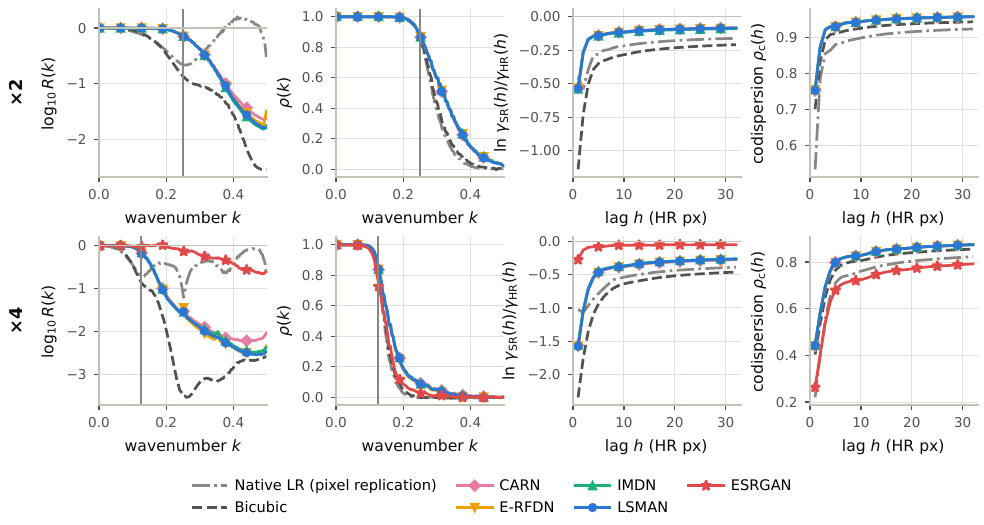}
  \caption{Spatial structure on Sat-Pool at $\times2$ (top row) and $\times4$ (bottom
    row), seed 0. Columns from left: $\log_{10}$ power ratio $R(k)$ and band correlation
    $\rho(k)$ by wavenumber $k$ (cycles per high-resolution pixel; the vertical line is
    the Nyquist frequency $1/(2s)$ of the observation); log ratio of semivariances and
    codispersion $\rho_{\mathrm{c}}(h)$ by lag, averaged over four directions. The four
    networks overlap almost everywhere.}
  \label{fig:structure}
\end{figure}

Figure~\ref{fig:structure} and Table~\ref{tab:structure} compare the reconstructions with
the reference as spatial fields on Sat-Pool.

\medskip
\noindent\textbf{Below and above the Nyquist frequency.} Every network keeps the band
correlation above 0.99 up to the Nyquist frequency of the observation, and raises it
above that frequency from 0.39 (bicubic) to 0.51 at $\times2$ and from 0.28 to 0.37 at
$\times4$. The networks halve the variogram error of bicubic interpolation (0.21 against
0.52 at $\times2$; 0.70--0.72 against 1.25 at $\times4$) and recover more of the variance
of the reference (sill ratio 0.89 against 0.77 at $\times2$). Above the Nyquist
frequency, however, they restore only about 30\% of the power of the reference at
$\times2$ ($\log_{10} R_{\mathrm{sup}} \approx -0.52$) and 15\% at $\times4$
($\approx -0.81$).

\begin{figure}[tbp]
  \centering
  \includegraphics[width=\textwidth]{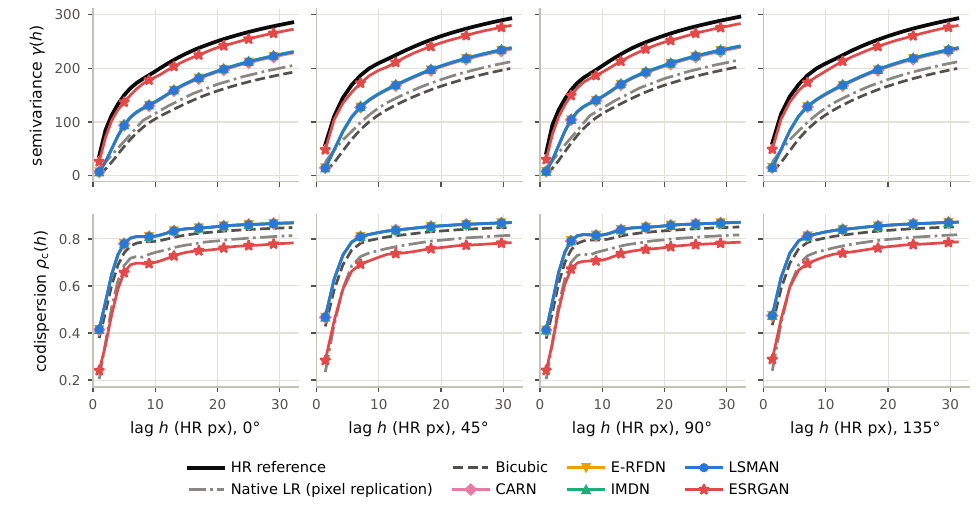}
  \caption{Directional semivariograms (top, grey levels$^2$) and codispersion with the
    reference (bottom) at $\times4$ on Sat-Pool, along $0^\circ$, $45^\circ$, $90^\circ$
    and $135^\circ$ (left to right). Diagonal lags are $\sqrt{2}$ pixels per step. The
    $\times2$ version is in Appendix~\ref{app:structure}.}
  \label{fig:directional}
\end{figure}

\medskip
\noindent\textbf{Energy and agreement.} ESRGAN restores about 78\% of the power above the
Nyquist frequency ($\log_{10} R_{\mathrm{sup}} = -0.11$) and has the smallest variogram
error of any method (0.15), yet its band correlation there (0.21) is lower than that of
bicubic interpolation (0.28). Beyond the shortest lags, its codispersion with the
reference is also lower than that of every interpolator and network, in every direction
(Figure~\ref{fig:directional}). ESRGAN generates texture with the right statistics in the
wrong place, that is, it invents detail. Pixel replication shows the same problem from the
other side. Its blocky edges add aliased power, so its power ratio above the Nyquist
frequency ($-0.22$ at $\times2$) is closer to the reference than that of any network,
while its band correlation (0.21) is the lowest. Measures of energy (power ratio,
variogram) and measures of agreement (band correlation, codispersion) must therefore be
read together.

\medskip
\noindent\textbf{Direction.} Directional variograms (Figure~\ref{fig:directional}, Table~\ref{tab:directional}) show greater structural loss along the image axes than along the diagonals for all methods. For LSMAN at $\times4$, variogram error is 0.78--0.79 at $0^\circ$ and $90^\circ$, compared with 0.63--0.65 at $45^\circ$ and $135^\circ$. Although part of this reflects the longer diagonal lag distance, reconstruction also alters the field structure itself. The reference is strongly anisotropic, with a minimum-to-maximum directional range ratio of 0.49, whereas at $\times4$ the reconstructed fields are more isotropic (0.59--0.60 for the networks and 0.63 for bicubic interpolation; at $\times2$ the networks stay at 0.50). Reconstruction at $\times4$ therefore weakens directional structure. This effect is not captured by PSNR, SSIM or omnidirectional measures, and is an example of a structural loss concealed by pixel-based metrics.

\medskip
\noindent\textbf{Coherence.} The coherence residual of bicubic interpolation is about
2 grey levels, an order of magnitude above that of the networks (0.20--0.31 at $\times2$),
and that of ESRGAN (2.4) is the largest, because invented detail does not average back to
the observation.

\begin{table}[tbp]
  \centering
  \caption{Pixel fidelity, spatial structure and coherence on Sat-Pool (735 tiles; seed
    means). $\varepsilon_\gamma$: variogram error; $\log_{10}R_{\mathrm{sup}}$ and
    $\rho_{\mathrm{sup}}$: power ratio and band correlation above the Nyquist frequency of
    the observation; $\bar\rho_{\mathrm{c}}$: codispersion at lags 1--4; sill and range:
    ratios of the fitted omnidirectional sill and range to those of the reference; $r_D$:
    coherence residual (grey levels). Arrows give the better direction. Sat-Test21 in
    Appendix~\ref{app:structure}.}
  \label{tab:structure}
  \footnotesize
\begin{tabular}{@{}lrrrrrrrrr@{}}
\toprule
 & \multicolumn{2}{c}{Pixel} & \multicolumn{6}{c}{Spatial structure} & Coherence \\
\cmidrule(lr){2-3}\cmidrule(lr){4-9}\cmidrule(lr){10-10}
Method & PSNR & SSIM & $\varepsilon_{\gamma}\downarrow$ & $\log_{10}R_{\mathrm{sup}}$ & $\rho_{\mathrm{sup}}\uparrow$ & $\bar\rho_{\mathrm{c}}\uparrow$ & sill & range & $r_{D}\downarrow$ \\
\midrule
\multicolumn{10}{@{}l}{\emph{Scale $\times2$}} \\
Native LR & 32.99 & 0.842 & 0.344 & $-$0.22 & 0.207 & 0.760 & 0.88 & 0.98 & 1.50 \\
Bicubic & 34.33 & 0.868 & 0.519 & $-$1.17 & 0.394 & 0.835 & 0.77 & 0.63 & 1.99 \\
Lanczos & 34.79 & 0.881 & 0.427 & $-$1.23 & 0.464 & 0.846 & 0.80 & 0.61 & 1.57 \\
CARN & 35.73 & 0.906 & 0.207 & $-$0.52 & 0.510 & 0.870 & 0.89 & 0.76 & 0.31 \\
E-RFDN & 35.76 & 0.906 & 0.204 & $-$0.52 & 0.514 & 0.871 & 0.89 & 0.76 & 0.23 \\
IMDN & 35.76 & 0.906 & 0.208 & $-$0.52 & 0.513 & 0.870 & 0.89 & 0.75 & 0.21 \\
LSMAN & 35.74 & 0.906 & 0.208 & $-$0.52 & 0.510 & 0.870 & 0.89 & 0.75 & 0.20 \\
LSMAN-$\sigma$ & 35.63 & 0.904 & 0.216 & $-$0.52 & 0.495 & 0.867 & 0.89 & 0.76 & 0.41 \\
\addlinespace
\multicolumn{10}{@{}l}{\emph{Scale $\times4$}} \\
Native LR & 29.45 & 0.649 & 0.847 & $-$0.53 & 0.138 & 0.458 & 0.77 & 1.13 & 1.98 \\
Bicubic & 30.32 & 0.686 & 1.249 & $-$1.46 & 0.282 & 0.579 & 0.62 & 0.64 & 2.29 \\
Lanczos & 30.53 & 0.697 & 1.121 & $-$1.54 & 0.321 & 0.590 & 0.64 & 0.59 & 1.87 \\
CARN & 31.00 & 0.730 & 0.702 & $-$0.81 & 0.368 & 0.620 & 0.73 & 0.57 & 0.83 \\
E-RFDN & 31.03 & 0.730 & 0.716 & $-$0.82 & 0.366 & 0.620 & 0.73 & 0.57 & 0.57 \\
IMDN & 31.03 & 0.730 & 0.704 & $-$0.80 & 0.366 & 0.619 & 0.73 & 0.58 & 0.46 \\
LSMAN & 31.03 & 0.730 & 0.713 & $-$0.81 & 0.368 & 0.620 & 0.73 & 0.57 & 0.56 \\
LSMAN-$\sigma$ & 30.98 & 0.728 & 0.723 & $-$0.83 & 0.357 & 0.616 & 0.73 & 0.56 & 0.60 \\
ESRGAN & 28.65 & 0.625 & 0.147 & $-$0.11 & 0.213 & 0.460 & 0.92 & 0.84 & 2.40 \\
\bottomrule
\end{tabular}

\end{table}

\medskip
\noindent\textbf{What the structure measures add.} Structure measures disagree with PSNR in 60 of 128 method pairs (Table~\ref{tab:disagreement}), but almost all disagreements involve the energy measures, power ratio (54 pairs) and variogram error (29). These arise mainly from pixel replication, whose aliasing adds high-frequency energy, and ESRGAN, whose synthetic texture restores energy without preserving location. By contrast, the agreement measures almost always align with PSNR, differing in only three pairs for band correlation and one for codispersion, while the energy and agreement measures rank the four networks identically within measurement precision. The value of the structure measures is therefore diagnostic rather than comparative: a high power ratio with low band correlation indicates invented detail, whereas a low power ratio with high band correlation indicates lost detail. Only the coherence residual reorders the networks: at $\times2$ it favours LSMAN over the higher-PSNR E-RFDN (on both sets) and IMDN (on Sat-Pool), whereas at $\times4$ on Sat-Pool it favours IMDN over LSMAN. Structure measures thus distinguish reconstruction behaviours that a single fidelity score cannot.

\begin{table}[tbp]
  \centering
  \caption{Pairs of methods for which PSNR and a structure measure give significant,
    opposite verdicts (Holm-adjusted Wilcoxon tests on scene means, $p<0.05$). Methods:
    pixel replication, bicubic, Lanczos, CARN, E-RFDN, IMDN, LSMAN, LSMAN-$\sigma$,
    and ESRGAN at $\times4$.}
  \label{tab:disagreement}
  \small
\begin{tabular}{@{}lrrrrr@{}}
\toprule
 & \multicolumn{2}{c}{Sat-Test21} & \multicolumn{2}{c}{Sat-Pool} & \\
\cmidrule(lr){2-3}\cmidrule(lr){4-5}
Structure measure opposing PSNR & $\times2$ & $\times4$ & $\times2$ & $\times4$ & Total \\
\midrule
Method pairs & 28 & 36 & 28 & 36 & 128 \\
\midrule
band correlation $\rho_{\mathrm{sup}}$ & 0 & 0 & 0 & 3 & 3 \\
variogram error $\varepsilon_{\gamma}$ & 2 & 9 & 4 & 14 & 29 \\
codispersion $\bar\rho_{\mathrm{c}}$ & 0 & 0 & 0 & 1 & 1 \\
power ratio $|\log_{10}R_{\mathrm{sup}}|$ & 10 & 15 & 10 & 19 & 54 \\
coherence residual $r_D$ & 4 & 1 & 4 & 3 & 12 \\
\midrule
\textbf{any of the five} & \textbf{12} & \textbf{15} & \textbf{12} & \textbf{21} & \textbf{60} \\
\bottomrule
\end{tabular}

\end{table}

\section{Perceptual Comparison}
\label{sec:esrgan}

ESRGAN optimises a perceptual and adversarial objective and gives up PSNR by design, so it
is also compared on a perceptual measure (Table~\ref{tab:perceptual}). It is far better on
LPIPS (0.21 against 0.49 for LSMAN on Sat-Pool; 0.12 against 0.27 on Urban100) and worse
on PSNR, SSIM and band correlation above the Nyquist frequency. ESRGAN gives the
reconstruction that looks best to a human observer, but for an analysis that relies on
where the detail is, it is worse than bicubic interpolation above the Nyquist frequency.
This is the perception--distortion trade-off of \citet{blau2018perception} seen in the
spatial structure of the reconstruction.

\begin{table}[tbp]
  \centering
  \caption{Perceptual comparison at $\times4$. LPIPS (AlexNet; lower is better) beside
    PSNR, SSIM and the structure measures above the Nyquist frequency of the observation.
    LSMAN is the seed-0 network.}
  \label{tab:perceptual}
  \small
\begin{tabular}{@{}llrrrrr@{}}
\toprule
Set & Method & PSNR & SSIM & LPIPS$\downarrow$ & $\rho_{\mathrm{sup}}\uparrow$ & $\log_{10}R_{\mathrm{sup}}$ \\
\midrule
Urban100 & Bicubic & 23.14 & 0.658 & 0.473 & -- & -- \\
 & LSMAN (seed 0) & 25.22 & 0.756 & 0.274 & -- & -- \\
 & ESRGAN & 24.37 & 0.734 & 0.123 & -- & -- \\
\addlinespace
Sat-Test21 & Bicubic & 28.93 & 0.646 & 0.600 & 0.281 & $-$1.43 \\
 & LSMAN (seed 0) & 29.65 & 0.698 & 0.506 & 0.382 & $-$0.78 \\
 & ESRGAN & 27.38 & 0.591 & 0.224 & 0.225 & $-$0.12 \\
\addlinespace
Sat-Pool & Bicubic & 30.32 & 0.686 & 0.570 & 0.282 & $-$1.46 \\
 & LSMAN (seed 0) & 31.04 & 0.731 & 0.488 & 0.371 & $-$0.81 \\
 & ESRGAN & 28.65 & 0.625 & 0.214 & 0.213 & $-$0.11 \\
\bottomrule
\end{tabular}

\end{table}

\section{Coherence and the Null Space}
\label{sec:coherence-results}

The coherence penalty weight was chosen on DIV2K validation PSNR alone: $\lambda = 0.1$
changes PSNR by less than 0.01\,dB, while $\lambda = 1$ costs 0.12\,dB and $\lambda = 10$
costs 0.56--0.69\,dB. Table~\ref{tab:coherence} shows the effect of the penalty. Under the
bicubic operator it lowers the coherence residual on Sat-Pool from 0.20 to 0.14 grey
levels at $\times2$ and from 0.56 to 0.15 at $\times4$, at no cost in PSNR. Under the
block-average operator it lowers it from 0.26 to 0.15 and from 0.49 to 0.11, and the
constraint layer lowers it further, to 0.14 and 0.07. These two values are the floor set
by rounding the output to 8 bits, $1/(\sqrt{12}\,s)$ grey levels (0.144 and 0.072), so the
constraint layer is coherent up to quantisation, again without loss of PSNR. The reference
itself has a coherence residual of 0.29 grey levels (Section~\ref{sec:coherence}), so with
the penalty or the constraint layer every reconstruction is at least as coherent with the
observation as the true scene.

\begin{table}[tbp]
  \centering
  \caption{Coherence-constrained objectives (seed means). PSNR in dB; $r_D$: coherence
    residual with the observation, RMSE in grey levels, using the operator that made the
    observation. The two operators define different problems and are not comparable with
    each other.}
  \label{tab:coherence}
  \small
\begin{tabular}{@{}lrrrrrr@{}}
\toprule
 & \multicolumn{3}{c}{$\times2$} & \multicolumn{3}{c}{$\times4$} \\
\cmidrule(lr){2-4}\cmidrule(lr){5-7}
Objective & PSNR & $r_D\downarrow$ & PSNR & PSNR & $r_D\downarrow$ & PSNR \\
 & Sat-Pool & Sat-Pool & DIV2K-val & Sat-Pool & Sat-Pool & DIV2K-val \\
\midrule
\multicolumn{7}{@{}l}{\emph{Bicubic observation operator}} \\
LSMAN & 35.74 & 0.201 & 35.43 & 31.03 & 0.562 & 29.92 \\
LSMAN + penalty ($\lambda=0.1$) & 35.73 & 0.144 & 35.43 & 31.04 & 0.149 & 29.92 \\
\addlinespace
\multicolumn{7}{@{}l}{\emph{Block-average observation operator (change of support)}} \\
LSMAN & 35.36 & 0.257 & 34.94 & 30.85 & 0.492 & 29.67 \\
LSMAN + penalty ($\lambda=0.1$) & 35.37 & 0.150 & 34.93 & 30.86 & 0.105 & 29.67 \\
LSMAN + constraint layer & 35.37 & 0.143 & 34.95 & 30.86 & 0.073 & 29.68 \\
\bottomrule
\end{tabular}

\end{table}

Coherence does not require retraining. Five steps of iterative back-projection applied to
the output of the unconstrained network bring its coherence residual to 0.12 ($\times2$)
and 0.06 ($\times4$) grey levels, lower than the trained penalty, while PSNR and band
correlation rise slightly (Table~\ref{tab:projection}).

\begin{table}[tbp]
  \centering
  \caption{Post-hoc projection of the unconstrained LSMAN (seed 0) onto the set of
    reconstructions coherent with the bicubic observation, by iterative back-projection.}
  \label{tab:projection}
  \small
\begin{tabular}{@{}lrrrrrr@{}}
\toprule
 & \multicolumn{3}{c}{$\times2$, Sat-Pool} & \multicolumn{3}{c}{$\times4$, Sat-Pool} \\
\cmidrule(lr){2-4}\cmidrule(lr){5-7}
Post-hoc projection & PSNR & $r_D\downarrow$ & $\rho_{\mathrm{sup}}$ & PSNR & $r_D\downarrow$ & $\rho_{\mathrm{sup}}$ \\
\midrule
none (network output) & 35.75 & 0.198 & 0.512 & 31.04 & 0.532 & 0.371 \\
5 back-projection steps & 35.76 & 0.118 & 0.512 & 31.06 & 0.062 & 0.371 \\
20 back-projection steps & 35.76 & 0.118 & 0.513 & 31.06 & 0.060 & 0.372 \\
\bottomrule
\end{tabular}

\end{table}

Coherence costs nothing because it concerns almost none of the error.
Table~\ref{tab:nullspace} splits the error of the block-average models with
\eqref{eq:nullsplit}. At least 99.7\% of the error energy lies in the null space of the
observation operator, for every model and both scales. The penalty and the constraint
layer remove the part of the error that the observation could correct, which is less than
0.4\% of it, and the total error changes by about 0.2\%. What reconstruction gets wrong is
what the observation cannot determine, and the choice among consistent fields is made by
the prior that the network has learned.

\begin{table}[tbp]
  \centering
  \caption{Reconstruction error on Sat-Test21 (luma RMSE, grey levels) and the share of its
    energy in the null space of the block-average operator, which the observation cannot
    see.}
  \label{tab:nullspace}
  \small
\begin{tabular}{@{}lrrrr@{}}
\toprule
 & \multicolumn{2}{c}{$\times2$} & \multicolumn{2}{c}{$\times4$} \\
\cmidrule(lr){2-3}\cmidrule(lr){4-5}
Model (block-average operator) & error RMSE & null-space share (\%) & error RMSE & null-space share (\%) \\
\midrule
LSMAN & 5.14 & 99.81 & 8.70 & 99.69 \\
LSMAN + penalty & 5.14 & 99.81 & 8.69 & 99.95 \\
LSMAN + constraint layer & 5.14 & 99.80 & 8.69 & 99.96 \\
\bottomrule
\end{tabular}

\end{table}

Since the null-space component is not identified by the data, a point estimate should come
with an uncertainty. The spread of the three independently trained LSMAN networks is a
weak indicator of it. Per pixel, its rank correlation with the error of their mean is only
0.11 at $\times2$ and 0.15 at $\times4$, although the 10\% of pixels with the largest
spread have 1.32 ($\times2$) and 1.41 ($\times4$) times the mean error of the others. Seed
ensembles therefore give a poor estimate of the uncertainty that the null space implies,
and a calibrated estimate remains open.

The dominance of the null-space component may be largely a consequence of the degradation considered here rather than an intrinsic property of learned reconstruction. The low-resolution inputs are generated by a known, noiseless and deterministic operator, so networks trained to invert it are already highly coherent with the observation at $\times2$ (0.20 grey levels versus 0.29 for the quantised reference), leaving little range-space error for a coherence constraint to correct. Real imagery presents a more difficult setting: the point-spread function, sensor noise and compression are uncertain, the observation operator is imperfectly known, and coherence with the true operator is no longer guaranteed. Although the decomposition of \eqref{eq:nullsplit} remains valid, its balance between range-space and null-space error may differ substantially. Repeating the analysis under a blind degradation pipeline would therefore test whether null-space dominance persists when the degradation is unknown. Under the standard evaluation protocol, however, coherence is a necessary property that can be enforced at negligible cost, yet constrains little of the actual reconstruction error.
\chapter{Applications}
\label{ch:applications}

\section{Land-Use Classification}
\label{sec:downstream}

Table~\ref{tab:ucm} shows the UC Merced accuracy under the matched protocol, in which the
probe is trained and tested under the same condition. With the ResNet-50 probe, the
reference reaches 95.95\%. Pixel replication of the observation loses 5.8 points at
$\times2$ and 11.5 points at $\times4$, and bicubic interpolation recovers most of this
loss (95.00\% and 94.48\%). Learned reconstruction adds little. LSMAN reaches 95.52\% and
94.52\%, not significantly different from bicubic interpolation (Holm-adjusted McNemar
$p = 0.48$ and $1.0$), and only IMDN at $\times2$ differs significantly (96.29\%,
$p = 0.008$). With the MobileNetV2 probe, no reconstruction differs from bicubic
interpolation at either scale. At $\times4$, every condition remains significantly below
the reference ($p \le 0.03$ with ResNet-50).

\begin{table}[tbp]
  \centering
  \caption{UC Merced land-use accuracy, matched protocol (probe trained and tested on the
    same condition; five folds blocked by near-duplicate groups; 2{,}100 images). $p$:
    Holm-adjusted exact McNemar test against bicubic interpolation. Gap recovered with
    95\% cluster-bootstrap interval.}
  \label{tab:ucm}
  \footnotesize
\begin{tabular}{@{}lrrlrrl@{}}
\toprule
 & \multicolumn{3}{c}{ResNet-50 probe} & \multicolumn{3}{c}{MobileNetV2 probe} \\
\cmidrule(lr){2-4}\cmidrule(lr){5-7}
Input & Acc.\ (\%) & $p$ & Gap recovered [95\% CI] & Acc.\ (\%) & $p$ & Gap recovered [95\% CI] \\
\midrule
\multicolumn{7}{@{}l}{\emph{Scale $\times2$}} \\
Native LR & 90.2 & $<$0.001 & -- & 94.0 & 0.018 & -- \\
Bicubic & 95.0 & -- & 0.83 [0.69, 0.98] & 95.7 & -- & 0.85 [0.43, 1.38] \\
Bicubic + sharpening & 95.0 & 1.000 & 0.84 [0.68, 1.01] & 95.0 & 0.964 & 0.54 [-0.07, 1.00] \\
IMDN & 96.3 & 0.008 & 1.06 [0.92, 1.23] & 95.4 & 1.000 & 0.73 [0.23, 1.19] \\
LSMAN & 95.5 & 0.483 & 0.93 [0.79, 1.07] & 96.0 & 1.000 & 1.05 [0.66, 1.67] \\
LSMAN + penalty & 95.8 & 0.204 & 0.97 [0.84, 1.12] & 95.9 & 1.000 & 0.95 [0.52, 1.56] \\
HR reference & 96.0 & 0.188 & -- & 96.0 & 1.000 & -- \\
\addlinespace
\multicolumn{7}{@{}l}{\emph{Scale $\times4$}} \\
Native LR & 84.5 & $<$0.001 & -- & 92.1 & $<$0.001 & -- \\
Bicubic & 94.5 & -- & 0.87 [0.78, 0.96] & 94.9 & -- & 0.72 [0.46, 0.98] \\
Bicubic + sharpening & 94.1 & 1.000 & 0.84 [0.75, 0.93] & 94.3 & 0.295 & 0.56 [0.26, 0.80] \\
IMDN & 94.5 & 1.000 & 0.88 [0.78, 0.97] & 94.0 & 0.295 & 0.50 [0.22, 0.71] \\
LSMAN & 94.5 & 1.000 & 0.88 [0.78, 0.96] & 93.9 & 0.238 & 0.46 [0.16, 0.69] \\
LSMAN + penalty & 94.5 & 1.000 & 0.87 [0.78, 0.96] & 94.3 & 0.295 & 0.57 [0.29, 0.82] \\
HR reference & 96.0 & 0.031 & -- & 96.0 & 0.234 & -- \\
\bottomrule
\end{tabular}

\end{table}

Three further checks support this result. First, the sharpening control is never
significantly better than bicubic interpolation and is worse in seven of the eight
backbone, protocol and scale combinations, so a preference of the feature extractor for
sharpened input does not explain any gain. Second, the protocol matters more than the
reconstruction. When the probe is trained on reference images and applied to the
condition (HR-trained protocol, Appendix~\ref{app:tasks}), pixel replication collapses
accuracy to 40.4\% ($\times2$) and 11.5\% ($\times4$), and interpolation recovers most of
it (91.4\% and 81.1\%). Learned reconstruction then changes accuracy by $-0.3$ to $+3.1$
points, significantly in six of the twelve comparisons with bicubic interpolation, and
sharpening loses 11.5 points at $\times2$ with ResNet-50. Third, a ResNet-50 fine-tuned
end to end under each condition reaches 97.5--98.6\% for every input, pixel replication
included, and no condition differs significantly from bicubic interpolation. Over both
backbones, both protocols and both scales, learned reconstruction changes accuracy
relative to bicubic interpolation by $-1.0$ to $+3.1$ points and is significantly
different in 7 of 24 comparisons. In the preliminary experiment, which used LSMAN-$\sigma$, a MobileNetV2 probe and a stratified 80/20 split, LSMAN-$\sigma$ appeared to exceed the
reference (97.38\% against 97.14\%). Under the same matched protocol with blocked folds and paired tests, no reconstruction
differs significantly from the reference at $\times2$, and the largest point estimate
above it (IMDN, gap recovered 1.06) has an interval that includes 1.

\section{Tracking Change in Built Assets}
\label{sec:changedetection}

Monitoring infrastructure means comparing the same location at two dates and deciding
what has changed, so change detection depends on resolving individual structures and on
reconstructing both dates alike. Table~\ref{tab:levir} shows the LEVIR-CD F1,
bootstrapped over source images. Reconstruction recovers part of the gap between the
observation and the reference: at $\times4$, the ResNet-50 F1 rises from 0.650 (pixel
replication) to 0.730 (bicubic) and 0.733--0.753 (networks), against 0.793 for the
reference. No reconstruction differs significantly from bicubic interpolation in any of
the 16 comparisons (Holm-adjusted $p \ge 0.46$; F1 differences $-0.015$ to $+0.024$). In
the preliminary experiment (LSMAN-$\sigma$, MobileNetV2 features), reconstruction at $\times2$ exceeded the reference (0.768 against
0.756). With the ResNet-50 probe used here it does not (LSMAN 0.770 against 0.793). With MobileNetV2,
bicubic interpolation (0.797) scores above the reference (0.782), but the difference is
well inside the bootstrap interval ($p = 1.0$). Reconstruction cannot add information,
and differences of this size are within the noise of the test.

\begin{table}[tbp]
  \centering
  \caption{LEVIR-CD building-change detection (2{,}048 test patches; training and test
    patches from disjoint source images). F1 differences against bicubic interpolation
    with Holm-adjusted bootstrap $p$-values over source images.}
  \label{tab:levir}
  \footnotesize
\begin{tabular}{@{}lrrrlrl@{}}
\toprule
 & \multicolumn{4}{c}{ResNet-50 probe} & \multicolumn{2}{c}{MobileNetV2 probe} \\
\cmidrule(lr){2-5}\cmidrule(lr){6-7}
Input & F1 & Precision & Recall & $\Delta$F1 vs bicubic ($p_{\mathrm{Holm}}$) & F1 & $\Delta$F1 vs bicubic ($p_{\mathrm{Holm}}$) \\
\midrule
\multicolumn{7}{@{}l}{\emph{Scale $\times2$}} \\
Native LR & 0.726 & 0.716 & 0.737 & $-$0.046 (0.036) & 0.769 & $-$0.028 (0.186) \\
Bicubic & 0.772 & 0.787 & 0.758 & -- & 0.797 & -- \\
Bicubic + sharpening & 0.767 & 0.772 & 0.762 & $-$0.005 (1.000) & 0.794 & $-$0.004 (1.000) \\
IMDN & 0.789 & 0.799 & 0.779 & $+$0.017 (0.476) & 0.789 & $-$0.008 (1.000) \\
LSMAN & 0.770 & 0.779 & 0.761 & $-$0.002 (1.000) & 0.782 & $-$0.015 (1.000) \\
LSMAN + penalty & 0.770 & 0.782 & 0.759 & $-$0.002 (1.000) & 0.783 & $-$0.014 (1.000) \\
HR reference & 0.793 & 0.810 & 0.776 & $+$0.020 (0.410) & 0.782 & $-$0.015 (1.000) \\
\addlinespace
\multicolumn{7}{@{}l}{\emph{Scale $\times4$}} \\
Native LR & 0.650 & 0.667 & 0.634 & $-$0.079 (0.003) & 0.726 & $-$0.028 (0.306) \\
Bicubic & 0.730 & 0.719 & 0.741 & -- & 0.754 & -- \\
Bicubic + sharpening & 0.740 & 0.734 & 0.746 & $+$0.010 (0.882) & 0.767 & $+$0.013 (1.000) \\
IMDN & 0.746 & 0.744 & 0.747 & $+$0.016 (0.882) & 0.768 & $+$0.014 (1.000) \\
LSMAN & 0.753 & 0.746 & 0.761 & $+$0.024 (0.460) & 0.748 & $-$0.006 (1.000) \\
LSMAN + penalty & 0.733 & 0.726 & 0.741 & $+$0.004 (0.882) & 0.761 & $+$0.007 (1.000) \\
HR reference & 0.793 & 0.810 & 0.776 & $+$0.063 (0.003) & 0.782 & $+$0.029 (0.306) \\
\bottomrule
\end{tabular}

\end{table}

\section{False Change}
\label{sec:falsechange}

If two observations of an unchanged scene differ only in noise or compression, a model
that fills the null space differently for each reports a change that did not happen.
Table~\ref{tab:falsechange} measures this on the 1{,}113 unchanged LEVIR-CD test
patches. Bicubic interpolation reduces the difference between the two observations
(amplification 0.82--0.86). Every learned reconstruction amplifies it, by 1.27--1.48, and
Laplacian sharpening by up to 1.94. When the observations differ only in their JPEG
compression (quality 85 against 95), 40.5\% of the pixels of LSMAN change by more than 10
grey levels at $\times2$, against 17.5\% for bicubic interpolation and 27.3\% for the
observation itself. The change probe is fairly robust at $\times2$ (at most 1.9\% false
positives for any network), but at $\times4$ with compression it flags 5.8\% (LSMAN),
8.6\% (IMDN) and 8.8\% (LSMAN with the coherence penalty) of the unchanged patches,
against 3.2\% for bicubic interpolation (all realisations in Appendix~\ref{app:tasks}).
The coherence penalty does not reduce false change, as expected from
Section~\ref{sec:coherence-results}: the difference between the two reconstructions lies
in the null space, which coherence does not constrain.

\begin{table}[tbp]
  \centering
  \caption{False change between two reconstructions of the same unchanged scene, whose
    observations differ only in noise ($\sigma = 1$ grey level) or only in JPEG quality
    (85 against 95). amp.: RMS difference between the reconstructions relative to that
    between the observations; px$>$10: share of pixels with change magnitude above 10 grey
    levels; FP: share of patches the ResNet-50 change probe flags as changed.}
  \label{tab:falsechange}
  \small
\begin{tabular}{@{}lrrrrrr@{}}
\toprule
 & \multicolumn{3}{c}{noise $\sigma{=}1$} & \multicolumn{3}{c}{JPEG $q$ 85 vs 95} \\
\cmidrule(lr){2-4}\cmidrule(lr){5-7}
Input & amp. & px$>$10 (\%) & FP (\%) & amp. & px$>$10 (\%) & FP (\%) \\
\midrule
\multicolumn{7}{@{}l}{\emph{Scale $\times2$}} \\
Native LR & 1.00 & 0.0 & 0.0 & 1.00 & 27.3 & 3.0 \\
Bicubic & 0.86 & 0.0 & 0.0 & 0.82 & 17.5 & 0.0 \\
Bicubic + sharpening & 1.94 & 0.8 & 0.0 & 1.91 & 56.6 & 0.2 \\
IMDN & 1.37 & 0.2 & 0.0 & 1.39 & 40.7 & 1.8 \\
LSMAN & 1.30 & 0.1 & 0.0 & 1.38 & 40.5 & 1.8 \\
LSMAN + penalty & 1.36 & 0.1 & 0.0 & 1.40 & 41.2 & 1.9 \\
\addlinespace
\multicolumn{7}{@{}l}{\emph{Scale $\times4$}} \\
Native LR & 1.00 & 0.0 & 0.4 & 1.00 & 34.6 & 32.5 \\
Bicubic & 0.86 & 0.0 & 0.0 & 0.84 & 25.1 & 3.2 \\
Bicubic + sharpening & 1.15 & 0.0 & 0.0 & 1.11 & 38.0 & 1.2 \\
IMDN & 1.48 & 0.8 & 0.1 & 1.41 & 46.7 & 8.6 \\
LSMAN & 1.35 & 0.3 & 0.0 & 1.37 & 46.0 & 5.8 \\
LSMAN + penalty & 1.42 & 0.3 & 0.0 & 1.39 & 46.7 & 8.8 \\
\bottomrule
\end{tabular}

\end{table}

Two practical implications follow. First, bicubic interpolation is the only method that attenuates differences between repeated observations, as its low-pass filtering suppresses the high-frequency variation introduced by noise and compression; all learned reconstructions, as well as the sharpening control, amplify it. For bi-temporal analysis of compressed imagery, interpolation is therefore the safer input unless the change detector is trained specifically on reconstructed image pairs. Second, the change probe is sensitive to image representation. At $\times4$, pixel replication causes compression differences alone to trigger change in 32.5\% of unchanged patches, indicating that part of the apparent benefit of resampling arises from removing block artefacts that activate an ImageNet feature extractor. The false-change rates in Table~\ref{tab:falsechange} should therefore be viewed as properties of the reconstruction--detector combination, and could likely be reduced with detectors trained on matched reconstructed inputs.

\section{Small Objects and Agricultural Parcels}
\label{sec:smallobjects}

Object recognition on NWPU VHR-10 is saturated: every condition exceeds 95\% overall
accuracy. Grouped by size terciles of the longest side of the objects in the reference
(below 62 pixels, 62--84 and 84 or more), the smallest objects show no significant
difference between learned reconstruction and bicubic interpolation at either scale, so
this experiment is reported as a subgroup analysis in Appendix~\ref{app:tasks}. The
parcel-delineation and canopy-counting experiment of the preliminary study scored the
reconstructions against the same detectors applied to the reference image. It measures
the self-consistency of the detectors under degradation, not counting accuracy, and is
reported only as such in Appendix~\ref{app:agri} until a manual reference is available.

\section{Combined Results Across Tasks}
\label{sec:synthesis}

\begin{figure}[tbp]
  \centering
  \includegraphics[width=\textwidth]{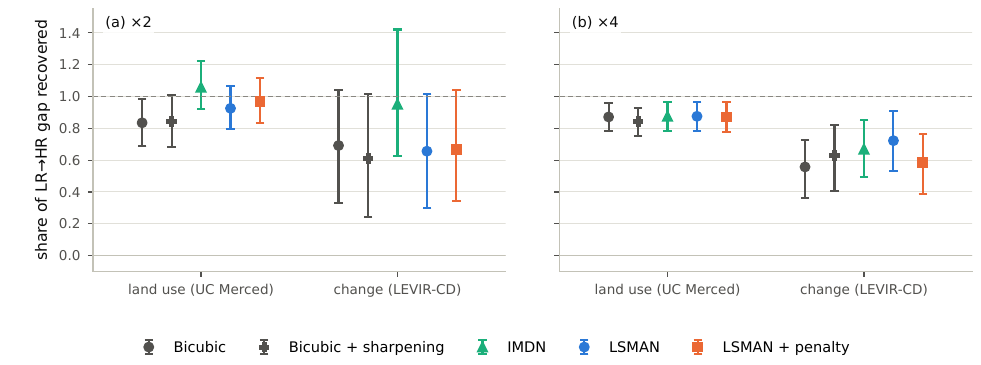}
  \caption{Share of the low-to-high-resolution performance gap recovered, with 95\%
    cluster-bootstrap intervals, at $\times2$ (a) and $\times4$ (b): UC Merced land use
    (ResNet-50 probe, matched protocol) and LEVIR-CD change detection (ResNet-50 probe).
    1 is the reference and 0 the observation. No entry was suppressed: every gap between
    observation and reference was significant.}
  \label{fig:gap}
\end{figure}

Figure~\ref{fig:gap} summarises the two tasks by the share of the gap between the
observation and the reference that each condition recovers, with 95\% intervals. For land
use, bicubic interpolation already recovers 83\% of the gap at $\times2$ and 87\% at
$\times4$, and the intervals of the networks overlap it (LSMAN 93\% and 88\%). For change
detection the intervals are wide (0.24--1.42 at $\times2$) and overlap for every
condition. The data therefore do not show that learned reconstruction recovers more of
the gap than interpolation at either scale. Most of the task benefit comes from
presenting the observation at the resolution that the analysis model expects, which
interpolation already does.

The tasks underpinning this conclusion are scene-level. Land-use classification depends on the arrangement of broad cover types, while building-change detection on $256 \times 256$ patches relies on structures tens of pixels across; neither requires the finest detail available in the reference. Consistent with this, a fine-tuned ResNet-50 achieves 98\% accuracy on pixel-replicated input, indicating that UC~Merced is close to saturated for competent models. These tasks were selected because they provide reference labels suitable for paired, spatially blocked evaluation. The task most likely to benefit from reconstruction, counting individual canopies, lacks an independent manual reference and is therefore reported only as detector self-consistency (Appendix~\ref{app:agri}). The evidence thus indicates that learned reconstruction does not improve scene-level analysis beyond interpolation, while its value for counting and delineation remains unresolved until suitable reference data become available.

\section{The Downlink Budget}
\label{sec:downlink}

\begin{figure}[tbp]
  \centering
  \includegraphics[width=\textwidth]{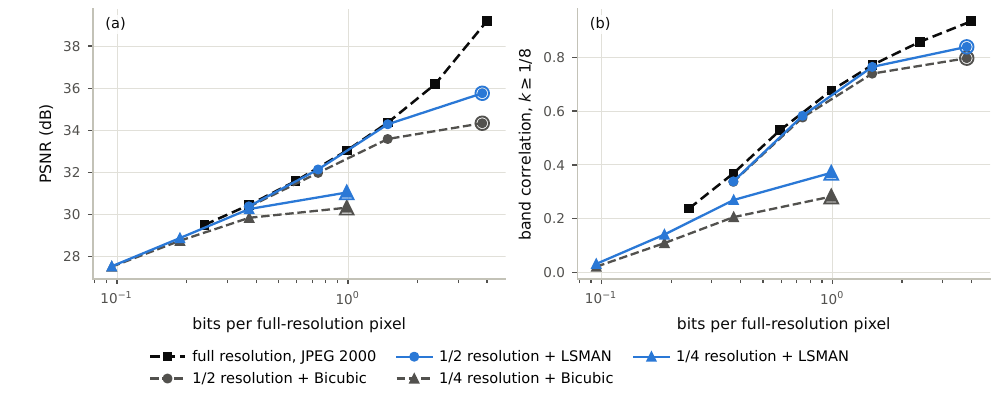}
  \caption{Rate--distortion on Sat-Pool: PSNR (a) and band correlation above the
    $\times4$ Nyquist frequency (b) against transmitted bits per full-resolution pixel,
    for full-resolution JPEG\,2000 and for reduced-resolution transmission (lossless,
    open markers; JPEG\,2000 at compression ratios 4, 8 and 16) reconstructed by bicubic
    interpolation or LSMAN.}
  \label{fig:rd}
\end{figure}

Transmitting images at reduced resolution and reconstructing them on the ground reduces
the downlink volume. The relevant comparison is with what an operator would do anyway: transmit at full resolution with
stronger lossy compression, for which JPEG\,2000
\citep{taubman2002jpeg2000} is a widely used standard. Figure~\ref{fig:rd} and Table~\ref{tab:rd}
compare the two strategies at the same number of transmitted bits per full-resolution
pixel on Sat-Pool. Lossless transmission at full resolution costs 14.5 bits per pixel, or
181.5\,MB and 14.5\,s at 100\,Mbit/s for a $10{,}000 \times 10{,}000$ scene. At every rate,
full-resolution JPEG\,2000 matches or beats reduced-resolution transmission followed by
LSMAN. At about 3.9 bits per pixel it gives 39.18\,dB against 35.76\,dB. At about 1.49
bits per pixel the two are equal (34.37 and 34.28\,dB; band correlation 0.77 and 0.76). At
about 1 bit per pixel JPEG\,2000 gives 33.04\,dB against 31.04\,dB for quarter-resolution
transmission, with a band correlation above the $\times4$ Nyquist frequency of 0.68
against 0.37. Land-use accuracy at matched bitrate does not differ between the two
strategies (95.19\% against 95.52\% at 3.95 bits per pixel, Holm $p = 0.77$; 94.62\%
against 94.52\% at 1.06 bits per pixel, $p = 0.93$; Table~\ref{tab:downlinktasks}). LSMAN
improves on bicubic interpolation of the same transmission at every rate (by 1.42\,dB for
half-resolution lossless transmission), but reduced-resolution transmission is not a
better compression scheme than full-resolution JPEG\,2000. Reconstruction is therefore most
useful for imagery that exists only at low resolution, such as archives and coarser
sensors, and not for saving downlink.

The comparison is not entirely symmetric in computational cost. JPEG\,2000 imposes a substantial onboard processing burden, motivating lighter standards such as CCSDS~122.0 \citep{ccsds2017image}, whereas reduced-resolution transmission compresses only a quarter of the pixels at $\times2$. For platforms constrained by onboard compute, reduced-resolution transmission therefore remains a practical option at intermediate rates, where performance is comparable (34.37 versus 34.28\,dB at approximately 1.5 bits per pixel). What this comparison rules out is the claim that learned reconstruction recovers more information than full-resolution compression loses at the same bitrate.

\begin{table}[tbp]
  \centering
  \caption{Transmission strategies at matched bitrate on Sat-Pool, grouped by rate. Scene
    volume and downlink time are for a $10{,}000 \times 10{,}000$ scene at 100\,Mbit/s;
    $\rho_{k\geq1/8}$: band correlation above the $\times4$ Nyquist frequency.}
  \label{tab:rd}
  \footnotesize
\begin{tabular}{@{}lrrrrrr@{}}
\toprule
Transmission strategy & bits / pixel & scene (MB) & downlink (s) & PSNR & SSIM & $\rho_{k\geq1/8}$ \\
\midrule
full resolution, JPEG 2000 (ratio 6) & 3.99 & 49.8 & 3.99 & 39.18 & 0.953 & 0.934 \\
1/2 resolution, lossless + bicubic & 3.81 & 47.6 & 3.81 & 34.34 & 0.868 & 0.797 \\
1/2 resolution, lossless + LSMAN & 3.81 & 47.6 & 3.81 & 35.76 & 0.906 & 0.839 \\
\addlinespace
full resolution, JPEG 2000 (ratio 16) & 1.49 & 18.6 & 1.49 & 34.37 & 0.865 & 0.771 \\
1/2 resolution, JPEG 2000 (ratio 4) + bicubic & 1.48 & 18.6 & 1.48 & 33.59 & 0.839 & 0.739 \\
1/2 resolution, JPEG 2000 (ratio 4) + LSMAN & 1.48 & 18.6 & 1.48 & 34.28 & 0.864 & 0.764 \\
\addlinespace
full resolution, JPEG 2000 (ratio 24) & 0.99 & 12.3 & 0.99 & 33.04 & 0.818 & 0.677 \\
1/4 resolution, lossless + bicubic & 0.99 & 12.3 & 0.99 & 30.32 & 0.686 & 0.282 \\
1/4 resolution, lossless + LSMAN & 0.99 & 12.3 & 0.99 & 31.04 & 0.731 & 0.371 \\
\addlinespace
full resolution, JPEG 2000 (ratio 64) & 0.37 & 4.6 & 0.37 & 30.45 & 0.695 & 0.368 \\
1/2 resolution, JPEG 2000 (ratio 16) + LSMAN & 0.37 & 4.6 & 0.37 & 30.39 & 0.690 & 0.339 \\
1/4 resolution, JPEG 2000 (ratio 4) + LSMAN & 0.37 & 4.7 & 0.37 & 30.26 & 0.685 & 0.270 \\
\bottomrule
\end{tabular}

\end{table}

\begin{table}[tbp]
  \centering
  \caption{Land-use accuracy at matched bitrate (UC Merced, ResNet-50 probe, matched
    protocol): full-resolution JPEG\,2000 against reduced-resolution lossless
    transmission reconstructed by LSMAN; Holm-adjusted McNemar tests.}
  \label{tab:downlinktasks}
  \small
\begin{tabular}{@{}lrr@{}}
\toprule
Transmission strategy (UC Merced, ResNet-50 probe) & Accuracy (\%) & $p_{\mathrm{Holm}}$ vs JPEG 2000 at equal rate \\
\midrule
HR reference (uncompressed) & 96.0 & -- \\
full resolution, JPEG 2000 at 3.95 bits/pixel & 95.2 & -- \\
1/2 resolution, lossless (3.95 bits/pixel) + LSMAN & 95.5 & 0.773 \\
full resolution, JPEG 2000 at 1.06 bits/pixel & 94.6 & -- \\
1/4 resolution, lossless (1.06 bits/pixel) + LSMAN & 94.5 & 0.929 \\
\bottomrule
\end{tabular}

\end{table}

\section{Deployability}
\label{sec:deployment}

The final consideration is whether LSMAN can be deployed on resource-limited
ground-station hardware. In the preliminary experiment, LSMAN-$\sigma$ took 235\,ms per tile in its training
framework and 590\,ms after export to TensorFlow Lite, measured with different runtimes and
unstated thread counts. Table~\ref{tab:deployment} times every runtime with one harness on
the same tiles (a $128 \times 128$ observation at $\times2$ and $64 \times 64$ at
$\times4$, both giving a $256 \times 256$ tile), on an Intel Xeon Silver 4215R processor
at 3.2\,GHz, as the median of 50 runs after 10 warm-up runs, with the thread count stated.
The thread count alone changes latency threefold (477 against 156\,ms for the $\times2$
network in PyTorch), which explains the preliminary discrepancy. The $\times4$ network is
faster than the $\times2$ network because its convolutions run on a grid four times
smaller (Section~\ref{ch:lsman}). Half precision was slower than single precision in the
preliminary TensorFlow Lite tests, as expected for processors without half-precision arithmetic, so
the deployment path tested here is 8-bit integer quantisation
\citep{jacob2018quantization} with ONNX Runtime, calibrated on DIV2K patches. Quantising
every convolution except the first and the last reduces latency by 1.7--1.9 times on one
thread (256\,ms at $\times2$ and 111\,ms at $\times4$) and the model size to
0.72--0.84\,MB. A $10{,}000 \times 10{,}000$ scene contains about 1{,}526 output tiles, so
four processor threads reconstruct it with the 8-bit model in about 3.6 minutes at $\times2$, and one RTX 3090
graphics processor in about 2 seconds.

\begin{table}[tbp]
  \centering
  \caption{Inference cost of LSMAN for one $256 \times 256$ output tile (Intel Xeon
    Silver 4215R, 3.2\,GHz; median of 50 runs after 10 warm-up runs).}
  \label{tab:deployment}
  \small
\begin{tabular}{@{}lrrrr@{}}
\toprule
 & Model & \multicolumn{2}{c}{CPU latency per tile (ms)} & GPU throughput \\
\cmidrule(lr){3-4}
Runtime & size (MB) & 1 thread & 4 threads & (tiles/s) \\
\midrule
\multicolumn{5}{@{}l}{\emph{Scale $\times2$, tile 128$\rightarrow$256}} \\
PyTorch, fp32 & -- & 477 & 156 & -- \\
ONNX Runtime, fp32 & 1.71 & 452 & 166 & -- \\
ONNX Runtime, int8 (all but first/last conv) & 0.72 & 256 & 140 & -- \\
ONNX Runtime, int8 (block convs only) & 1.01 & 326 & 171 & -- \\
PyTorch, fp32, RTX 3090, batch 1 & -- & -- & -- & 216 \\
PyTorch, fp32, RTX 3090, batch 64 & -- & -- & -- & 862 \\
\addlinespace
\multicolumn{5}{@{}l}{\emph{Scale $\times4$, tile 64$\rightarrow$256}} \\
PyTorch, fp32 & -- & 204 & 77 & -- \\
ONNX Runtime, fp32 & 2.57 & 195 & 74 & -- \\
ONNX Runtime, int8 (all but first/last conv) & 0.84 & 111 & 56 & -- \\
ONNX Runtime, int8 (block convs only) & 1.87 & 165 & 76 & -- \\
PyTorch, fp32, RTX 3090, batch 1 & -- & -- & -- & 206 \\
PyTorch, fp32, RTX 3090, batch 64 & -- & -- & -- & 2225 \\
\bottomrule
\end{tabular}

\end{table}

Super-resolution networks can be sensitive to quantisation, so accuracy was measured again
under it (Table~\ref{tab:int8}). Pixel fidelity barely changes (at most 0.05\,dB and
0.002 SSIM), and land-use accuracy does not differ significantly from single precision.
Coherence, however, degrades: the coherence residual rises from 0.21 to 0.37 grey levels
at $\times2$ and from 0.62 to 0.75 at $\times4$. Keeping the first and last convolutions
in single precision is not enough, while quantising only the block convolutions preserves
coherence (0.24 and 0.64) at a smaller speed-up. A deployment that relies on coherence
should therefore check it after quantisation, or restore it with the post-hoc projection
of Section~\ref{sec:coherence-results}.

\begin{table}[tbp]
  \centering
  \caption{LSMAN under 8-bit integer quantisation (ONNX Runtime): reconstruction and
    structure on Sat-Test21, and UC Merced land-use accuracy (ResNet-50 probe, matched
    protocol) with Holm-adjusted McNemar tests against single precision.}
  \label{tab:int8}
  \small
\begin{tabular}{@{}lrrrrrr@{}}
\toprule
 & \multicolumn{4}{c}{Sat-Test21} & \multicolumn{2}{c}{UC Merced, ResNet-50 probe} \\
\cmidrule(lr){2-5}\cmidrule(lr){6-7}
Precision & PSNR & SSIM & $r_D\downarrow$ & $\rho_{\mathrm{sup}}$ & Accuracy (\%) & $p_{\mathrm{Holm}}$ vs fp32 \\
\midrule
\multicolumn{7}{@{}l}{\emph{Scale $\times2$}} \\
fp32 & 34.40 & 0.895 & 0.213 & 0.525 & 95.5 & -- \\
int8, all but first/last conv & 34.35 & 0.895 & 0.374 & 0.520 & 96.1 & 0.136 \\
int8, block convs only & 34.39 & 0.895 & 0.240 & 0.524 & 95.9 & 0.275 \\
\addlinespace
\multicolumn{7}{@{}l}{\emph{Scale $\times4$}} \\
fp32 & 29.65 & 0.698 & 0.618 & 0.382 & 94.5 & -- \\
int8, all but first/last conv & 29.62 & 0.696 & 0.750 & 0.376 & 95.0 & 0.326 \\
int8, block convs only & 29.65 & 0.698 & 0.638 & 0.381 & 94.5 & 1.000 \\
\bottomrule
\end{tabular}

\end{table}

\chapter{Conclusion}
\label{ch:conclusion}

\section{Summary}

This study evaluated learned single-image super-resolution for satellite imagery as a spatial downscaling problem using a calibrated measurement pipeline that reproduces published baselines within 0.022\,dB. Four lightweight networks trained under a common computational budget served as the evaluation instrument. Five conclusions follow. 

First, learned reconstruction consistently improves pixel-level fidelity relative to bicubic interpolation, by 1.39--1.48\,dB at $\times2$ on satellite imagery. However, architectural differences contribute little: the four networks perform within 0.04\,dB of one another, and removing attention from the reference model changes PSNR by less than 0.025\,dB. 

Second, under the standard protocol of a known, noiseless degradation, at least 99.7\% of reconstruction error lies in the null space of the observation operator. Observation coherence can therefore be enforced to the quantisation limit through penalties, constraint layers or back-projection without reducing fidelity, yet it removes less than 0.4\% of the error. Most reconstruction error is associated with information that the observation cannot determine. 

Third, pixel metrics and structure measures provide complementary information. Energy measures reward both genuine recovery and misplaced detail, agreement measures largely mirror PSNR among comparable networks, and only their combination explains how a reconstruction achieves its score. Directional analysis further reveals a structural loss that pixel metrics overlook: reconstruction makes the field more isotropic than the reference, increasing the range-anisotropy ratio from 0.49 to 0.59 at $\times4$. 

Fourth, learned reconstruction provides little benefit for the scene-level analyses examined here. Across paired, spatially blocked evaluations using two feature extractors and a sharpening control, it changes land-use accuracy by $-1.0$ to $+3.1$ percentage points relative to bicubic interpolation, significantly in 7 of 24 comparisons (6 of them when the classifier is trained on high-resolution images), and building-change F1 by $-0.015$ to $+0.024$, never significantly. Reconstruction also amplifies spurious differences between repeated observations of unchanged scenes by 1.3--1.5$\times$, whereas interpolation attenuates them. 

Fifth, reduced-resolution transmission followed by reconstruction does not outperform full-resolution compression. At matched bitrates, JPEG\,2000 matches or exceeds the reconstruction pipeline in fidelity, preservation of structure above the Nyquist frequency and land-use classification accuracy. 

These results indicate that learned reconstruction is most valuable when imagery exists only at low resolution, such as in historical archives or from inherently coarse sensors, rather than as a mechanism for reducing downlink requirements. For bi-temporal analysis, interpolation remains the safer input. They also suggest that evaluation should extend beyond pixel fidelity to include spatial structure, observation coherence and task-based assessment under paired statistical testing and spatial validation. Two questions remain open and are the ones most likely to alter these conclusions. The first is whether reconstruction improves object-level analyses, which requires independently annotated reference data. The second is whether null-space dominance persists under realistic, uncertain degradations involving unknown blur, noise and compression. Addressing these questions would determine the extent to which the present findings generalise beyond the standard super-resolution benchmark setting.

\section{Limitations}
\label{sec:limitations}

Several limitations bear on how far these conclusions extend. First, LSMAN and its competitors were trained for 30{,}000 steps, one to two orders of magnitude fewer than the published models, so their absolute benchmark values are lower; at this budget the attention gates of LSMAN give no measurable benefit, and whether they would with longer training is not tested. Second, every network was trained for bicubic decimation of clean images and evaluated on JPEG-compressed tiles from one 50\,cm product over three countries, with no real pair of low- and high-resolution acquisitions; real degradations, other sensors and paired datasets such as WorldStrat or SEN2VEN$\mu$S \citep{cornebise2022open,michel2022sen2venus} may change the results. Third, reconstruction is ill-posed, yet only point estimates are evaluated: the spread of three seeds is a weak indicator of error, and no calibrated uncertainty is reported. Finally, the task results rest on two frozen ImageNet backbones and one fine-tuned network, three seeds and one blocked split per dataset, with near-duplicate clusters standing in for spatial blocks where coordinates are missing, and the canopy and parcel experiment has no manual reference, so it supports no claim about counting accuracy.



\section*{Code and Data Availability}

The implementation code is publicly available at
\url{https://doi.org/10.5281/zenodo.23306933}. All datasets used in this study are
publicly available, and the repository provides instructions for sourcing them.

\section*{Funding}
This research did not receive any specific grant from funding agencies in the public, commercial, or not-for-profit sectors.

\section*{Declaration of competing interest}
The authors declare that they have no known competing financial interests or personal relationships that could have appeared to influence the work reported in this paper.

\section*{Declaration of generative AI and AI-assisted technologies in the writing process}
During the preparation of this work the authors used Claude (Anthropic) to draft and edit parts of the manuscript text, to write and troubleshoot code for the analysis, and to generate the scripts that produce the tables and figures from the experimental results. After using this tool, the authors reviewed, tested and edited the content and code as needed and take full responsibility for the content of the published article.

\FloatBarrier
\bibliographystyle{unsrtnat}
\bibliography{references}

\clearpage
\appendix
\chapter{Data, Satellite Tiles and Calibration Results}
\label{app:data}

\section{Datasets}
\label{app:datasets}

Several datasets were used to train the models, evaluate their performance, and assess their utility on downstream tasks. Table~\ref{tab:datasets} lists all datasets used in the study. Low-resolution images for training and evaluation are regenerated from the corresponding high-resolution images using the calibrated degradation pipeline (Section~\ref{sec:calibration}). No low-resolution files from third-party mirrors are used for scoring, and no image is centre-cropped.

\begin{table}[tbp]
  \centering
  \caption{Datasets.}
  \label{tab:datasets}
  \small
  \begin{tabular}{@{}l l r l@{}}
    \toprule
    Dataset & Role & Images & Content \\
    \midrule
    DIV2K training     & training                 & $800$    & natural images, 2K resolution \\
    DIV2K validation   & selection, evaluation    & $100$    & natural images, 2K resolution \\
    Set5, Set14        & evaluation               & $5$, $14$ & natural images \\
    BSD100             & evaluation               & $100$    & natural images \\
    Urban100           & evaluation               & $100$    & urban scenes with repeated structure \\
    Sat-Test21         & evaluation               & $21$     & satellite tiles, 15 scenes, $256\times256$, RGB \\
    Sat-Pool           & evaluation               & $735$    & satellite tiles, 190 scenes, $256\times256$, RGB \\
    UC Merced          & land-use classification  & $2{,}100$ & aerial, 21 classes, $256\times256$ \\
    LEVIR-CD           & change detection         & $2{,}048$ test patches & bi-temporal aerial, $256\times256$ \\
    NWPU VHR-10        & object recognition       & $3{,}896$ objects & ten object classes \\
    \bottomrule
  \end{tabular}
\end{table}

\section{Calibration of the Measurement Path}
\label{app:calibration-path}

This section evaluates the measurement settings used to reproduce the reference bicubic baseline, focusing on low-resolution image generation and luminance handling. Table~\ref{tab:calibration} compares the reference bicubic PSNR values with
those obtained using the different measurement settings. The comparison
documents the effects of low-resolution image generation and luminance
handling on benchmark reproduction. 

\begin{table}[tbp]
  \centering
  \caption{Bicubic interpolation calibration across benchmark datasets and
  scale factors. PSNR (dB) on luminance, excluding a border of $s$ pixels. Prelim.\ table: bicubic
  values of the preliminary experiment's benchmark table. Mirror-LR path: low-resolution files from a
  public mirror, made without trimming the high-resolution image to a multiple of $s$. Corrected path:
  low-resolution images regenerated by antialiased bicubic decimation after trimming; single-channel
  files scored on raw intensity. HR not divisible: images whose size is not a multiple of $s$.}
  \label{tab:calibration}
  \small
\begin{tabular}{@{}llrrrrrrr@{}}
\toprule
 & & & HR not & Prelim. & Mirror-LR & Corrected & Published & Corrected \\
Set & Scale & Images & divisible & table & path & path & value & $-$ published \\
\midrule
Set5 & $\times2$ & 5 & 0 & 33.68 & 33.68 & 33.682 & 33.66 & $+$0.022 \\
Set14 & $\times2$ & 14 & 3 & 29.91 & 29.91 & 30.236 & 30.24 & $-$0.004 \\
BSD100 & $\times2$ & 100 & 100 & 27.42 & 27.42 & 29.565 & 29.56 & $+$0.005 \\
Urban100 & $\times2$ & 100 & 0 & 26.88 & 26.88 & 26.880 & 26.88 & $-$0.000 \\
Set5 & $\times4$ & 5 & 0 & 28.43 & 28.43 & 28.431 & 28.42 & $+$0.011 \\
Set14 & $\times4$ & 14 & 4 & 25.71 & 25.71 & 25.997 & 26.00 & $-$0.003 \\
BSD100 & $\times4$ & 100 & 100 & 25.58 & 25.59 & 25.963 & 25.96 & $+$0.003 \\
Urban100 & $\times4$ & 100 & 0 & 23.14 & 23.14 & 23.142 & 23.14 & $+$0.002 \\
\bottomrule
\end{tabular}

\end{table}

\section{Calibration Results}
\label{app:calibration}

We examined the effects of the decimation filter and luminance convention
on bicubic PSNR, as well as the effect of correcting the measurement path
on previously obtained model results. Table~\ref{tab:conventions} compares
bicubic PSNR under different evaluation conventions. Table~\ref{tab:draftcorrected}
compares LSMAN-$\sigma$ as measured in the preliminary experiment (mirror-LR path) and after re-scoring
on the corrected path, alongside bicubic interpolation evaluated with the
corrected path.

The corrected evaluation changes the measured PSNR by different amounts
across datasets. For example, the BSD100 result of LSMAN-$\sigma$ at $\times2$ changes from 27.24\,dB in the preliminary experiment to 31.46\,dB on
the corrected path. This comparison illustrates how evaluation conventions
can affect measured reconstruction quality and the interpretation of model
performance.

\begin{table}[tbp]
  \centering
  \caption{Bicubic $\times2$ PSNR (dB) on correctly aligned images under the two
    conventions that change measured values most: the decimation filter and the
    luminance definition.}
  \label{tab:conventions}
  \small

\begin{tabular}{@{}llrrr@{}}
\toprule
 & & Full-range & Studio-swing & Studio-swing luma, \\
Set & Decimation & luma & luma & raw single-channel files \\
\midrule
Set14  & Library default (no antialiasing) & 29.53 & 30.85 & 30.76 \\
Set14  & Antialiased bicubic downsampling & 29.01 & 30.33 & 30.24 \\
BSD100 & Library default (no antialiasing) & 28.64 & 29.96 & 29.96 \\
BSD100 & Antialiased bicubic downsampling & 28.24 & 29.56 & 29.56 \\
\bottomrule
\end{tabular}
\end{table}

\begin{table}[tbp]
  \centering
  \caption{LSMAN-$\sigma$ (logistic gates and output, 15{,}000 training steps) as measured in the
    preliminary experiment and re-scored on the corrected path, with bicubic interpolation on the
    corrected path (PSNR, dB). Its preliminary DIV2K value was measured on centre-cropped images.}
  \label{tab:draftcorrected}
  \small
  \begin{tabular}{@{}llrrrrrr@{}}
\toprule
Scale & Model and path & Set5 & Set14 & BSD100 & Urban100 & DIV2K-val & Sat-Test21 \\
\midrule
$\times2$ & LSMAN-$\sigma$, preliminary & 36.98 & 31.59 & 27.24 & 29.81 & 33.81 & 34.29 \\
$\times2$ & LSMAN-$\sigma$, corrected path & 36.98 & 32.57 & 31.46 & 29.81 & 34.92 & 34.27 \\
$\times2$ & Bicubic, corrected path & 33.68 & 30.24 & 29.56 & 26.88 & 32.44 & 32.94 \\
\addlinespace
$\times4$ & LSMAN-$\sigma$, preliminary & 30.91 & 26.92 & 26.15 & 24.73 & 29.59 & 29.61 \\
$\times4$ & LSMAN-$\sigma$, corrected path & 30.90 & 27.70 & 27.01 & 24.72 & 29.59 & 29.60 \\
$\times4$ & Bicubic, corrected path & 28.43 & 26.00 & 25.96 & 23.14 & 28.11 & 28.93 \\
\bottomrule
\end{tabular}

\end{table}

\section{Satellite Tiles: Provenance and Arrangement}
\label{app:tiles}

The tiles were collected without a record of their source, which was therefore established from
their content. The pool contains 5{,}108 tiles of $256 \times 256$ pixels from 190 scenes,
named \texttt{<scene>\_sat\_<row><column>} with row and column indices from
0 to 8. Each scene therefore forms a $9 \times 9$ grid of contiguous tiles
covering at least $2{,}304 \times 2{,}304$ pixels; the 21 Sat-Test21 tiles
are drawn from this pool. This layout matches the $2{,}448 \times 2{,}448$-pixel
scenes in the DeepGlobe Land Cover Classification dataset
\citep{demir2018deepglobe}, whose filenames follow the form
\texttt{<id>\_sat.jpg}. To verify the match, each of the 803 scenes in the dataset's training split
was divided into the same $9 \times 9$ grid, and each grid tile was compared
with every pool tile using the RMS difference of $16 \times 16$ block means.
All 5{,}108 pool tiles match tiles from the training scenes within 0.16 grey
levels, at the row and column positions encoded in their filenames. The matches
pair the 190 pool scenes one to one with 190 DeepGlobe scenes. In the remaining
613 training scenes, the closest grid tile differs from every pool tile by at
least 1.7 grey levels (median 6.6), confirming that the matches are unambiguous.
The 21 Sat-Test21 tiles, representing 15 scenes, are included in these matches.
The matching script and scene correspondence list are released with the code.

DeepGlobe distributes RGB imagery from DigitalGlobe's Vivid+ product at 50\,cm ground
sample distance, collected over Thailand, Indonesia and India, as JPEG files without
per-scene acquisition dates \citep{demir2018deepglobe}. No tile was resampled, so the
tiles have a 50\,cm ground sample distance and each covers $128 \times 128$\,m; the
144-pixel strips at the right and bottom edges of each scene are not used. The tiles
carry the scenes' JPEG compression, which none of the networks saw during training.

\begin{figure}[tbp]
  \centering
  \includegraphics[width=\textwidth]{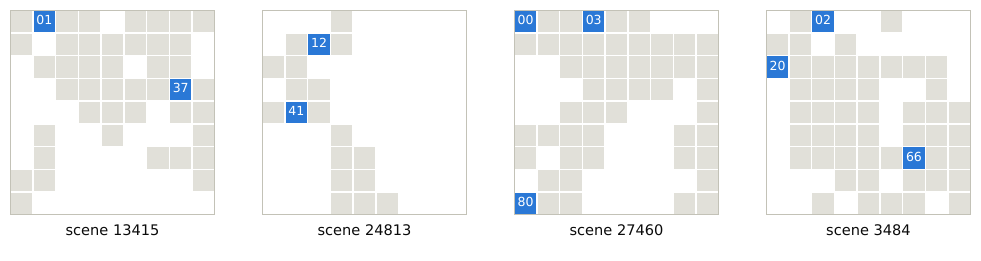}
  \caption{Positions of the Sat-Test21 tiles (blue, labelled with their row and column)
    among the pool tiles (grey) of the four scenes that contribute more than one test
    tile. Each scene is cut into a $9 \times 9$ grid of $256 \times 256$ tiles, each
    covering $128 \times 128$\,m.}
  \label{fig:footprints}
\end{figure}

Because a scene's tiles are spatially adjacent, tiles from one scene are not independent
samples. The 21 tiles of Sat-Test21 come from 15 scenes: eleven scenes contribute one
tile, two contribute two and two contribute three. Figure~\ref{fig:footprints} shows the
positions of the test tiles within the four scenes that contribute more than one, and
Table~\ref{tab:tiles} gives their separations. No two test tiles touch; the closest pair
is 362 pixels (181\,m) apart. Sat-Pool takes at most four tiles from each of 190 scenes,
none of them in Sat-Test21. All satellite statistics resample scenes
(Section~\ref{sec:stats}).

\begin{table}[tbp]
  \centering
  \caption{Separation of Sat-Test21 tiles from the same scene: distance between tile
    centres and gap between tile edges, in tile widths, pixels and metres (50\,cm
    ground sample distance).}
  \label{tab:tiles}
  \small
\begin{tabular}{@{}lrrrrr@{}}
\toprule
Scene & Tile pair & Centre distance (tiles) & Gap (tiles) & Gap (px) & Gap (m) \\
\midrule
13415 & 01 -- 37 & 6.71 & 5.39 & 1379 & 690 \\
24813 & 12 -- 41 & 3.16 & 2.00 & 512 & 256 \\
27460 & 00 -- 03 & 3.00 & 2.00 & 512 & 256 \\
27460 & 00 -- 80 & 8.00 & 7.00 & 1792 & 896 \\
27460 & 03 -- 80 & 8.54 & 7.28 & 1864 & 932 \\
3484 & 02 -- 20 & 2.83 & 1.41 & 362 & 181 \\
3484 & 02 -- 66 & 7.21 & 5.83 & 1493 & 746 \\
3484 & 20 -- 66 & 7.21 & 5.83 & 1493 & 746 \\
\bottomrule
\end{tabular}

\end{table}

\FloatBarrier
\chapter{Training Details and Additional Reconstruction Results}
\label{app:training}

\section{Competitor Configurations}

The three competitors follow their official architectures and are trained exactly as
LSMAN (Table~\ref{tab:training}). IMDN \citep{hui2019lightweight} uses six information
multi-distillation blocks of 64 channels with contrast-aware channel attention. E-RFDN
\citep{liu2020residual,zhang2020aim} uses four residual feature distillation blocks of 50
channels with enhanced spatial attention. CARN \citep{ahn2018fast} is the single-scale
build with 64 channels; the 1.59\,M parameters usually quoted for CARN refer to the
multi-scale model. All upsample by sub-pixel convolution. ESRGAN is the authors' public
$\times4$ checkpoint \citep{wang2019esrgan}.

\section{Convergence}
\label{sec:convergence}

\begin{figure}[tbp]
  \centering
  \includegraphics[width=\textwidth]{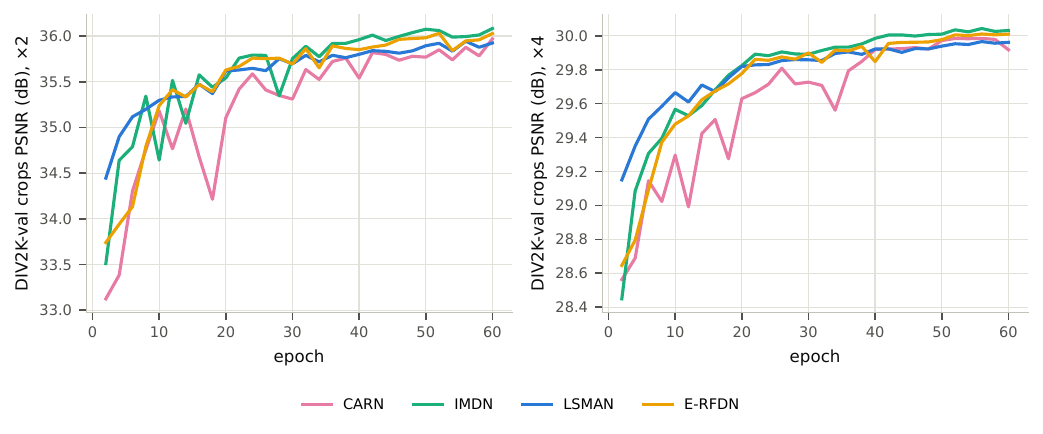}
  \caption{Validation PSNR during training (DIV2K validation crops, seed 0) at $\times2$
    (left) and $\times4$ (right).}
  \label{fig:training}
\end{figure}

Figure~\ref{fig:training} shows the validation PSNR of the seed-0 networks during
training on fixed crops of the DIV2K validation images. Table~\ref{tab:convergence}
summarises the final PSNR and its change over the last 10\% of the 30,000-step
budget. At $\times 2$, CARN, LSMAN, and E-RFDN still show positive gains at the
end of training, with CARN recording the largest increase (+0.090~dB). IMDN,
by contrast, shows a small decline ($-0.012$~dB), suggesting that its validation
performance has largely plateaued under this training setup. LSMAN's gain of
$0.032$~dB indicates that its validation performance was still improving,
although more slowly than CARN's.

At $\times 4$, the final gains are smaller, ranging from $-0.019$ to
$+0.014$~dB. These changes suggest that validation PSNR was approaching a
plateau for all four models within the available budget, although small residual
changes remained for some networks. The final PSNR values are also closely
grouped: the spread between the highest and lowest values is just 0.16~dB at
$\times 2$ and 0.11~dB at $\times 4$. Thus, the fixed training budget produces
similar final reconstruction scores across architectures, but it does not
guarantee that every model has reached convergence. In particular, the positive
late-stage gains for CARN at $\times 2$ indicate that its final score may still
benefit from additional training.

\begin{table}[tbp]
  \centering
  \caption{Convergence at the end of the 30{,}000-step budget (seed 0).}
  \label{tab:convergence}
  \small
\begin{tabular}{@{}llrrc@{}}
\toprule
Scale & Model & Final validation PSNR & Gain over the last 10\% of steps & Still improving \\
\midrule
$\times2$ & CARN & 35.97 & $+$0.090 & yes \\
$\times2$ & IMDN & 36.08 & $-$0.012 & no \\
$\times2$ & LSMAN & 35.92 & $+$0.032 & yes \\
$\times2$ & E-RFDN & 36.03 & $+$0.027 & yes \\
$\times4$ & CARN & 29.92 & $-$0.019 & no \\
$\times4$ & IMDN & 30.03 & $+$0.011 & no \\
$\times4$ & LSMAN & 29.96 & $+$0.014 & no \\
$\times4$ & E-RFDN & 30.01 & $+$0.014 & no \\
\bottomrule
\end{tabular}

\end{table}

\section{Structural Similarity on the Benchmarks}
\label{sec:benchmarks-ssim}

Table~\ref{tab:benchmarks-ssim} reports SSIM for the matched-budget comparison
in Table~\ref{tab:benchmarks}. The learned models consistently outperform
bicubic and Lanczos interpolation across all five benchmarks at both scales,
showing that the networks recover structural information beyond what these
interpolation methods preserve. However, the differences among the learned
models are small at $\times 2$ and somewhat larger at $\times 4$.

At $\times 2$, IMDN achieves the highest SSIM on Set5, Set14, Urban100 and DIV2K-val,
while E-RFDN leads on BSD100. LSMAN scores between 0.0006 and
0.0037 below the best result on each benchmark. At $\times 4$, IMDN achieves
the highest score on BSD100, Urban100, and DIV2K-val, while CARN and IMDN
tie on Set5 and CARN leads on Set14. LSMAN remains below the best-performing
model on every benchmark, with differences ranging from 0.0026 to 0.0060.
Its relative shortfall is therefore more apparent at the larger scale.

These results complement the parameter counts in Table~\ref{tab:benchmarks}.
LSMAN has 355,737 parameters at $\times 2$, fewer than CARN, E-RFDN, and
IMDN, yet its SSIM is consistently slightly lower. At $\times 4$, its
605,145 parameters are also fewer than those of CARN and IMDN, but its
structural similarity scores remain lower. The results therefore support a
qualified conclusion: LSMAN offers a compact alternative with competitive
reconstruction quality, but it does not achieve the best structural similarity
under the matched training budget.

The reported standard deviations across seeds are at most 0.001,
indicating limited variation between the three runs, but this alone does not
establish that every difference is statistically significant across images.
Overall, SSIM reinforces the strong advantage of learned reconstruction over
interpolation while showing that LSMAN's parameter efficiency comes with a
modest loss in structural similarity relative to the strongest competing models.

\begin{table}[tbp]
  \centering
  \caption{Matched-budget comparison, SSIM (mean over three seeds; standard deviations
    are at most 0.001).}
  \label{tab:benchmarks-ssim}
  \small
\begin{tabular}{@{}lrrrrrr@{}}
\toprule
Method & Parameters & Set5 & Set14 & BSD100 & Urban100 & DIV2K-val \\
\midrule
\multicolumn{7}{@{}l}{\emph{Scale $\times2$}} \\
Bicubic & -- & 0.9305 & 0.8692 & 0.8436 & 0.8407 & 0.9040 \\
Lanczos & -- & 0.9366 & 0.8795 & 0.8552 & 0.8507 & 0.9117 \\
\addlinespace
CARN & 964{,}163 & 0.9590 & 0.9132 & 0.8960 & 0.9161 & 0.9413 \\
E-RFDN & 417{,}212 & 0.9591 & 0.9132 & 0.8964 & 0.9173 & 0.9415 \\
IMDN & 694{,}404 & 0.9592 & 0.9135 & 0.8962 & 0.9174 & 0.9416 \\
LSMAN & 355{,}737 & 0.9586 & 0.9123 & 0.8949 & 0.9137 & 0.9403 \\
\addlinespace
\multicolumn{7}{@{}l}{\emph{Scale $\times4$}} \\
Bicubic & -- & 0.8113 & 0.7031 & 0.6682 & 0.6582 & 0.7745 \\
Lanczos & -- & 0.8183 & 0.7113 & 0.6760 & 0.6655 & 0.7809 \\
\addlinespace
CARN & 1{,}111{,}875 & 0.8875 & 0.7718 & 0.7277 & 0.7610 & 0.8278 \\
E-RFDN & 433{,}448 & 0.8861 & 0.7704 & 0.7269 & 0.7589 & 0.8270 \\
IMDN & 715{,}176 & 0.8875 & 0.7714 & 0.7281 & 0.7614 & 0.8280 \\
LSMAN & 605{,}145 & 0.8846 & 0.7688 & 0.7255 & 0.7554 & 0.8253 \\
\bottomrule
\end{tabular}

\end{table}

\FloatBarrier
\chapter{Additional Structure Results}
\label{app:structure}

\begin{figure}[tbp]
  \centering
  \includegraphics[width=\textwidth]{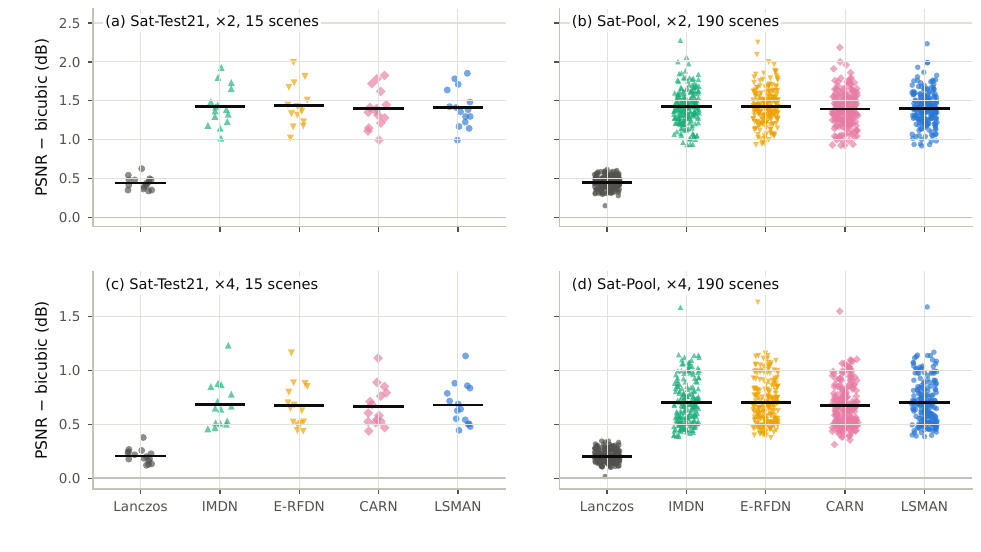}
  \caption{Per-scene PSNR gain over bicubic interpolation (scene means of per-tile
    differences; seed-averaged networks) on Sat-Test21 and Sat-Pool at $\times2$ and
    $\times4$. Bars mark the means.}
  \label{fig:paired}
\end{figure}

\begin{figure}[tbp]
  \centering
  \includegraphics[width=\textwidth]{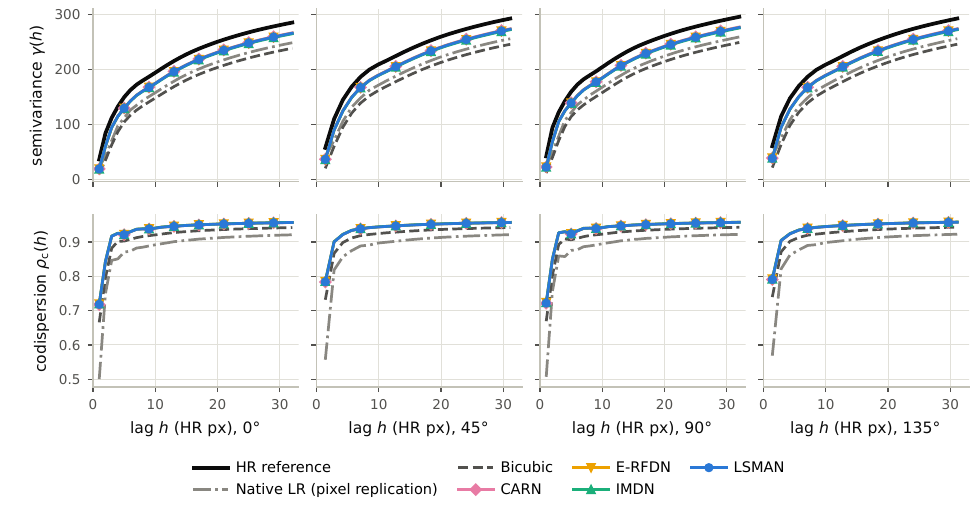}
  \caption{Directional semivariograms (top) and codispersion with the reference (bottom)
    at $\times2$ on Sat-Pool, along $0^\circ$, $45^\circ$, $90^\circ$ and $135^\circ$.}
  \label{fig:directional-x2}
\end{figure}

Figure~\ref{fig:paired} shows the per-scene PSNR gain of each method over bicubic
interpolation, the paired view of Table~\ref{tab:satellite}: every scene gains, by at
least 0.92\,dB at $\times2$ and 0.31\,dB at $\times4$ for every network, and the four
networks' distributions coincide. Table~\ref{tab:structure-test21} repeats
Table~\ref{tab:structure} on Sat-Test21, Table~\ref{tab:directional} resolves the
variogram error and codispersion by direction, and Figure~\ref{fig:directional-x2} is the
$\times2$ counterpart of Figure~\ref{fig:directional}.

\begin{table}[tbp]
  \centering
  \caption{Pixel fidelity, spatial structure and coherence on Sat-Test21 (21 tiles,
    15 scenes); columns as in Table~\ref{tab:structure}.}
  \label{tab:structure-test21}
  \footnotesize
\begin{tabular}{@{}lrrrrrrrrr@{}}
\toprule
 & \multicolumn{2}{c}{Pixel} & \multicolumn{6}{c}{Spatial structure} & Coherence \\
\cmidrule(lr){2-3}\cmidrule(lr){4-9}\cmidrule(lr){10-10}
Method & PSNR & SSIM & $\varepsilon_{\gamma}\downarrow$ & $\log_{10}R_{\mathrm{sup}}$ & $\rho_{\mathrm{sup}}\uparrow$ & $\bar\rho_{\mathrm{c}}\uparrow$ & sill & range & $r_{D}\downarrow$ \\
\midrule
\multicolumn{10}{@{}l}{\emph{Scale $\times2$}} \\
Native LR & 31.63 & 0.824 & 0.338 & $-$0.19 & 0.202 & 0.762 & 0.87 & 0.97 & 1.72 \\
Bicubic & 32.94 & 0.852 & 0.511 & $-$1.16 & 0.389 & 0.836 & 0.77 & 0.62 & 2.28 \\
Lanczos & 33.38 & 0.867 & 0.421 & $-$1.23 & 0.458 & 0.847 & 0.80 & 0.62 & 1.81 \\
CARN & 34.38 & 0.895 & 0.200 & $-$0.49 & 0.524 & 0.872 & 0.89 & 0.76 & 0.32 \\
E-RFDN & 34.41 & 0.896 & 0.199 & $-$0.49 & 0.530 & 0.873 & 0.89 & 0.76 & 0.24 \\
IMDN & 34.40 & 0.896 & 0.201 & $-$0.50 & 0.527 & 0.873 & 0.89 & 0.76 & 0.21 \\
LSMAN & 34.39 & 0.895 & 0.203 & $-$0.51 & 0.525 & 0.873 & 0.89 & 0.76 & 0.22 \\
LSMAN-$\sigma$ & 34.27 & 0.893 & 0.209 & $-$0.50 & 0.505 & 0.869 & 0.89 & 0.76 & 0.44 \\
\addlinespace
\multicolumn{10}{@{}l}{\emph{Scale $\times4$}} \\
Native LR & 28.09 & 0.607 & 0.832 & $-$0.48 & 0.137 & 0.460 & 0.79 & 1.06 & 2.30 \\
Bicubic & 28.93 & 0.646 & 1.232 & $-$1.43 & 0.281 & 0.582 & 0.60 & 0.55 & 2.68 \\
Lanczos & 29.14 & 0.659 & 1.103 & $-$1.53 & 0.322 & 0.592 & 0.61 & 0.51 & 2.18 \\
CARN & 29.62 & 0.698 & 0.677 & $-$0.76 & 0.378 & 0.623 & 0.72 & 0.51 & 0.89 \\
E-RFDN & 29.63 & 0.697 & 0.694 & $-$0.79 & 0.376 & 0.622 & 0.72 & 0.52 & 0.64 \\
IMDN & 29.64 & 0.698 & 0.680 & $-$0.76 & 0.378 & 0.622 & 0.72 & 0.53 & 0.50 \\
LSMAN & 29.64 & 0.698 & 0.689 & $-$0.77 & 0.377 & 0.623 & 0.72 & 0.51 & 0.65 \\
LSMAN-$\sigma$ & 29.60 & 0.696 & 0.699 & $-$0.78 & 0.366 & 0.620 & 0.72 & 0.52 & 0.66 \\
ESRGAN & 27.38 & 0.591 & 0.132 & $-$0.12 & 0.225 & 0.468 & 0.90 & 0.73 & 2.57 \\
\bottomrule
\end{tabular}

\end{table}

\begin{table}[tbp]
  \centering
  \caption{Direction-resolved structure on Sat-Pool: variogram error (lags 1--8) and
    codispersion at lag 1 along each direction, and the anisotropy ratio (minimum over
    maximum fitted directional range) of the reconstruction and of the reference.}
  \label{tab:directional}
  \footnotesize
\begin{tabular}{@{}lrrrrrrrrc@{}}
\toprule
 & \multicolumn{4}{c}{Variogram error $\varepsilon_{\gamma}$ by direction} & \multicolumn{4}{c}{Codispersion at lag 1 by direction} & Anisotropy \\
\cmidrule(lr){2-5}\cmidrule(lr){6-9}
Method & 0$^\circ$ & 45$^\circ$ & 90$^\circ$ & 135$^\circ$ & 0$^\circ$ & 45$^\circ$ & 90$^\circ$ & 135$^\circ$ & SR / HR \\
\midrule
\multicolumn{10}{@{}l}{\emph{Scale $\times2$}} \\
Bicubic & 0.583 & 0.468 & 0.567 & 0.459 & 0.666 & 0.732 & 0.669 & 0.740 & 0.56 / 0.49 \\
IMDN & 0.238 & 0.184 & 0.229 & 0.180 & 0.718 & 0.784 & 0.723 & 0.791 & 0.50 / 0.49 \\
LSMAN & 0.238 & 0.184 & 0.230 & 0.180 & 0.718 & 0.783 & 0.721 & 0.790 & 0.50 / 0.49 \\
\addlinespace
\multicolumn{10}{@{}l}{\emph{Scale $\times4$}} \\
Bicubic & 1.378 & 1.136 & 1.356 & 1.125 & 0.379 & 0.428 & 0.376 & 0.433 & 0.63 / 0.49 \\
IMDN & 0.785 & 0.636 & 0.767 & 0.627 & 0.413 & 0.465 & 0.411 & 0.472 & 0.59 / 0.49 \\
LSMAN & 0.794 & 0.645 & 0.776 & 0.634 & 0.415 & 0.466 & 0.412 & 0.474 & 0.59 / 0.49 \\
ESRGAN & 0.163 & 0.128 & 0.164 & 0.132 & 0.241 & 0.284 & 0.242 & 0.288 & 0.47 / 0.49 \\
\bottomrule
\end{tabular}

\end{table}

\FloatBarrier
\chapter{Additional Task Results}
\label{app:tasks}

\section{Land Use: HR-Trained Probe and Fine-Tuned Network}

Table~\ref{tab:ucm-hr} repeats Table~\ref{tab:ucm} with probes trained on the reference
images and applied to each condition. Table~\ref{tab:ucm-ft} reports a ResNet-50
fine-tuned end to end under each condition (same blocked folds). No condition differs
significantly from bicubic interpolation; only IMDN at $\times4$ falls significantly below
the reference (97.5\% against 98.6\%, Holm-adjusted $p = 0.011$).

\begin{table}[tbp]
  \centering
  \caption{UC Merced land-use accuracy, HR-trained protocol (probe trained on reference
    images, applied to the condition). Columns as in Table~\ref{tab:ucm}.}
  \label{tab:ucm-hr}
  \footnotesize
\begin{tabular}{@{}lrrlrrl@{}}
\toprule
 & \multicolumn{3}{c}{ResNet-50 probe} & \multicolumn{3}{c}{MobileNetV2 probe} \\
\cmidrule(lr){2-4}\cmidrule(lr){5-7}
Input & Acc.\ (\%) & $p$ & Gap recovered [95\% CI] & Acc.\ (\%) & $p$ & Gap recovered [95\% CI] \\
\midrule
\multicolumn{7}{@{}l}{\emph{Scale $\times2$}} \\
Native LR & 40.4 & $<$0.001 & -- & 66.1 & $<$0.001 & -- \\
Bicubic & 91.4 & -- & 0.92 [0.89, 0.94] & 92.0 & -- & 0.87 [0.83, 0.90] \\
Bicubic + sharpening & 79.9 & $<$0.001 & 0.71 [0.66, 0.76] & 90.5 & 0.016 & 0.82 [0.77, 0.86] \\
IMDN & 93.4 & $<$0.001 & 0.95 [0.94, 0.97] & 92.8 & 0.490 & 0.89 [0.86, 0.93] \\
LSMAN & 92.7 & 0.033 & 0.94 [0.92, 0.96] & 92.7 & 0.490 & 0.89 [0.86, 0.92] \\
LSMAN + penalty & 92.7 & 0.033 & 0.94 [0.92, 0.96] & 92.7 & 0.490 & 0.89 [0.86, 0.92] \\
HR reference & 96.0 & $<$0.001 & -- & 96.0 & $<$0.001 & -- \\
\addlinespace
\multicolumn{7}{@{}l}{\emph{Scale $\times4$}} \\
Native LR & 11.5 & $<$0.001 & -- & 38.0 & $<$0.001 & -- \\
Bicubic & 81.1 & -- & 0.82 [0.80, 0.85] & 78.2 & -- & 0.69 [0.65, 0.74] \\
Bicubic + sharpening & 79.8 & 0.166 & 0.81 [0.78, 0.83] & 77.4 & 0.171 & 0.68 [0.64, 0.72] \\
IMDN & 80.8 & 1.000 & 0.82 [0.80, 0.85] & 81.2 & 0.004 & 0.75 [0.71, 0.78] \\
LSMAN & 81.2 & 1.000 & 0.83 [0.80, 0.85] & 80.3 & 0.0499 & 0.73 [0.69, 0.77] \\
LSMAN + penalty & 81.1 & 1.000 & 0.82 [0.80, 0.85] & 80.4 & 0.0499 & 0.73 [0.69, 0.77] \\
HR reference & 96.0 & $<$0.001 & -- & 96.0 & $<$0.001 & -- \\
\bottomrule
\end{tabular}

\end{table}

\begin{table}[tbp]
  \centering
  \caption{UC Merced land-use accuracy of a ResNet-50 fine-tuned end to end under each
    condition (12 epochs, five blocked folds), with Holm-adjusted McNemar tests.}
  \label{tab:ucm-ft}
  \small
\begin{tabular}{@{}lrrr@{}}
\toprule
Input & Accuracy (\%) & $p_{\mathrm{Holm}}$ vs bicubic & $p_{\mathrm{Holm}}$ vs HR \\
\midrule
\multicolumn{4}{@{}l}{\emph{Scale $\times2$}} \\
Native LR & 98.6 & 1.000 & 1.000 \\
Bicubic & 98.4 & -- & 1.000 \\
Bicubic + sharpening & 98.3 & 1.000 & 1.000 \\
IMDN & 98.5 & 1.000 & 1.000 \\
LSMAN & 98.2 & 1.000 & 1.000 \\
LSMAN + penalty & 98.3 & 1.000 & 1.000 \\
HR reference & 98.6 & 1.000 & -- \\
\addlinespace
\multicolumn{4}{@{}l}{\emph{Scale $\times4$}} \\
Native LR & 98.0 & 1.000 & 0.330 \\
Bicubic & 98.3 & -- & 0.432 \\
Bicubic + sharpening & 98.1 & 1.000 & 0.432 \\
IMDN & 97.5 & 0.128 & 0.011 \\
LSMAN & 98.0 & 1.000 & 0.354 \\
LSMAN + penalty & 98.2 & 1.000 & 0.432 \\
HR reference & 98.6 & 1.000 & -- \\
\bottomrule
\end{tabular}

\end{table}

\section{Object Recognition by Size}

Objects of NWPU VHR-10 are cropped from each reconstructed image at their reference
boxes, resized to $224 \times 224$ identically for every condition, and classified with
a ResNet-50 probe; folds are blocked by near-duplicate image groups. Size groups are the
terciles of the longest box side in the reference (62 and 84 pixels). The task is
saturated, and Table~\ref{tab:nwpu} is reported as a subgroup analysis.

\begin{table}[tbp]
  \centering
  \caption{NWPU VHR-10 object recognition accuracy (\%) by size tercile (3{,}896
    objects), with Holm-adjusted McNemar tests against bicubic interpolation.}
  \label{tab:nwpu}
  \footnotesize
\begin{tabular}{@{}lrrrrrrrr@{}}
\toprule
 & \multicolumn{2}{c}{small ($<$62 px)} & \multicolumn{2}{c}{medium (62--84 px)} & \multicolumn{2}{c}{large ($\ge$84 px)} & \multicolumn{2}{c}{all} \\
\cmidrule(lr){2-3}\cmidrule(lr){4-5}\cmidrule(lr){6-7}\cmidrule(lr){8-9}
Input & Acc. & $p_{\mathrm{Holm}}$ & Acc. & $p_{\mathrm{Holm}}$ & Acc. & $p_{\mathrm{Holm}}$ & Acc. & $p_{\mathrm{Holm}}$ \\
\midrule
\multicolumn{9}{@{}l}{\emph{Scale $\times2$}} \\
Native LR & 96.8 & 0.010 & 98.6 & 0.718 & 98.9 & 0.804 & 98.1 & 0.005 \\
Bicubic & 98.3 & -- & 99.0 & -- & 99.0 & -- & 98.8 & -- \\
Bicubic + sharpening & 97.5 & 0.173 & 98.8 & 0.906 & 99.6 & 0.047 & 98.6 & 1.000 \\
IMDN & 97.8 & 0.355 & 99.3 & 0.906 & 99.5 & 0.328 & 98.8 & 1.000 \\
LSMAN & 97.9 & 0.719 & 99.4 & 0.867 & 99.6 & 0.262 & 99.0 & 1.000 \\
LSMAN + penalty & 98.0 & 0.719 & 99.5 & 0.422 & 99.4 & 0.359 & 99.0 & 1.000 \\
HR reference & 99.4 & 0.013 & 99.5 & 0.547 & 99.7 & 0.059 & 99.5 & $<$0.001 \\
\addlinespace
\multicolumn{9}{@{}l}{\emph{Scale $\times4$}} \\
Native LR & 96.6 & 1.000 & 95.4 & 0.034 & 95.2 & $<$0.001 & 95.7 & $<$0.001 \\
Bicubic & 96.9 & -- & 97.2 & -- & 97.6 & -- & 97.2 & -- \\
Bicubic + sharpening & 97.1 & 1.000 & 97.5 & 1.000 & 97.9 & 0.835 & 97.5 & 0.649 \\
IMDN & 97.6 & 0.748 & 96.9 & 1.000 & 98.1 & 0.835 & 97.5 & 0.649 \\
LSMAN & 97.4 & 1.000 & 97.4 & 1.000 & 98.7 & 0.080 & 97.8 & 0.105 \\
LSMAN + penalty & 97.2 & 1.000 & 97.2 & 1.000 & 98.5 & 0.173 & 97.7 & 0.376 \\
HR reference & 99.4 & $<$0.001 & 99.5 & $<$0.001 & 99.7 & $<$0.001 & 99.5 & $<$0.001 \\
\bottomrule
\end{tabular}

\end{table}

\section{False Change: All Realisations}

Table~\ref{tab:falsechange-full} extends Table~\ref{tab:falsechange} to all four pairs
of observations: noise at $\sigma = 1$ and $\sigma = 2$ grey levels, compression at
quality 85 against 95, and noise combined with compression at quality 85.

\begin{table}[tbp]
  \centering
  \caption{False change on 1{,}113 unchanged LEVIR-CD test patches for all four pairs of
    observations. Columns as in Table~\ref{tab:falsechange}.}
  \label{tab:falsechange-full}
  \scriptsize
\begin{tabular}{@{}lrrrrrrrrrrrr@{}}
\toprule
 & \multicolumn{3}{c}{noise $\sigma{=}1$} & \multicolumn{3}{c}{noise $\sigma{=}2$} & \multicolumn{3}{c}{JPEG $q$ 85 vs 95} & \multicolumn{3}{c}{noise $\sigma{=}1$ + JPEG 85} \\
\cmidrule(lr){2-4}\cmidrule(lr){5-7}\cmidrule(lr){8-10}\cmidrule(lr){11-13}
Input & amp. & px$>$10 (\%) & FP (\%) & amp. & px$>$10 (\%) & FP (\%) & amp. & px$>$10 (\%) & FP (\%) & amp. & px$>$10 (\%) & FP (\%) \\
\midrule
\multicolumn{13}{@{}l}{\emph{Scale $\times2$}} \\
Native LR & 1.00 & 0.0 & 0.0 & 1.00 & 0.7 & 0.0 & 1.00 & 27.3 & 3.0 & 1.00 & 6.9 & 0.2 \\
Bicubic & 0.86 & 0.0 & 0.0 & 0.82 & 0.1 & 0.0 & 0.82 & 17.5 & 0.0 & 0.84 & 3.0 & 0.0 \\
Bicubic + sharpening & 1.94 & 0.8 & 0.0 & 1.92 & 34.8 & 0.0 & 1.91 & 56.6 & 0.2 & 1.87 & 24.0 & 0.0 \\
IMDN & 1.37 & 0.2 & 0.0 & 1.35 & 8.7 & 0.0 & 1.39 & 40.7 & 1.8 & 1.35 & 13.4 & 0.1 \\
LSMAN & 1.30 & 0.1 & 0.0 & 1.27 & 6.3 & 0.0 & 1.38 & 40.5 & 1.8 & 1.34 & 13.1 & 0.1 \\
LSMAN + penalty & 1.36 & 0.1 & 0.0 & 1.33 & 7.9 & 0.0 & 1.40 & 41.2 & 1.9 & 1.36 & 13.5 & 0.1 \\
\addlinespace
\multicolumn{13}{@{}l}{\emph{Scale $\times4$}} \\
Native LR & 1.00 & 0.0 & 0.4 & 1.00 & 0.7 & 0.8 & 1.00 & 34.6 & 32.5 & 1.00 & 7.7 & 3.2 \\
Bicubic & 0.86 & 0.0 & 0.0 & 0.83 & 0.1 & 0.0 & 0.84 & 25.1 & 3.2 & 0.85 & 3.8 & 0.0 \\
Bicubic + sharpening & 1.15 & 0.0 & 0.0 & 1.12 & 3.5 & 0.0 & 1.11 & 38.0 & 1.2 & 1.12 & 9.5 & 0.0 \\
IMDN & 1.48 & 0.8 & 0.1 & 1.42 & 11.1 & 0.3 & 1.41 & 46.7 & 8.6 & 1.42 & 15.1 & 1.0 \\
LSMAN & 1.35 & 0.3 & 0.0 & 1.32 & 8.1 & 0.2 & 1.37 & 46.0 & 5.8 & 1.38 & 14.5 & 0.1 \\
LSMAN + penalty & 1.42 & 0.3 & 0.0 & 1.38 & 10.0 & 0.2 & 1.39 & 46.7 & 8.8 & 1.40 & 14.8 & 0.7 \\
\bottomrule
\end{tabular}

\end{table}

\FloatBarrier
\chapter{Agricultural Parcels and Canopies: Detector Self-Consistency}
\label{app:agri}

Precision agriculture works per parcel and per plant: where a field boundary runs and how
many trees stand in an orchard. This experiment tests whether these two quantities
survive decimation and reconstruction, using the 21 Sat-Test21 tiles. Parcels are
delineated by a graph-based segmentation and tree canopies are detected as blobs, and the
same two detectors, with the same settings, are applied to the reference tile and to
every condition.

No manual annotation of parcels or canopies exists for these tiles, so the detector
output on the reference tile is used as the reference. The reference tile therefore
scores perfectly by construction, and every other condition is scored by how closely the
detectors reproduce on it what they found on the reference tile. This measures the
\emph{self-consistency} of the detectors under degradation, not the accuracy of
counting or delineation: it cannot tell whether a detected canopy or boundary is real,
and a reconstruction that sharpens texture can raise agreement with a contrast-sensitive
detector whether or not the sharpened canopies exist. Accuracy requires a manual
reference, with canopy centres and parcel boundaries annotated on the reference tiles
without viewing any reconstruction. Until such a reference is available, the results
below describe detector agreement only, and no claim about counting or delineation
accuracy is drawn from them.

\medskip
\noindent\textbf{Procedure.} Parcels are segmented with the graph-based method of
\citet{felzenszwalb2004efficient} (scale 200, $\sigma = 0.8$, minimum segment 200
pixels) and scored by boundary F1 within 2 pixels (1\,m at the tiles' 50\,cm ground
sample distance) and by the intersection over union of each reference parcel with its
best-overlapping segment. Canopies are detected as Laplacian-of-Gaussian blobs
\citep{lindeberg1998feature} ($\sigma$ from 2 to 6, threshold 0.04) and matched one to
one within 6 pixels (3\,m). Table~\ref{tab:agri} reports
agreement with the detectors applied to the reference tile on the 21 Sat-Test21 tiles,
including the models trained under the block-average operator with and without the
coherence penalty and the constraint layer. Table~\ref{tab:agri-sens} reports the
sensitivity of canopy agreement to the blob threshold (averaged over $\sigma_{\max} \in
\{4, 6, 8\}$) and of boundary agreement to the segmentation scale.

\begin{table}[tbp]
  \centering
  \caption{Detector self-consistency on Sat-Test21: agreement of the parcel and canopy
    detectors applied to each condition with the same detectors applied to the
    reference tile. Not an accuracy measure (see text).}
  \label{tab:agri}
  \small
\begin{tabular}{@{}lrrrrrr@{}}
\toprule
 & \multicolumn{2}{c}{Parcels} & \multicolumn{4}{c}{Canopies} \\
\cmidrule(lr){2-3}\cmidrule(lr){4-7}
Input & boundary F1 & parcel IoU & precision & recall & F1 & $|$count error$|$ (\%) \\
\midrule
\multicolumn{7}{@{}l}{\emph{Scale $\times2$}} \\
Native LR & 0.575 & 0.369 & 0.995 & 0.800 & 0.886 & 19.6 \\
Bicubic & 0.593 & 0.380 & 0.995 & 0.820 & 0.898 & 17.6 \\
Bicubic + sharpening & 0.576 & 0.423 & 0.686 & 0.988 & 0.807 & 46.5 \\
IMDN & 0.615 & 0.444 & 0.982 & 0.971 & 0.977 & 1.5 \\
LSMAN & 0.610 & 0.421 & 0.978 & 0.968 & 0.973 & 1.3 \\
LSMAN + penalty & 0.586 & 0.410 & 0.980 & 0.973 & 0.976 & 1.2 \\
\multicolumn{7}{@{}l}{\quad\emph{block-average operator}} \\
Native LR & 0.582 & 0.383 & 0.990 & 0.788 & 0.877 & 20.4 \\
LSMAN & 0.588 & 0.400 & 0.966 & 0.928 & 0.947 & 3.9 \\
LSMAN + penalty & 0.602 & 0.411 & 0.963 & 0.932 & 0.947 & 3.2 \\
LSMAN + constraint layer & 0.598 & 0.409 & 0.963 & 0.932 & 0.947 & 3.4 \\
\addlinespace
\multicolumn{7}{@{}l}{\emph{Scale $\times4$}} \\
Native LR & 0.472 & 0.304 & 0.981 & 0.395 & 0.556 & 59.7 \\
Bicubic & 0.428 & 0.223 & 0.994 & 0.352 & 0.512 & 64.5 \\
Bicubic + sharpening & 0.458 & 0.244 & 0.980 & 0.546 & 0.698 & 44.3 \\
IMDN & 0.508 & 0.289 & 0.974 & 0.623 & 0.758 & 36.0 \\
LSMAN & 0.507 & 0.268 & 0.979 & 0.611 & 0.750 & 37.5 \\
LSMAN + penalty & 0.507 & 0.269 & 0.973 & 0.624 & 0.758 & 35.9 \\
\multicolumn{7}{@{}l}{\quad\emph{block-average operator}} \\
Native LR & 0.476 & 0.318 & 0.974 & 0.441 & 0.600 & 54.7 \\
LSMAN & 0.490 & 0.272 & 0.976 & 0.582 & 0.726 & 40.4 \\
LSMAN + penalty & 0.483 & 0.273 & 0.974 & 0.595 & 0.736 & 38.9 \\
LSMAN + constraint layer & 0.502 & 0.274 & 0.973 & 0.591 & 0.732 & 39.2 \\
\bottomrule
\end{tabular}

\end{table}

\begin{table}[tbp]
  \centering
  \caption{Sensitivity of detector self-consistency to the Laplacian-of-Gaussian
    threshold (canopy F1) and to the Felzenszwalb scale (boundary F1). Conditions
    reconstructed under the block-average operator are omitted: their sensitivity runs
    used constraint-layer models with a ReLU output that did not train
    (Section~\ref{sec:lsmancoherence}).}
  \label{tab:agri-sens}
  \small
\begin{tabular}{@{}lrrrrrrrr@{}}
\toprule
 & \multicolumn{5}{c}{canopy F1 by LoG threshold} & \multicolumn{3}{c}{boundary F1 by Felzenszwalb scale} \\
\cmidrule(lr){2-6}\cmidrule(lr){7-9}
Input & 0.02 & 0.03 & 0.04 & 0.06 & 0.08 & 100 & 200 & 400 \\
\midrule
\multicolumn{9}{@{}l}{\emph{Scale $\times2$}} \\
Native LR & 0.913 & 0.898 & 0.886 & 0.832 & 0.793 & 0.623 & 0.575 & 0.513 \\
Bicubic & 0.914 & 0.904 & 0.898 & 0.857 & 0.833 & 0.634 & 0.593 & 0.481 \\
IMDN & 0.972 & 0.974 & 0.976 & 0.976 & 0.977 & 0.652 & 0.615 & 0.554 \\
LSMAN & 0.972 & 0.972 & 0.973 & 0.977 & 0.979 & 0.645 & 0.610 & 0.563 \\
LSMAN + penalty & 0.972 & 0.972 & 0.976 & 0.977 & 0.978 & 0.649 & 0.586 & 0.560 \\
\addlinespace
\multicolumn{9}{@{}l}{\emph{Scale $\times4$}} \\
Native LR & 0.670 & 0.615 & 0.555 & 0.421 & 0.323 & 0.564 & 0.473 & 0.381 \\
Bicubic & 0.616 & 0.561 & 0.511 & 0.398 & 0.308 & 0.498 & 0.428 & 0.385 \\
IMDN & 0.784 & 0.769 & 0.757 & 0.711 & 0.683 & 0.559 & 0.508 & 0.425 \\
LSMAN & 0.778 & 0.761 & 0.748 & 0.700 & 0.675 & 0.556 & 0.507 & 0.429 \\
LSMAN + penalty & 0.786 & 0.772 & 0.758 & 0.707 & 0.672 & 0.553 & 0.507 & 0.453 \\
\bottomrule
\end{tabular}

\end{table}

Two observations hold under every detector setting, but only as statements about
self-consistency. At $\times2$, learned reconstructions reproduce the detector's canopies
on the reference far more closely than interpolation does (canopy agreement F1 about
0.97 against 0.83--0.91 for bicubic interpolation), whereas at $\times4$ every condition
loses more than a quarter of them. The coherence penalty and the constraint layer change
neither parcel nor canopy agreement appreciably.

\end{document}